\documentclass[10pt,twocolumn,letterpaper]{article}

\usepackage[pagenumbers]{iccv}
\usepackage{booktabs}
\usepackage{multirow}
\usepackage{makecell}
\usepackage{colortbl}
\usepackage{graphicx}

\definecolor{iccvblue}{rgb}{0.21,0.49,0.74}
\usepackage[pagebackref,breaklinks,colorlinks,allcolors=iccvblue]{hyperref}

\title{How to Design and Train Your \\Implicit Neural Representation for Video Compression}

\author{
  Matthew Gwilliam \quad Roy Zhang \quad Namitha Padmanabhan \quad Hongyang Du \quad Abhinav Shrivastava \\
  University of Maryland, College Park \\
}

\begin{document}
\maketitle

\begin{abstract}
    Implicit neural representation (INR) methods for video compression have recently achieved visual quality and compression ratios that are competitive with traditional pipelines.
    However, due to the need for per-sample network training, the encoding speeds of these methods are too slow for practical adoption.
    We develop a library to allow us to disentangle and review the components of methods from the NeRV family, reframing their performance in terms of not only size-quality trade-offs, but also impacts on training time.
    We uncover principles for effective video INR design and propose a state-of-the-art configuration of these components, Rabbit NeRV (RNeRV).
    When all methods are given equal training time (equivalent to 300 NeRV epochs) for 7 different UVG videos at 1080p, RNeRV achieves +1.27\% PSNR on average compared to the best-performing alternative for each video in our NeRV library.
    We then tackle the encoding speed issue head-on by investigating the viability of hyper-networks, which predict INR weights from video inputs, to disentangle training from encoding to allow for real-time encoding.
    We propose masking the weights of the predicted INR during training to allow for variable, higher quality compression, resulting in 1.7\% improvements to both PSNR and MS-SSIM at 0.037 bpp on the UCF-101 dataset, and we increase hyper-network parameters by 0.4\% for 2.5\%/2.7\% improvements to PSNR/MS-SSIM with equal bpp and similar speeds.
\end{abstract}

\section{Introduction}
\label{sec:intro}

\begin{figure}[t]
  \centering
  \includegraphics[width=0.8\linewidth]{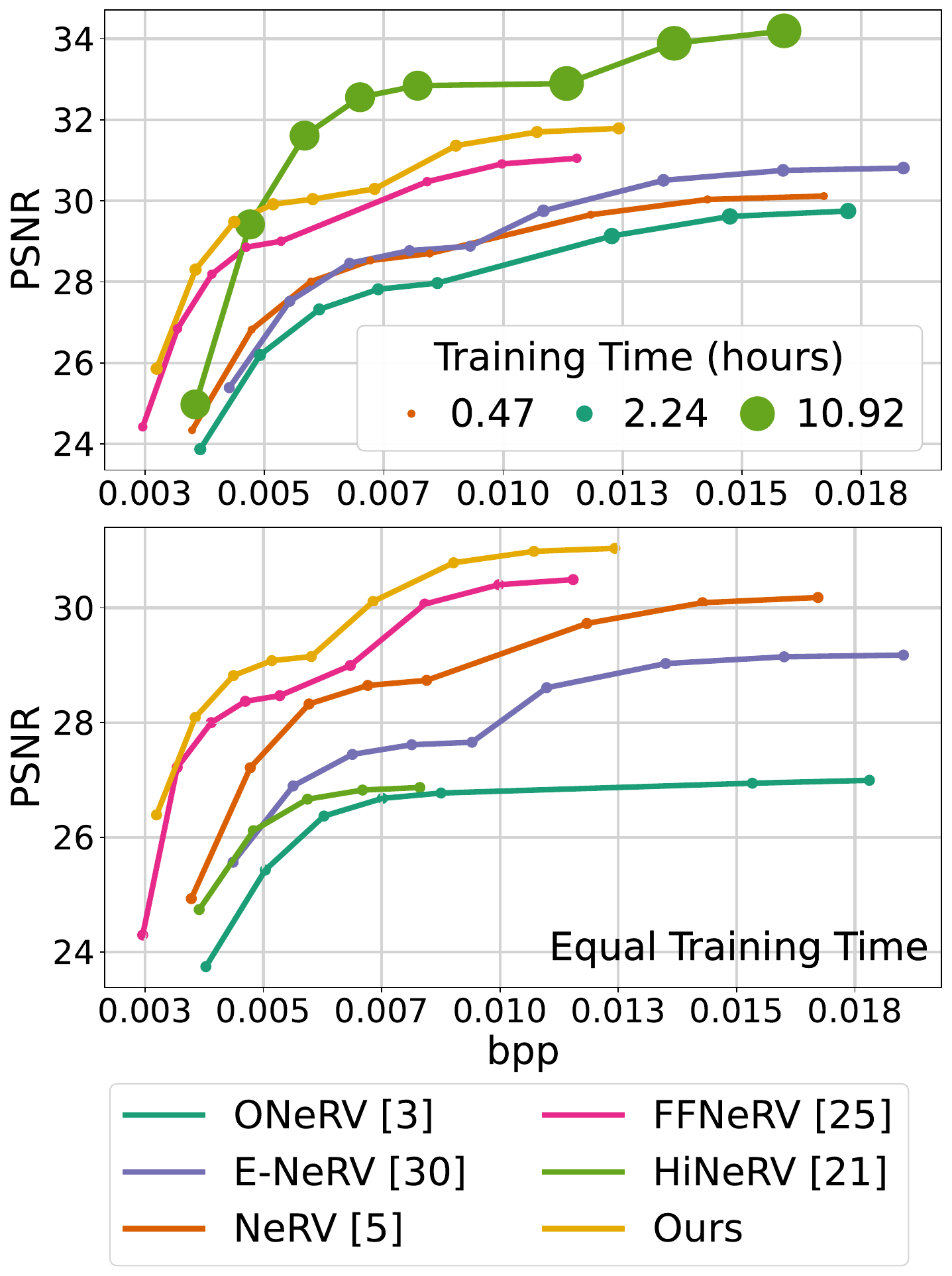}

   \caption{\textbf{Compression performance of INR-based video codecs} from the NeRV family. We examine not only size (bits per pixel) and quality (peak singal-noise ratio), but also encoding speed. Since INRs must be trained for each sample, the encoding speed is dominated by the training time. \textbf{(top)} PSNR/bpp with time as dot size. \textbf{(bottom)} PSNR/bpp for equal time (30 minutes on RTXA5000), averaged over 7 UVG videos at 1080p.}
   \label{fig:teaser}
\end{figure}

Implicit neural representations (INRs) are very appealing for video compression due to their fast decoding speeds.
Many neural network researchers have tried to apply deep learning to the video compression problem in attempts to (1) improve video quality while (2) reducing the storage size~\cite{Rippel_2019_ICCV,Agustsson_2020_CVPR}.
One popular family of approaches, Neural Video Compression (NVC)~\cite{li2021deepcontextualvideocompression}, offers promising performance in both those areas.
However, these methods suffer from slow decoding speeds, which limit adoption for practical use. 
By contrast, INRs boast real-time decoding speeds at low storage size, and recent developments have improved the reconstruction quality substantially, to the point of being more competitive with NVC and traditional codecs~\cite{kim2023c3highperformancelowcomplexityneural,kwan2023hinerv}.
However, since INR-based approaches require training a neural network for each sample individually as part of the encoding, encoding speeds are incredibly slow.

We consider size, quality, encoding time, and decoding speed as 4 equally important criteria for judging the effectiveness of video compression methods.
So, we propose to tackle video INR's greatest weakness, encoding speed (training time), head-on.
We find that in general the encoding time is both under-studied and under-valued.
While existing research will typically make comparisons fair in terms of size by measuring both the number of parameters and the bits-per-pixel (bpp), for encoding speed they will typically report results with equal epochs~\cite{chen2023hnerv,kwan2023hinerv}.
This neglects the actual real cost of training these methods, and encourages the introduction of parameters in configurations that results in large amounts of FLOPs and high wall time.

We illustrate the impact of such a paradigm in Figure~\ref{fig:teaser}. With equal training iterations, HiNeRV~\cite{kwan2023hinerv} dominates compared to its predecessors.
However, with equal training time, such by allowing all methods the amount of time it takes NeRV to run 300 epochs as measured on a single NVIDIA RTXA5000 GPU (30 minutes), the landscape changes drastically.
FFNeRV~\cite{Lee_2023} emerges as the most effective method on average in this setting.
By contrast, HiNeRV is more similar to the performance of the Original NeRV~\cite{chen2021nerv} (ONeRV), while the updated NeRV proposed in the HNeRV paper~\cite{chen2023hnerv} (which has a simpler stem) is significantly better than both.
However, as we show in Section~\ref{sec:experiments}, HiNeRV is still the best for longer encoding times.

With encoding speed concerns in mind, we disentangle NeRV-like methods to reveal principles for optimizing the desired combination of training time, storage size, and reconstruction quality.
We use these principles to formulate our Rabbit NeRV (RNeRV), which we name for its faster training times (and rabbits are associated with speed in many cultures).
We also analyze the differences between these methods through a qualitative lens by extending the XINC~\cite{Padmanabhan_2024_CVPR} method for the rest of the NeRV family.

To further address the encoding time issue, we turn towards hyper-networks~\cite{chen2022transformers}.
By taking a video as input and predicting the INR weights directly, one can skip the per-sample fitting process altogether, and instead train the hyper-network before encoding time.
We propose to use hyper-networks to compress entire videos, 8 frames at a time.
We refer to this architecture, based on NeRV-Enc and NeRV-Dec~\cite{chen2024fastencodingdecodingimplicit} as ``HyperNeRV.'' 
We devise a training strategy, Weight Token Masking, where we randomly mask parts of the hyper-network's weight predictions during training.
This enables us to mask the same parts at inference time, with minimal decrease in quality.
Since we do not need to store the masked parts, we are able to perform compression with a flexible number of bits that can be determined at encoding time (by choosing to mask, or not to mask).
We also show how we can modify size of the ``shared parameters'' of the hyper-network to achieve better video quality at equal compressed size.

In summary, we propose that in light of the promising improvements for video INR methods in terms of rate-distortion, the community must focus more on encoding time.
We make the following key contributions:
\begin{itemize}
    \item We develop a library that disentangles various state-of-the-art video INRs, and use this codebase to distill and elucidate best principles for NeRV design.
    \item We provide a concrete configuration, RNeRV, that achieves optimal performance for different encoding (training) time budgets -- 1.27\% PSNR and 0.72\% MS-SSIM improvements on average compared to the next-best method for 30 minutes training time on UVG.
    \item We propose weight token masking to enable the model to encode at 2 different storage sizes, with 1.7\% improvement to PSNR and MS-SSIM on UCF-101.
\end{itemize}

\section{Related Work}
\label{sec:related_work}

\begin{figure*}[t]
  \centering
   \includegraphics[width=0.90\linewidth]{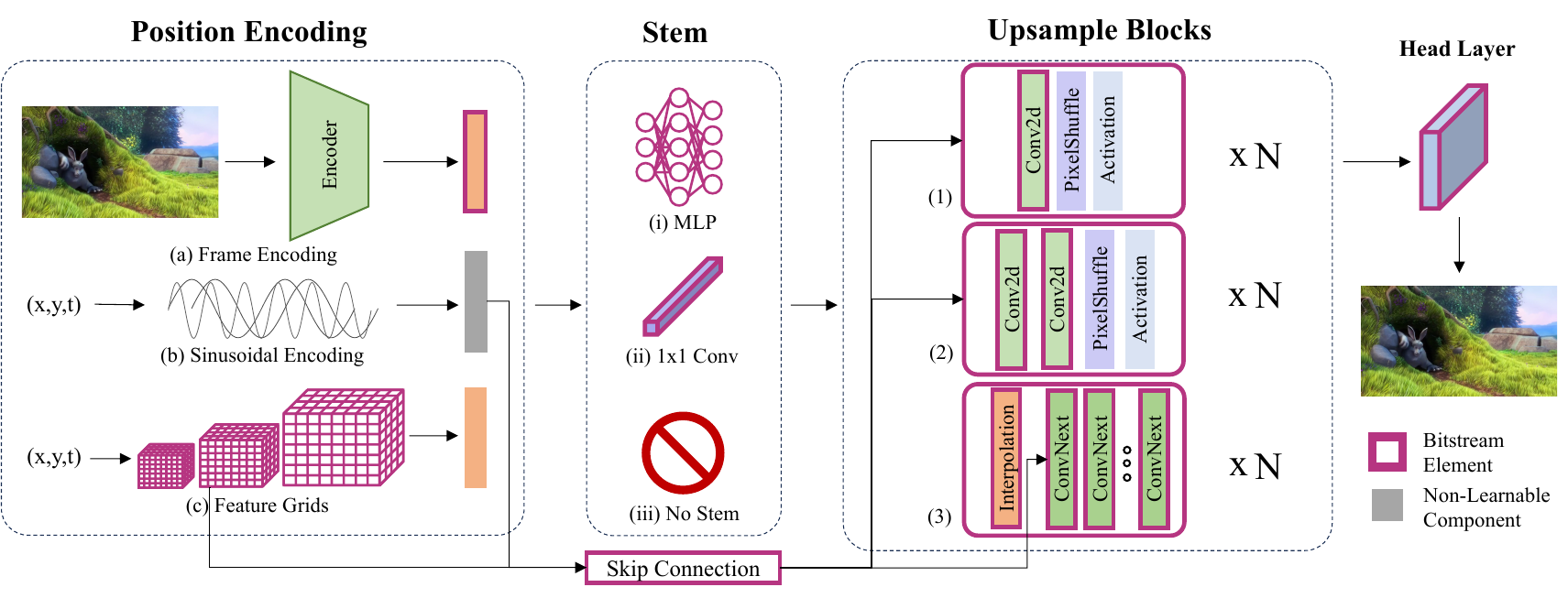}
   \caption{\textbf{Disentangling components within the NeRV family}. We isolate the various components of NeRV-like architectures and categorize critical parts of the design space -- positional encoding, stem, and upsample blocks. There are other components shown in the figure (skip connections) as well as many not shown, including critical choices like parameter distribution. In this work we analyze the impact of these various components and propose configurations based on different target optimizations of the time-quality-size trade-off.}
   \label{fig:disentangling_nervs_method}
\end{figure*}

\noindent\textbf{Video Compression.}
Since the rise of deep learning, many approaches have tried to supplant or enhance established video compression methods~\cite{mpeg,H264,hevc}.
Some focus on improving components of standard compression pipelines using deep learning~\cite{Rippel_2019_ICCV,liu2019neural,Agustsson_2020_CVPR,khani2021efficient,rippel2021elfvc}.
DCVC advocates for conditional coding instead of traditional predictive coding~\cite{li2021deepcontextualvideocompression}, and many works in the family of ``Neural Video Compression'' (NVC) follow this paradigm~\cite{Li_2022,Li_2023_CVPR,Li_2024_CVPR}.
In our work, instead of traditional pipelines or NVC, we focus on the potential of implicit neural representation for video compression.

\noindent\textbf{Implicit Neural Representation.}
With implicit neural representation (INR), a neural network is trained to map a set of coordinates to some signal, such as an image, video, or 3D representation~\cite{sitzmann2020implicit,tancik2020fourfeat,M_ller_2022,mildenhall2020nerf,chen2021nerv,xu2022signal,saragadam2023wire}.
An example INR could learn a multilayer perceptron to map $(x, y)$ positions (often with some positional encoding) to $(r, g, b)$ tuples, such that an image can be learned and stored as neural network weights.
These networks can be designed to be much smaller than the images they represent, such that one major application of INR has been image compression~\cite{dupont2021coin,dupont2022coin++,strümpler2022implicitneuralrepresentationsimage,ladune2023coolchiccoordinatebasedlowcomplexity}.
Other works focus more directly on video compression~\cite{chen2021nerv,zhang2021implicitneuralvideocompression,kim2022scalable,maiya2023nirvana,kwan2024nvrcneuralvideorepresentation}, and some are designed with both in mind~\cite{kim2023c3highperformancelowcomplexityneural}.
In this work,, we focus on video compression, and even more specifically we study the popular NeRV family of models~\cite{li2022enervexpediteneuralvideo,Zhang_2024_CVPR,Yan_2024_CVPR,Kim_2024,kwan2023hinerv,Lee_2023}, pioneered by Chen et al.~\cite{chen2021nerv}.
It uses convolution layers, typically upsampling from small feature maps via PixelShuffle~\cite{shi2016realtimesingleimagevideo}, in addition to MLPs and outputs all RGB values of a frame given the positional embedding of frame index $t$ as input.
More recent methods have abandoned the fixed positional encoding; some, such as HNeRV~\cite{chen2023hnerv}, use an image encoder to learn tiny content-based embeddings~\cite{chen2022cnerv,He_2023_CVPR,Zhao_2023_CVPR,Saethre_2024_CVPR,Zhao_2024_CVPR,xu2024vqnervvectorquantizedneural,wu2024qs}.
Others, such as FFNeRV~\cite{Lee_2023} and HiNeRV~\cite{kwan2023hinerv}, learn grid features (and such approaches are popular even outside the NeRV family~\cite{kim2022scalable,maiya2023nirvana,girish2023shacirascalablehashgridcompression}).

\noindent\textbf{Hyper-Networks.}
Some approaches leverage meta-learning, training a single hyper-network to avoid long per-sample training times associated with INR.
Given an input image or video, the hyper-network predicts INR network weights (``hypo-network'') that can reconstruct the input~\cite{chen2022transformers,kim2022generalizable}.
Some works predict INRs to generate novel images~\cite{skorokhodov2021adversarialgenerationcontinuousimages,Haydarov_2024_CVPR} or videos~\cite{yu2022generatingvideosdynamicsawareimplicit}.
Latent-INR proposes per-sample hyper-networks for video compression, and even aligns these representations with CLIP embeddings for semantic-aware INR~\cite{maiya2024latent}.
NeRV-Enc/NeRV-Dec~\cite{chen2024fastencodingdecodingimplicit} is the first work to apply non-implicit hyper-networks for video compression, sacrificing image quality and bitstream length for much faster encoding times.
While NeRV-Enc shows promising results for compressing a video as a collection of 8 evenly sampled frames, we are the first work to use hyper-networks to compress entire videos.
We also propose novel Weight Token Masking and slightly increasing hyper-network size for better overall performance.

\section{Method}
\label{sec:method}

We first explain how we disentangle the components of NeRV models in Section~\ref{subsec:disentangle_method}.
In Section~\ref{subsubsec:existing_methods} we describe the individual contributions of popular NeRV-like methods, and in Section~\ref{subsubsec:our_design_summary} we explain how we decompose these methods for further study and improvement.
As proof that such an investigation is useful, we use it to formulate our proposed RNeRV.
We describe our novel Weight Token Masking for improving the size-quality performance of hyper-network compression strategies in Section~\ref{subsec:weight_token_masking_method}.

\begin{figure*}[t]
  \centering
   \includegraphics[width=0.9\linewidth]{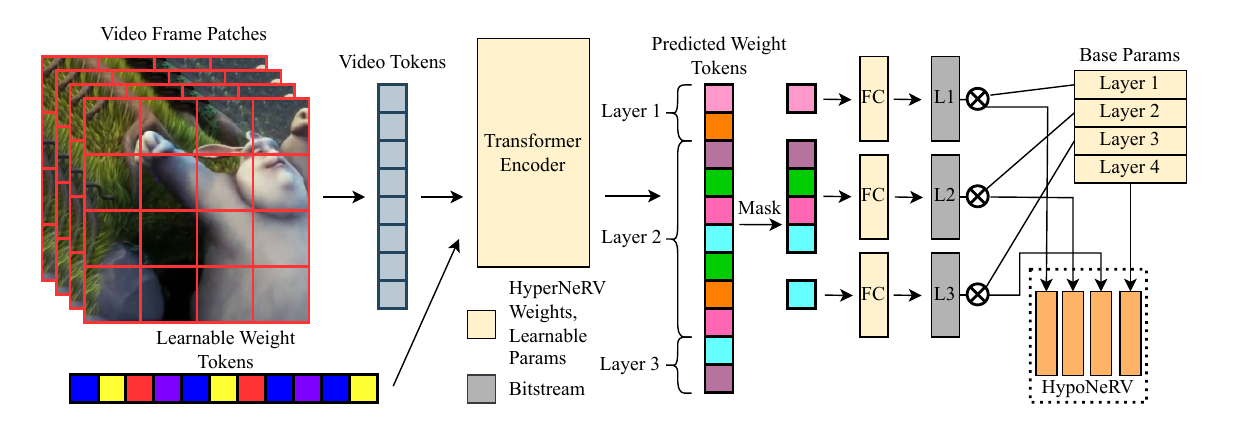}
   \caption{\textbf{Weight Token Masking.} We make hyper-networks have flexible encoding size at inference time. If we mask some of the predicted weight tokens for some portion of the samples (in our case, 50\%), we can also choose to mask the same set of tokens at inference for any given input clip (group of frames), with a small decrease in reconstruction quality. That is, we can flexibly reduce the bitrate by 2x post-training, during encoding, by choosing whether to mask some tokens for each group of frames.}
   \label{fig:hypernerv_masking_method}
\end{figure*}

\subsection{Disentangling Video INRs}
\label{subsec:disentangle_method}

\subsubsection{Existing Methods}
\label{subsubsec:existing_methods}

In this work, we primarily focus on the NeRV family of INR models, pioneered by the NeRV paper itself~\cite{chen2021nerv}.
The original NeRV trains a neural network, $\theta$ to predict a frame at some timestep, $I_t$, that is $I_t = \theta(t)$, using a reconstruction loss to minimize the difference between $I_t$ and $\theta(t)$.
NeRV starts by converting the frame index, $t$, to a sinusoidal positional encoding.
An MLP expands the dimension of the encoding, which is then reshaped to a matrix of shape $fc_w \times fc_h \times fc_{\text{dim}}$.
From there, NeRV gradually upsamples, with each layer first using a convolutional layer to expand the channel dimension (which is $fc_{\text{dim}}$ for the first layer), followed by a PixelShuffle to convert the additional channel dimensions to larger spatial size, then an activation.
Once the output has reached the size of the original frame, NeRV applies a final convolutional layer (called the ``head layer'') to convert the channel dimension to color channels.

E-NeRV~\cite{li2022enervexpediteneuralvideo} expands the stem of NeRV to disentangle and process an $xy$ embedding in addition to $t$. 
It also adds learnable skip connections from $t$ to every convolutional layer, and proposes a reworked convolutional layer for the first NeRV block (after the stem).
HNeRV~\cite{chen2023hnerv} principally proposes replacing the fixed, non-learnable positional encoding with a tiny, trainable ConvNext encoding which outputs content-adaptive embeddings.
It also proposes improvements for non-hybrid NeRV, by changing the MLP stem of NeRV to a simpler, larger stem consisting of a single fully-connected layer.
HNeRV highlights importance of proper parameter distribution and proposes adjusting kernel sizes and channel dimensions to ensure the later layers in the network have a good number of parameters.
Instead of a hybrid encoder, FFNeRV~\cite{Lee_2023} proposes a learnable grid for the positional encoding, eliminates the stem and shifts its responsibilities to the first NeRV block, and proposes a NeRV block that starts with an additional group-wise convolution.
It also proposes a flow-warping strategy for modeling motion in videos.
HiNeRV~\cite{kwan2023hinerv} introduces a broad array of improvements -- grid-based stem, grid-based skip connection, patchwise learning and decoding, and bilinear interpolation for upsampling with ConvNext to process features in the NeRV blocks.

\subsubsection{Our Definition of the Video INR Design Space}
\label{subsubsec:our_design_summary}

In Figure~\ref{fig:disentangling_nervs_method}, we provide an illustration of this design space.
Video INRs in the NeRV family are a combination of a position encoding, some stem to process that encoding, and then upsample blocks to transform that encoding to the resolution of the video.
Some leverage skip connections to improve performance.
Distribution of parameters, while not represented in the figure, is crucial for maintaining a healthy balance between speed and performance.
Since our work focuses on exploring this design space, we explain the possible options in the paragraphs below.

\noindent\textbf{Positional Encoding and Stem.}
Stems are typically designed with the position encoding in mind, so in this work we typically keep them paired for our explorations.
We can use sinusoidal positional encoding of $t$ with an MLP-style stem like NeRV and E-NeRV, or a single layer stem like NeRV.
We can use positional encoding of $x$ and $y$ with a transformer-based stem like E-NeRV.
We can use a grid encoding with a single layer stem like HiNeRV, or without a stem like FFNeRV.
We can use a content-adaptive encoder on frames or pixel differences like HNeRV or Diff-NeRV~\cite{li2022enervexpediteneuralvideo}, although for the sake of simplicity and scope, we do not investigate along this axis in this work.

\begin{table*}
\centering
\caption{\textbf{Video regression for $\textbf{1080}{\times}\textbf{1920}$ UVG videos} for the set of established video INR compression architectures supported in our library. We report PSNR$\uparrow$/MS-SSIM$\uparrow$~\cite{wang2003multiscale} for configurations with 1.5M and 3M parameters, all based on training and evaluation with our own library, using settings faithful to the original papers when possible. We include learnable grids and decoders in the parameter counts, since these are part of the bitstream. We do not include the HNeRV and DiffNeRV encoders.}
\label{tab:main_reconstruction_results}
\renewcommand{\tabcolsep}{6pt}
\renewcommand{\arraystretch}{1.3}
\resizebox{1.\linewidth}{!}{
\begin{tabular}{@{}l c ccccccc@{}}
\toprule
& & \multicolumn{7}{c}{Video} \\
\cmidrule(l){3-9}
Method & \#Params & Beauty & Bosphorus & HoneyBee & Jockey & ShakeNDry & YachtRide & ReadySetGo   \\
\midrule
\rowcolor{lightgray!30}& 1.5M & 30.80/0.8458 & 29.78/0.8568 & 33.69/0.9554 & 26.86/0.8016 & 28.76/0.8480 & 25.40/0.7727 & 21.34/0.7012 \\
\rowcolor{lightgray!30}\multirow{-2}{*}{ONeRV~\cite{chen2021nerv}} & 3M & 31.75/0.8632 & 31.69/0.9015 & 37.14/0.9773 & 28.61/0.8357 & 31.56/0.9091 & 26.49/0.8123 & 23.04/0.7721 \\
\multirow{2}{*}{E-NeRV~\cite{li2022enervexpediteneuralvideo}} & 1.5M & 31.28/0.8554 & 30.80/0.8809 & 37.27/0.9790 & 26.44/0.7907 & 32.14/0.9181 & 25.89/0.7932 & 21.58/0.7097 \\
& 3M & 32.65/0.8789 & 33.31/0.9281 & 38.81/0.9835 & 29.02/0.8372 & 33.68/0.9339 & 27.52/0.8485 & 23.74/0.7887 \\
\rowcolor{lightgray!30}& 1.5M & 31.39/0.8568 & 31.01/0.8900 & 34.57/0.9656 & 27.55/0.8120 & 30.13/0.8877 & 25.94/0.7967 & 21.90/0.7188 \\
\rowcolor{lightgray!30}\multirow{-2}{*}{NeRV~\cite{chen2023hnerv}} & 3M & 32.26/0.8730 & 32.54/0.9191 & 36.47/0.9761 & 29.48/0.8473 & 31.83/0.9162 & 27.06/0.8362 & 23.36/0.7753 \\
\multirow{2}{*}{FFNeRV~\cite{Lee_2023}} & 1.5M & 31.64/0.8607 & 30.14/0.8612 & 35.15/0.9681 & 28.93/0.8422 & 30.83/0.8978 & 26.00/0.7935 & 22.44/0.7420 \\
& 3M & 32.81/0.8816 & 32.50/0.9109 & 37.55/0.9797 & 31.41/0.8859 & 32.63/0.9242 & 27.54/0.8418 & 24.81/0.8222 \\
\rowcolor{lightgray!30} & 1.5M & \textbf{33.57}/\textbf{0.8947} & \textbf{35.54}/\textbf{0.9611} & \textbf{39.21}/\textbf{0.9844} & \textbf{32.25}/\textbf{0.8978} & \textbf{34.11}/\textbf{0.9386} & \textbf{29.38}/\textbf{0.8883} & \textbf{27.73}/\textbf{0.8991} \\
\rowcolor{lightgray!30}\multirow{-2}{*}{HiNeRV~\cite{kwan2023hinerv}} & 3M & \textbf{33.71}/\textbf{0.8987} & \textbf{37.43}/\textbf{0.9739} & \textbf{39.39}/\textbf{0.9849} & \textbf{34.71}/\textbf{0.9300} & \textbf{35.40}/\textbf{0.9543} & \textbf{30.39}/\textbf{0.9083} & \textbf{30.31}/\textbf{0.9362} \\
\multirow{2}{*}{HNeRV~\cite{chen2023hnerv}} & 1.5M & 31.96/0.8674 & 31.70/0.9000 & 37.87/0.9805 & 29.90/0.8652 & 32.87/0.9254 & 26.25/0.8062 & 24.00/0.8075 \\
& 3M & 32.84/0.8819 & 33.55/0.9306 & 38.72/0.9832 & 31.34/0.8841 & 33.62/0.9329 & 27.90/0.8577 & 25.72/0.8524 \\
\rowcolor{lightgray!30} & 1.5M & 32.89/0.8794 & 32.57/0.9113 & 37.91/0.9793 & 31.01/0.8845 & 32.82/0.9222 & 27.53/0.8354 & 24.98/0.8309 \\
\rowcolor{lightgray!30}\multirow{-2}{*}{DiffNeRV~\cite{Zhao_2023_CVPR}} & 3M & 33.63/0.8903 & 34.33/0.9372 & 39.19/0.9829 & 32.41/0.9001 & 34.07/0.9342 & 28.68/0.8669 & 26.50/0.8656 \\
\bottomrule
\end{tabular}
}
\end{table*}

\noindent\textbf{Blocks.}
We can use a basic NeRV block, consisting of a convolutional layer, PixelShuffle, and activation.
Alternatively, we can use an additional convolutional layer for all (FFNeRV) or the first (E-NeRV) blocks.
HiNeRV uses a ConvNext-based block with bilinear upsampling.

\noindent\textbf{Skips.}
E-NeRV and HiNeRV propose two different skips.
E-NeRV passes the $t$ positional encoding through a learned FC for each NeRV block, and within the block performs a layer normalization on the NeRV features followed by a fusion with the $t$ skip features.
HiNeRV passes local grid features (learnable per block) to an FC layer, and adds the output to the result of the block's bilinear upsampling.

\noindent\textbf{Other.}
Parameters are equally expensive to store regardless of location, but they are not equally expensive computationally.
Later parameters process larger feature maps, and are therefore much heavier in terms of FLOPs and slower in terms of wall time.
We control these as in prior works, with an expansion term $\texttt{exp}$, a reduction term $r$, and a kernel size term $\texttt{ks}$.
For most works, $\texttt{ks}$ is set to 3.
The expansion term governs the relationship between the input channels $ch_{\text{in}}$ and output channels $ch_{\text{out}}$ for the first layer specifically: $ch_{\text{out}} = ch_{\text{in}} * \texttt{exp}$.
The reduction term governs the change in channels from one layer to the next: $ch_{n+1} = ch_n / r$.
HNeRV changes $\texttt{ks}$ from $3$ to $1$ in the first layer and $3$ to $5$ in all other layers, and changes $r$ from its default 2 to $r=1.2$ (which allocates more parameters for later layers).
Such changes are good for quality, but hurt speed.

\subsection{Flexible Size via Weight Token Masking}
\label{subsec:weight_token_masking_method}

We adopt the hyper-network approach described in NeRV-Enc, although we re-implement the method since at the time of this writing their code is not public.
In this setup, a hypo-network is designed and initialized as a set of learnable parameters, corresponding to the weights and biases of each layer of the final hypo-network prediction.
These are sometimes referred to as ``shared'' parameters, since they are optimized across the entire dataset of videos.
In order to predict a hypo-network that corresponds to an actual video, the hyper-network backbone predicts weight tokens to modulate each layer of the base hypo-network.
The modulation is a simple elementwise multiplication between hyper-network predictions and the learnable shared parameters.
We give a more detailed walkthrough in Section~\ref{sec:hypernerv_walkthrough} of the Appendix.

Naturally, this sort of network can produce encodings at a single bitlength, modifiable only by quantization and compression approaches like arithmetic coding.
To introduce some flexibility, and owing to the fact that the network is already amenable to repeating and expanding the weight token predictions, we sometimes mask the weight tokens during training, as shown in Figure~\ref{fig:hypernerv_masking_method}.
That is, for half the samples during training, we disregard the latter half of the weight tokens of each layer.
The model then learns to encode necessary information in the first half of the weight tokens for each layer, and some information for higher quality in the latter half.
Then, at inference time, one can use the full length for higher quality, or half the length for a 2x improvement in storage size.

\section{Experiments}
\label{sec:experiments}

\begin{figure}[t]
    \centering
    \begin{minipage}{1.0\linewidth}
    \centering
        \includegraphics[width=0.7\textwidth]{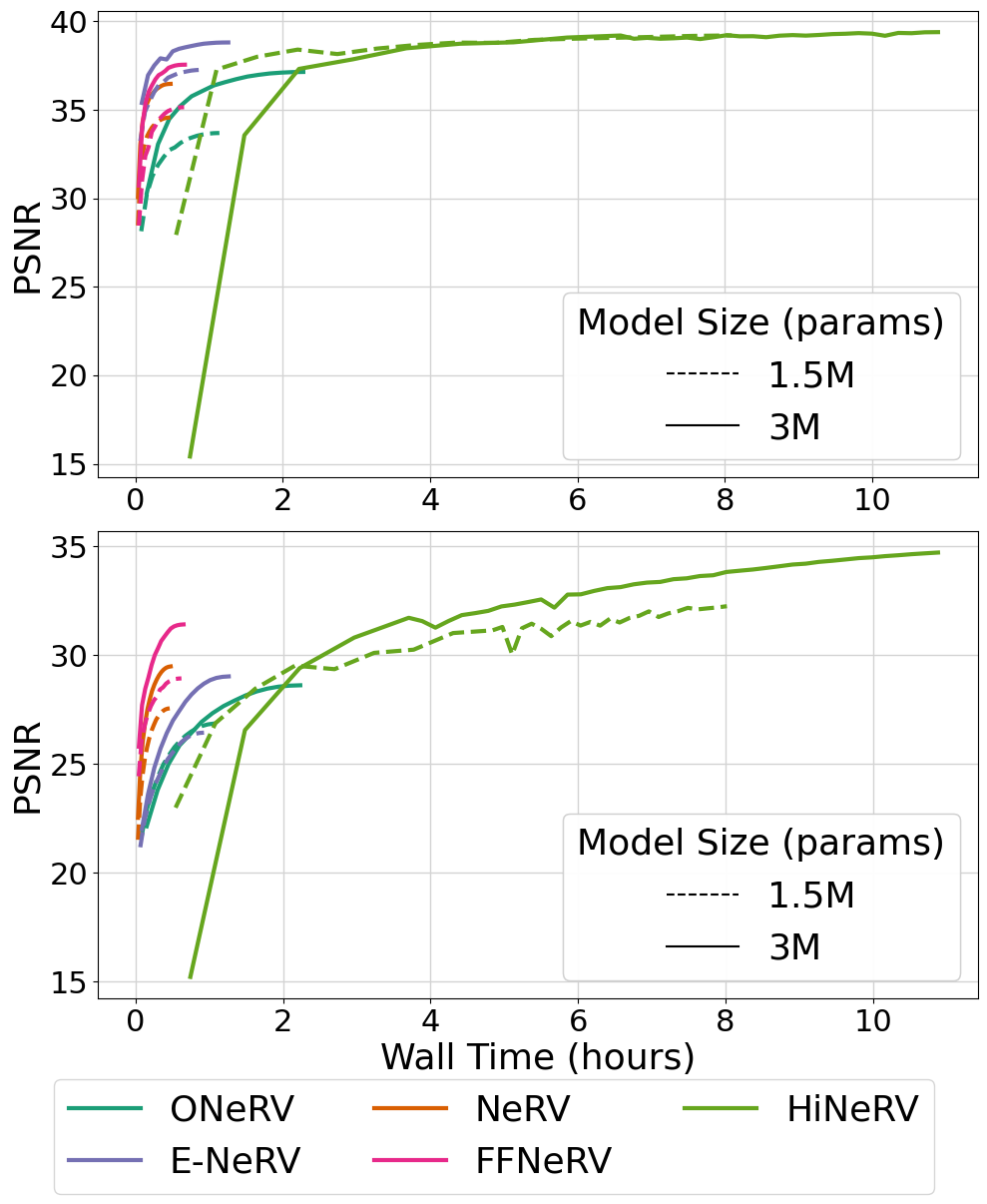}
        \caption{\textbf{Quality vs. wall time}, for UVG HoneyBee (top) and Jockey (bottom) at 1080p (Section~\ref{subsec:reproduce}).}
        \label{fig:quality_wall_time_main}
    \end{minipage}
\vspace{-1.0em}
\end{figure}

\begin{figure*}[t]
\begin{center}
    \begin{minipage}{0.32\linewidth}
    \includegraphics[width=1.0\textwidth]{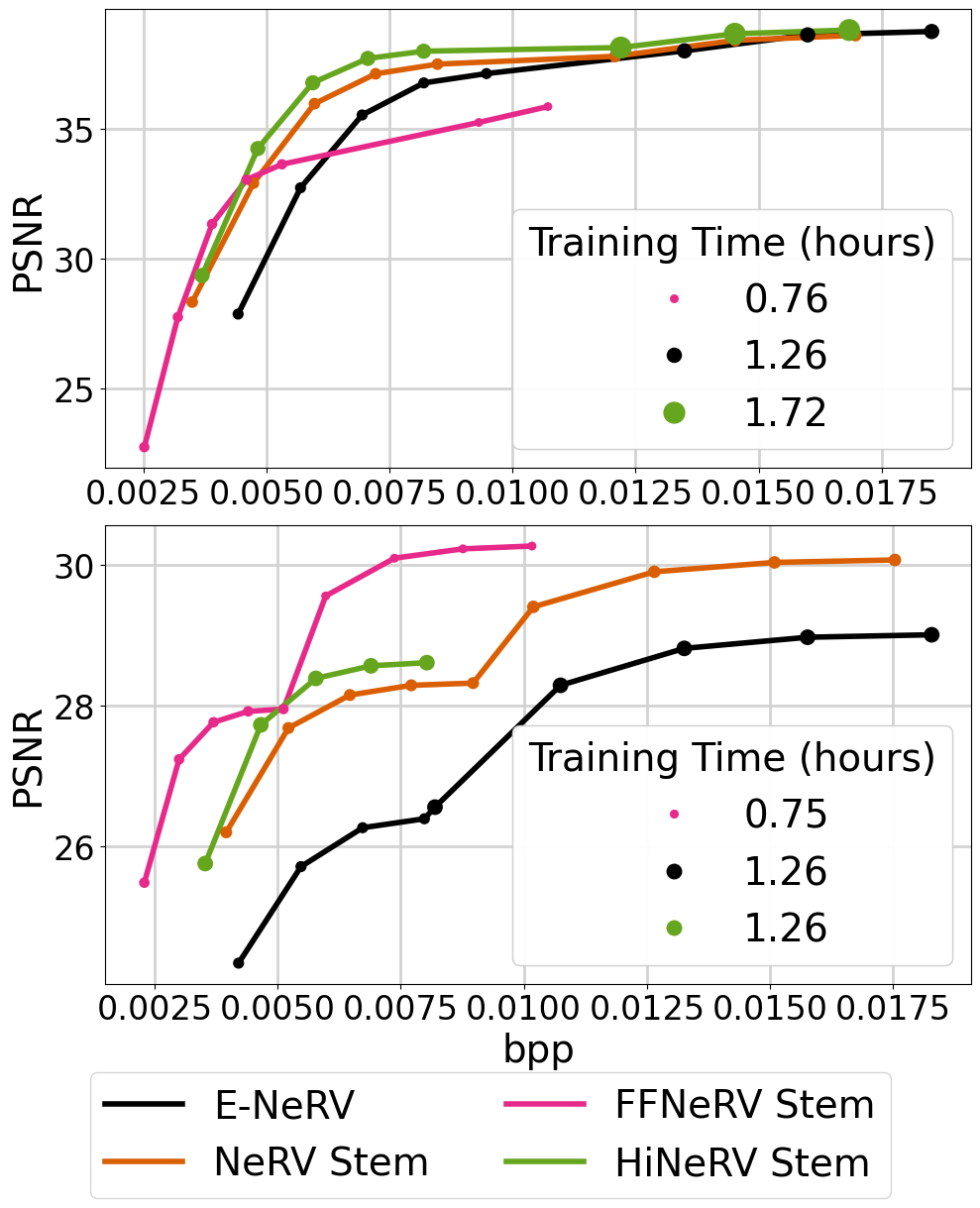}
        \caption{\textbf{Quality vs.\ size}, for HoneyBee (top) and Jockey (bottom) at 1080p for \textbf{E-NeRV} for various \textbf{position-stem} combinations (Section~\ref{subsec:investigating}).}
        \label{fig:enerv_stems_psnr_bpp}
    \end{minipage}
    \hfill
    \begin{minipage}{0.32\linewidth}
    \includegraphics[width=1.0\textwidth]{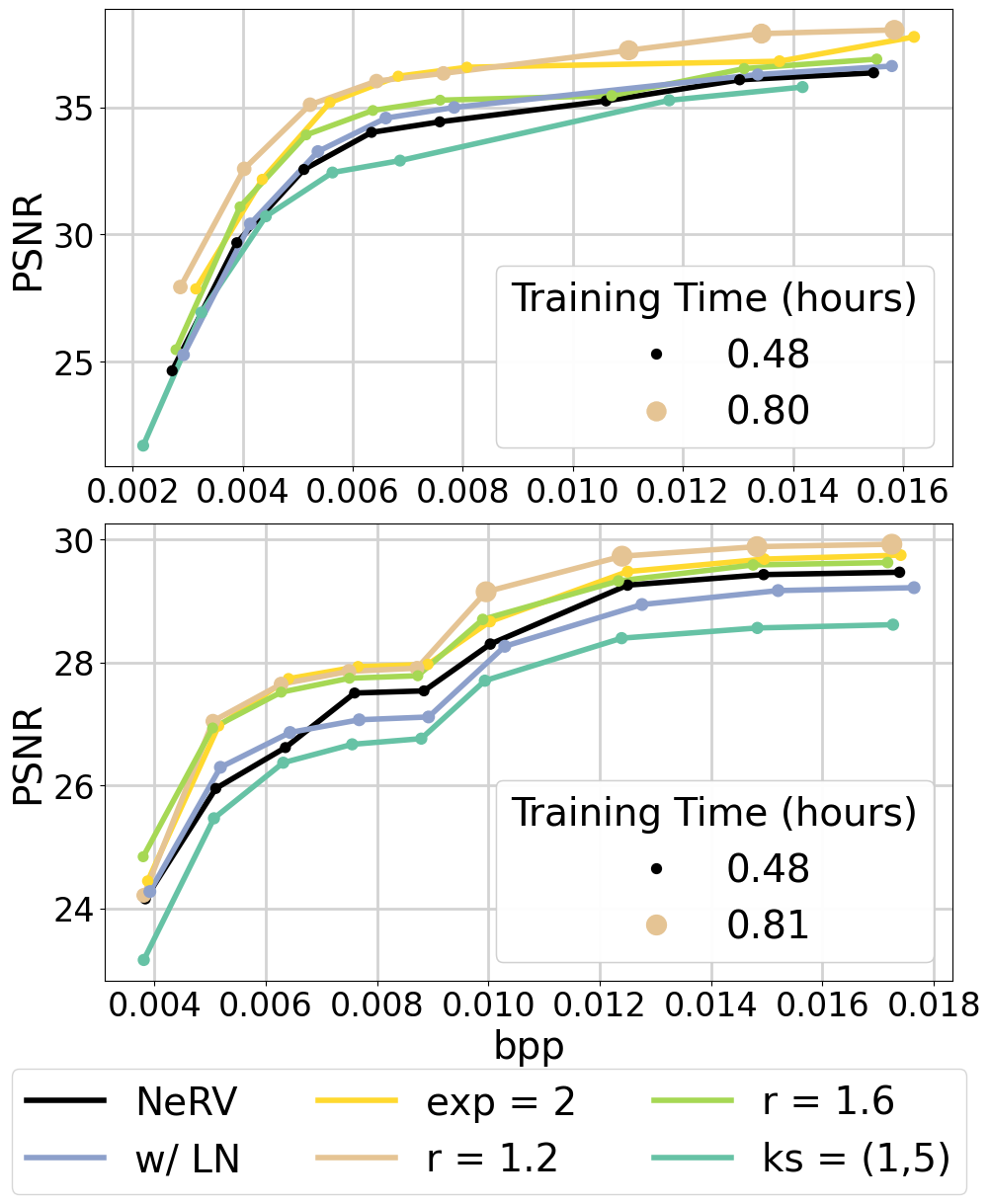}
        \caption{\textbf{Quality vs.\ size}, for HoneyBee (top) and Jockey (bottom) at 1080p for \textbf{NeRV} with different \textbf{parameter distributions} (Section~\ref{subsec:investigating}).}
        \label{fig:nerv_params_psnr_bpp}
    \end{minipage}
    \hfill
    \begin{minipage}{0.32\linewidth}
    \includegraphics[width=1.0\textwidth]{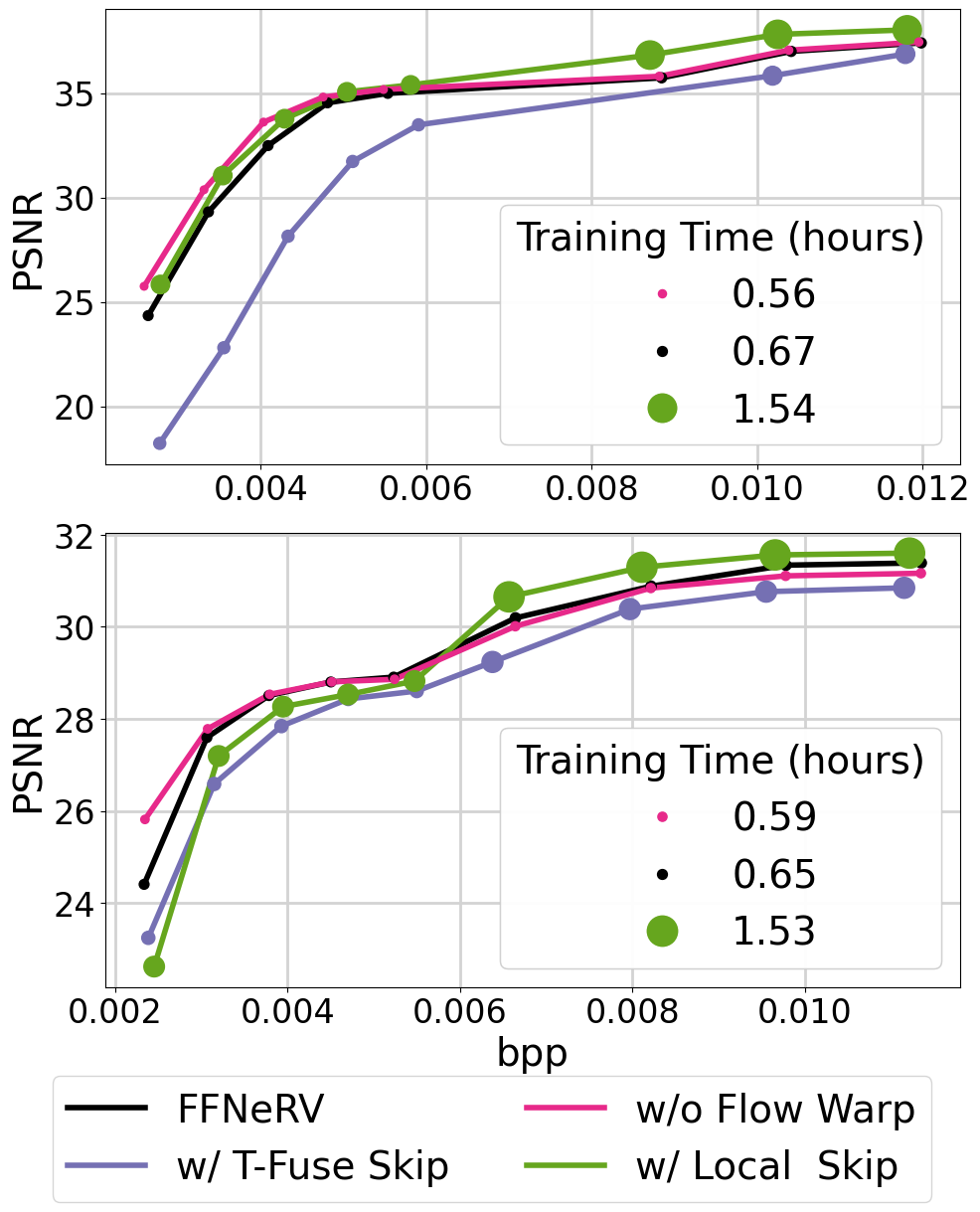}
        \caption{\textbf{Quality vs.\ size}, for HoneyBee (top) and Jockey (bottom) at 1080p for \textbf{FFNeRV} with different \textbf{skip} connections and without \textbf{flow} warping (Section~\ref{subsec:investigating}).}
        \label{fig:ffnerv_skips_flow_psnr_bpp}
    \end{minipage}
\vspace{-1.0em}
\end{center}
\end{figure*}

We run most of our experiments on the 7 videos from the UVG dataset~\cite{mercat2020uvg} listed in Table~\ref{tab:main_reconstruction_results}, always at 1080p ($1080{\times}1920$ resolution).
Except for ShakeNDry, which has 300 frames, all videos have 600 frames.
For our hyper-network experiments, we train on a 10,000 video subset of Kinetics-400~\cite{kay2017kineticshumanactionvideo} as in prior work~\cite{chen2024fastencodingdecodingimplicit}, and we test on a subset of UCF-101~\cite{soomro2012ucf101dataset101human} with 1 video per class.
For the hybrid methods HNeRV and DiffNeRV, we implement PixelShuffle with rectangular strides to allow operating with images at 16:9 aspect ratios, instead of cropping frames as in those papers.
We train all INR methods with the Adam optimizer~\cite{kingma2017adammethodstochasticoptimization}, with a cosine annealing learning rate and single epoch warmup.
We run on single GPUs, and wherever we benchmark wall-time, we use NVIDIA RTXA5000 GPUs.
Different GPUs could have different wall-times; however, we observe that the major trends in wall-time are consistent with FLOPs.
When we report bpp results, we use both the 1.5M and 3M parameter models, quantizing each to 8, 7, 6, 5, and 4 bits before performing arithmetic coding~\cite{mentzer2019practical}.
For more details, see Section~\ref{sec:implementation_details} in the Appendix.

We first verify our library's efficacy with by reproducing results for existing methods in Section~\ref{subsec:reproduce}.
We then investigate the effectiveness of disentangled INR components and suggest best principles for design in Section~\ref{subsec:investigating}.
We further benchmark the performance of existing models with short, medium, and long training times in Section~\ref{subsec:time}.
We formulate a state-of-the-art configuration, RNeRV, that performs especially well for short (2 minutes) and medium (30 minutes) encoding times.
For even faster encoding, we explore the use of hyper-networks and give results for our novel weight token masking strategy in Section~\ref{subsec:weight_token_masking_results}.
Finally, to better analyze what neurons from different NeRV methods learn in terms of both content and motion, we provide XINC~\cite{Padmanabhan_2024_CVPR} analysis using our extended version of their library that we implement in Section~\ref{sec:analysis}.

\subsection{Reproducing and Benchmarking Video INRs}
\label{subsec:reproduce}

\begin{figure*}[t]
\begin{center}
    \begin{minipage}{0.32\linewidth}
    \includegraphics[width=1.0\textwidth]{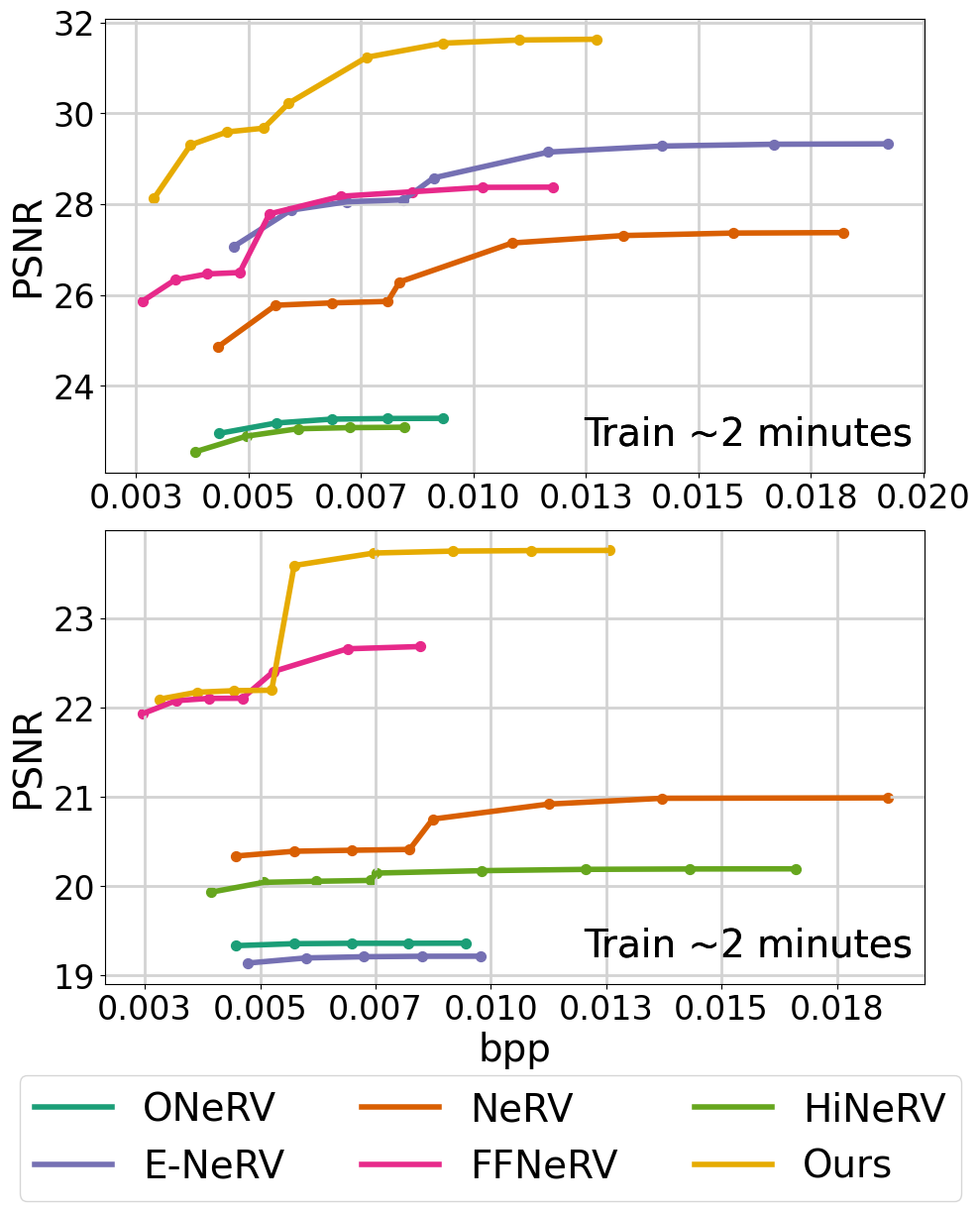}
        \caption{\textbf{Quality vs.\ size}, for HoneyBee (top) and Jockey (bottom) at 1080p with ``short'' training time (Section~\ref{subsec:time}).}
        \label{fig:short_time_psnr_bpp}
    \end{minipage}
    \hfill
    \begin{minipage}{0.32\linewidth}
    \includegraphics[width=1.0\textwidth]{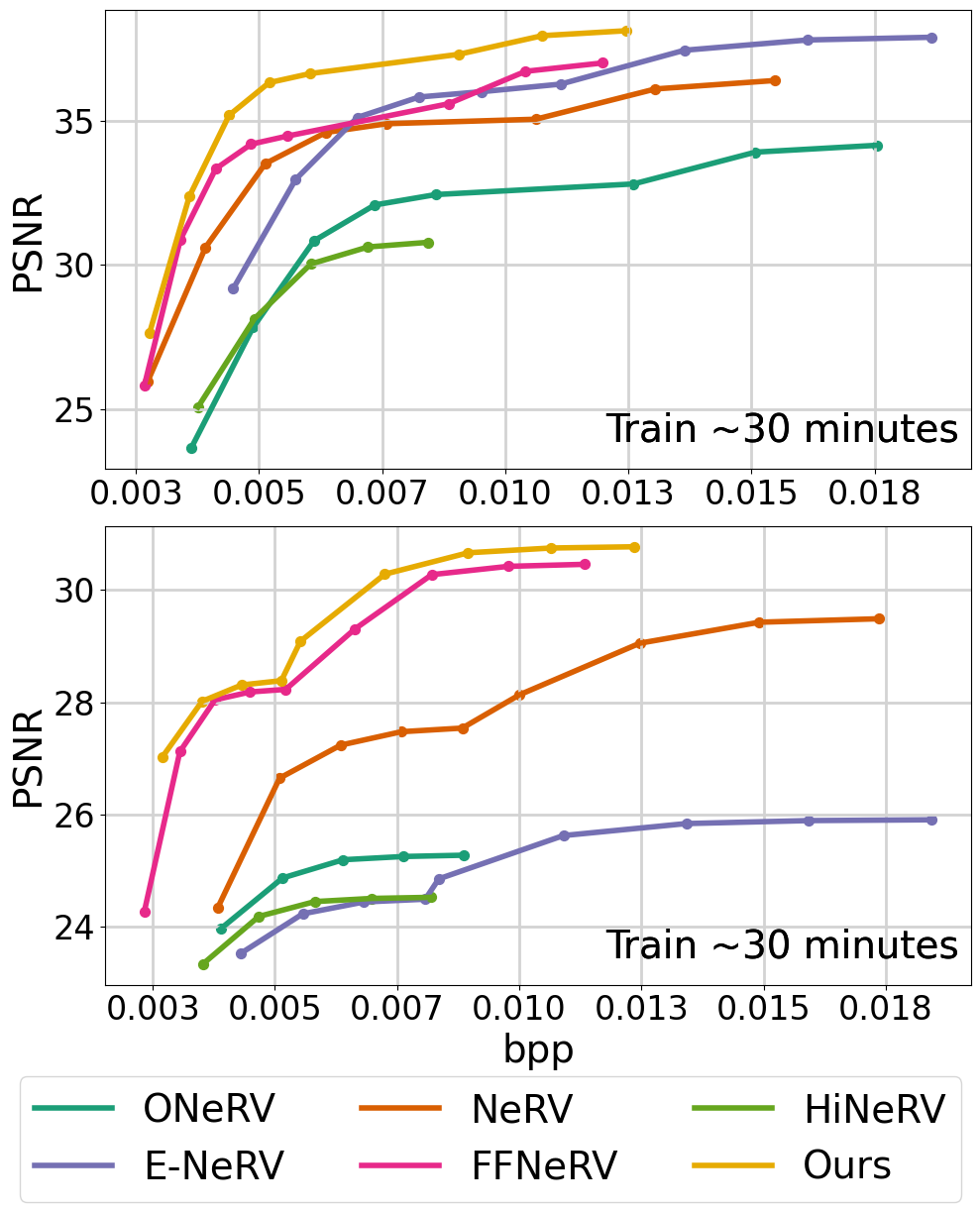}
        \caption{\textbf{Quality vs.\ size}, for HoneyBee (top) and Jockey (bottom) at 1080p with ``medium'' training time (Section~\ref{subsec:time}).}
        \label{fig:medium_time_psnr_bpp}
    \end{minipage}
    \hfill
    \begin{minipage}{0.32\linewidth}
    \includegraphics[width=1.0\textwidth]{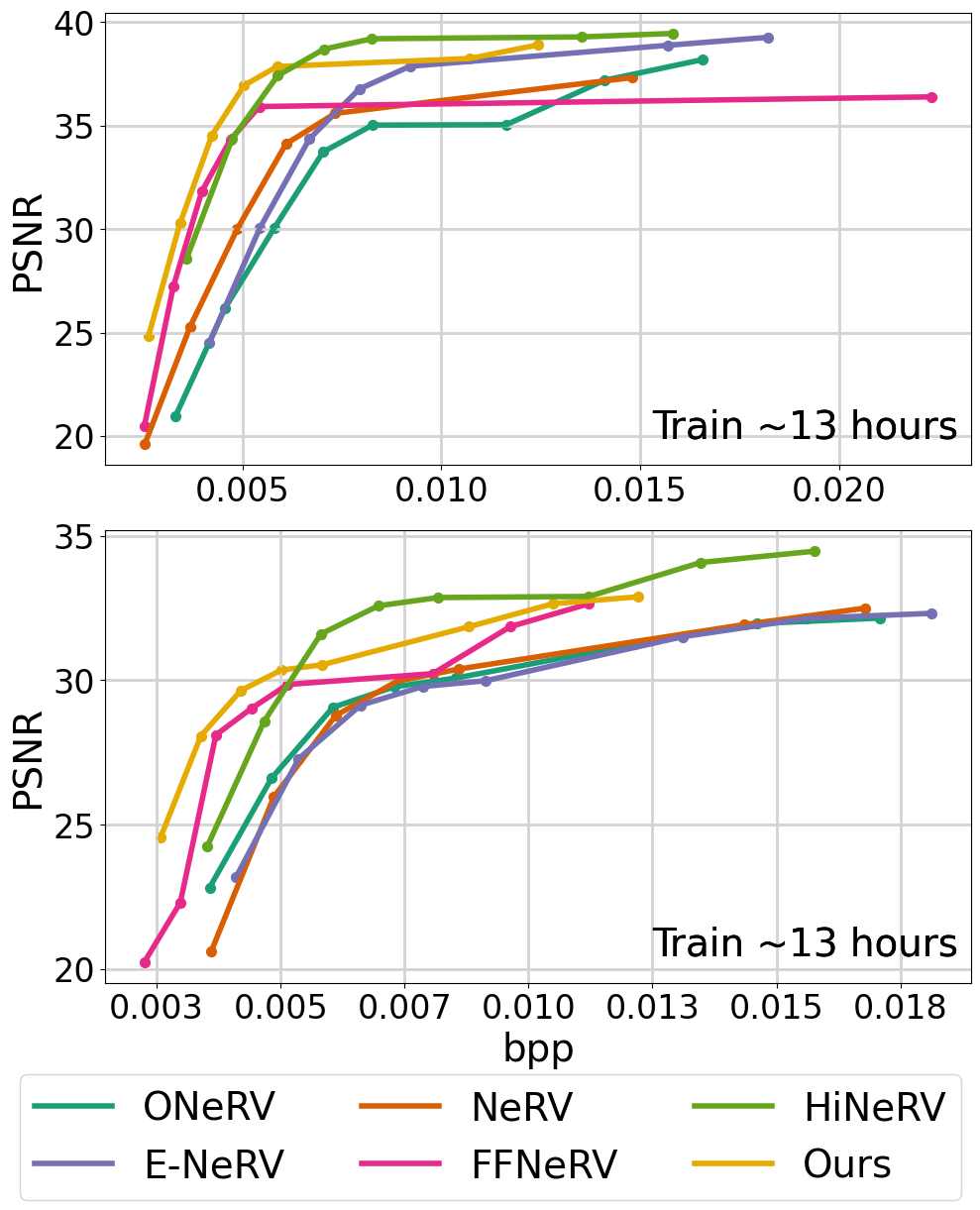}
        \caption{\textbf{Quality vs.\ size}, for HoneyBee (top) and Jockey (bottom) at 1080p with ``long'' training time (Section~\ref{subsec:time}).}
        \label{fig:long_time_psnr_bpp}
    \end{minipage}
    \vspace{-1.5em}
\end{center}
\end{figure*}

\begin{table}
\centering
\caption{\textbf{Weight Token Masking results.} We sacrifice some quality at full bit-width for superior quality at the more important half bit-width. We report bpp at 8-bit quantization.}
\label{tab:masking_results}
\resizebox{0.9\linewidth}{!}{
\begin{tabular}{@{}cc cc ccc@{}}
\toprule
\multicolumn{2}{c}{Masking?} & \multicolumn{2}{c}{\# Params} & \multicolumn{3}{c}{Metrics} \\
\cmidrule(l){1-2}
\cmidrule(l){3-4}
\cmidrule(l){5-7}
 Train & Val & Min & Max  & PSNR & MS-SSIM & bpp \\
\midrule
N & n/a         & 5.15k & 5.15k       & 23.22 & 0.6261 & 0.079 \\
$\checkmark$ & $\checkmark$           & 5.15k & 10.3k       & 23.71 & 0.6451 & 0.079 \\
$\checkmark$ & $\times$           & 5.15k & 10.3k       & 23.85 & 0.6492 & 0.157 \\
\midrule 
$\times$ & n/a         & 2.45k & 2.45k       & 22.55 & 0.6011 & 0.037 \\
$\checkmark$ & $\checkmark$           & 2.45k & 4.90k       & 22.81 & 0.6130 & 0.037 \\
$\checkmark$ & $\times$           & 2.45k & 4.90k       & 22.93 & 0.6163 & 0.075 \\
\bottomrule
\end{tabular}
}
\end{table}

\begin{table}
\centering
\caption{\textbf{Hypo-network size results.} We compare the 85.6k parameter hypo-network from NeRV-Enc with a larger one. We achieve better performance without increasing bitstream size.}
\label{tab:hypo_size_results}
\resizebox{0.85\linewidth}{!}{
\begin{tabular}{@{}cccccc@{}}
\toprule
 \multicolumn{2}{c}{\# Params} & \multicolumn{2}{c}{Quality} & \multicolumn{2}{c}{FPS} \\
\cmidrule(l){1-2}
\cmidrule(l){3-4}
\cmidrule(l){5-6}
Total & Unique & PSNR & MS-SSIM & Enc & Dec \\
\midrule
85.6k   & 24.1k & 25.29 & 0.7319 & \textbf{4455} & 632.7 \\
130k    & 24.1k & \textbf{25.92} & \textbf{0.7515} & 4206 & \textbf{717.1} \\
\bottomrule
\end{tabular}
}
\vspace{-1.0em}
\end{table}

We report reconstruction results for the set of video INR methods we reproduce in our library in Table~\ref{tab:main_reconstruction_results}.
We re-use published configurations when possible, otherwise we perform extensive hyperparameter searches to find best configurations for 1.5M and 3M parameters.
For example, we adjust the FFNeRV expansion from 8 to 4 to account for the smaller size compared to the 12M parameter models in their paper.
When training methods at equal epochs and equal decoder parameter counts, HiNeRV has the best results for every video.
In most cases, even the 1.5M parameter HiNeRV outperforms the competing methods at 3M parameters.

\begin{figure}[ht]
    \centering
    \begin{minipage}{1.0\linewidth}
        \centering
        \includegraphics[width=0.7\textwidth]{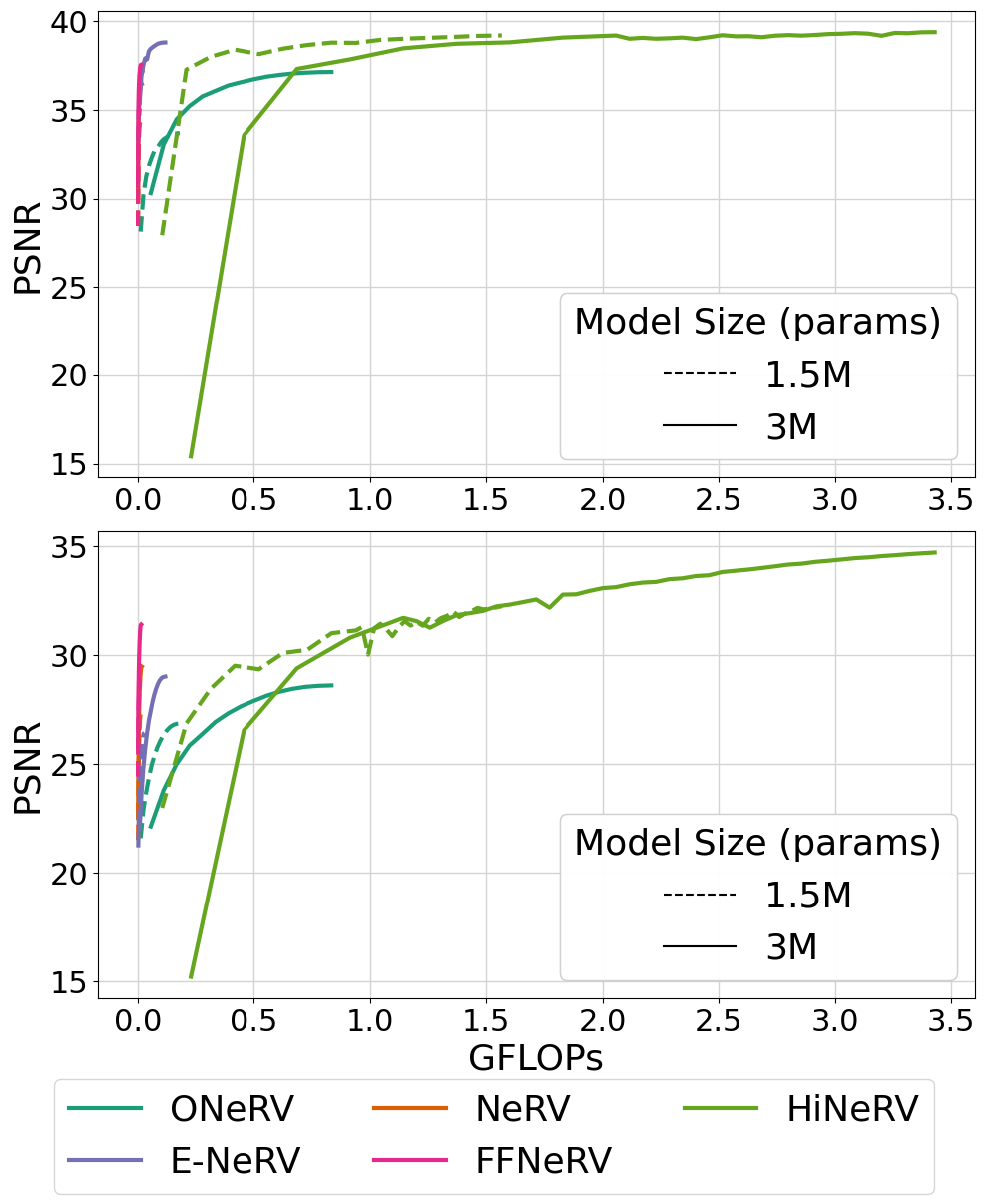}
        \caption{Reconstruction quality over GFLOPs, for UVG HoneyBee (top) and Jockey (bottom) at 1080p.}
        \label{fig:quality_flops_main}
    \end{minipage}
\end{figure}

However, one must also consider the training time. 
We provide quality-time results in terms of wall time in Figure~\ref{fig:quality_wall_time_main} (see Appendix for epochs in Figure~\ref{fig:quality_epochs_main} and FLOPs in Figure~\ref{fig:quality_flops_main}).
Although HiNeRV eventually achieves the best results, it takes significantly longer to do so than competing non-hybrid methods.
We also notice that while E-NeRV improves on the speed of the original NeRV (per the premise of the paper), FFNeRV is faster, while NeRV is the fastest.
This motivates two directions of investigation -- (1) which method is best with equal time, and (2) can we mix-and-match method components for good quality with less time?

\subsection{Investigating NeRV Design Decisions}
\label{subsec:investigating}

We show some key ablations here, with more in the Appendix.
We measure the impact of the position encoding/stem choice on E-NeRV in Figure~\ref{fig:enerv_stems_psnr_bpp}.
We look at parameter distribution for NeRV in Figure~\ref{fig:nerv_params_psnr_bpp} and the impact of adding skips and removing flow on FFNeRV in Figure~\ref{fig:ffnerv_skips_flow_psnr_bpp}.
We also consider blocks (see Appendix), but find that the FFNeRV, E-NeRV, and NeRV designs are roughly equivalent, while HiNeRV blocks are challenging to train without the other components of HiNeRV.

\begin{figure*}[t]
  \centering
   \includegraphics[width=.9\linewidth]{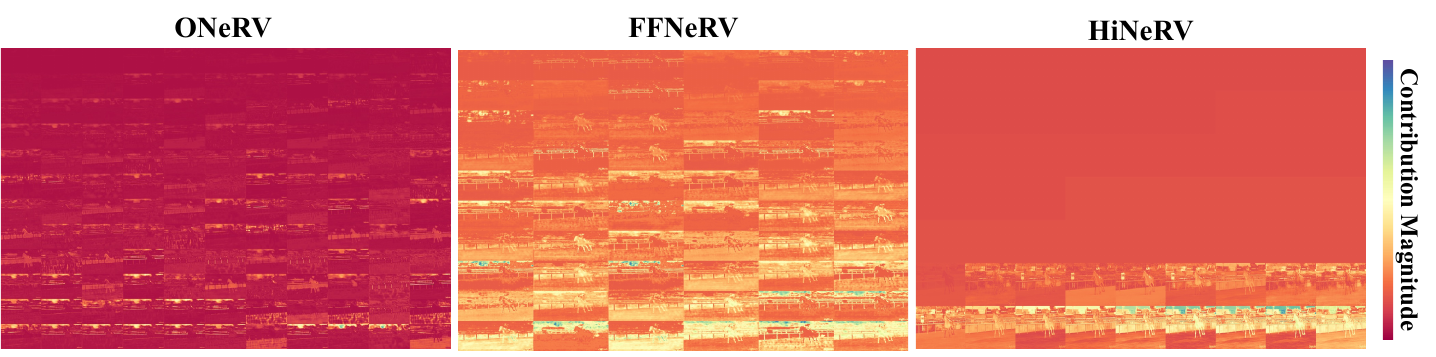}
   \caption{\textbf{XINC contribution maps} on 1080p Jockey. XINC dissects an INR to understand what parts of the visual signal are represented by each neuron (2D convolution kernel).
   We show contribution maps for the last (head) layer, sorted by magnitude for ease of comparison. 
   Darker red corresponds to lower contribution from that kernel for that spatial location, while blue/purple corresponds to high contribution. 
   ONeRV tends to show predominantly near-zero contributions, while most HiNeRV neurons lack discernible spatial correlations.
   }
   \label{fig:xinc_main}
    \vspace{-1.0em}
\end{figure*}

\begin{figure}[t]
  \centering
   \includegraphics[width=0.95\linewidth]{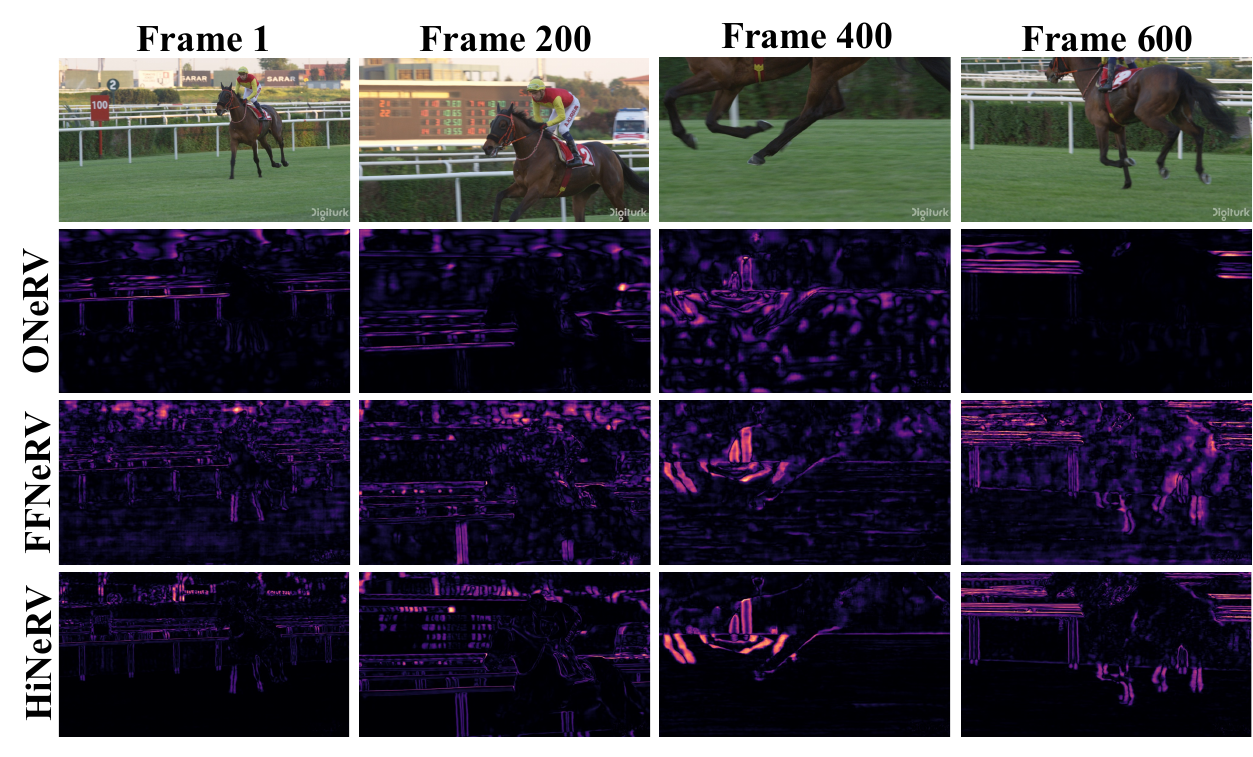}
    \caption{\textbf{XINC motion analysis} for the last (head) layer, for 1080p Jockey.
    We show fluctuation in total kernel contributions in response to motion between adjacent frames at various points in the video.
    While HiNeRV's changes are driven by fine differences between frames, FFNeRV shows more spatially distributed contribution changes and ONeRV exhibits much less structure.}
   \label{fig:xinc_motion}
    \vspace{-1.0em}
\end{figure}

Based on these ablations, we first observe the promise of the FFNeRV stem.
From Figure~\ref{fig:enerv_stems_psnr_bpp}, it achieves very strong results on the Jockey video.
In general, all other stems and positional encodings considered are better than the original E-NeRV stem and fixed positional encoding.
We also notice, from Figure~\ref{fig:nerv_params_psnr_bpp}, that distributing parameters to later layers is helpful for performance.
However, a key exception is doing this by changing the parameters, as in HNeRV, which seems to generally be unhelpful.
Additionally, redistributing the parameters can dramatically increase training times (with a nearly 2x gap between the original $r=2$ and $r=1.2$), and must therefore be done with care.
We observe that while the HiNeRV-style local skip helps FFNeRV in Figure~\ref{fig:ffnerv_skips_flow_psnr_bpp}, it is quite expensive.
We also find that omitting flow warping saves time, with minimal quality penalty.

\subsection{Optimizing NeRV Designs for Time}
\label{subsec:time}

Based on our findings in Section~\ref{subsec:investigating}, we propose a time-efficient alternative to existing models.
The first is a combination of the other methods -- we use the modified NeRV blocks from FFNeRV, the feature grids and stemless approach of FFNeRV, local skips and layer norms from E-NeRV, we change $r$ from 2 to 1.2 for the 1.5M model and 1.4 for the 3M parameter model (necessary to preserve the equal size) while reducing the first-layer expansion to 4 as well.
Notably we opt for the E-NeRV skip in spite of its struggles in Figure~\ref{fig:ffnerv_skips_flow_psnr_bpp}; we hypothesize it requires the layer normalization~\cite{ba2016layernormalization} to work well.

We show results for existing NeRV configurations compared to our proposed configuration, in terms of PSNR/bpp, for encoding time equivalent to 20 NeRV epochs (short) in Figure~\ref{fig:short_time_psnr_bpp}, 300 NeRV epochs (medium) in Figure~\ref{fig:medium_time_psnr_bpp}, and 300 HiNeRV epochs (long) in Figure~\ref{fig:long_time_psnr_bpp}.
As expected, HiNeRV struggles at short and medium training times, although it is notably sometimes better than E-NeRV and ONeRV.
We note that our configuration achieves the best results for both Jockey and HoneyBee at short and medium training times.
Additionally, our configuration achieves the best results at low bpp even for long training time.
Due to the strength and orthogonality of their individual contributions, when existing video INR methods are combined, they are worth even more than the sum of their parts.

\subsection{Hyper-network Improvements}
\label{subsec:weight_token_masking_results}

While we call 2 minutes ``short'' encoding time for the 600 frame, 1080p UVG videos, this is nowhere near the real-time encoding speeds that are desirable for many applications.
By contrast, hyper-network methods can perform encoding with a single forward pass.
However, hyper-networks like NeRV-Enc do not come close to the quality-size performance of other methods.
To help solve this, we introduce Weight Token Masking to enable hyper-networks to encode videos with more adaptive bitstream lengths.
We show results in Table~\ref{tab:masking_results} for our improved HyperNeRV with adaptive weight token masking, where masked tokens are not stored, as described in Section~\ref{sec:method}.
We achieve better PSNR and MS-SSIM than the baseline NeRV-Dec networks at equal bpp, and when we do not perform the masking on those same networks (doubling the size), we get slightly better quality.
While such small improvements would not justify the increase in size, this still signals a promising direction for exploration that we leave for future work.

We also examine the role of ``shared parameters'' (base hypo-network size) in Table~\ref{tab:hypo_size_results}.
We can increase the size of the hypo-network for significant improvements to MS-SSIM and PSNR at equal bpp.
Future work on hyper-networks ought to focus even more on the hypo-network design, optimizing the design of the base hypo-network to be as large as is useful (since it can be ``installed'' with the algorithm) while allowing the unique parameters to be as small as possible (since these are actually stored per-clip).

\subsection{Analysis}
\label{sec:analysis}

We finally perform some qualitative analysis to better understand the different NeRV methods. 
We extend XINC~\cite{Padmanabhan_2024_CVPR} to dissect and analyze representations not just from NeRV, but also for other NeRV variants in this paper, including hypo-NeRV (see the Appendix).
We show contribution maps for their head (last) layers in Figure~\ref{fig:xinc_main}.
We observe that ONeRV stands out compared to other methods with many kernels in the head layer giving low, near-zero contributions.
Notably HiNeRV has many maps without as much contrast as those from other methods.

In Figure~\ref{fig:xinc_motion} we show how the different networks handle motion, by taking the differences in total contributions (all maps summed) for adjacent frames.
Notably the best method, HiNeRV, seems to have very structured changes, focusing specifically on adapting contributions for the fine details that change between frames.
By contrast, FFNeRV, perhaps due to its flow-warping constraint, has contribution changes more spread out spatially.
In general the weaker method, ONeRV, noticeably has less structured changes due to the motion between frames.
For further analysis, including for hypo-networks, see the Appendix.

\section{Conclusion}
\label{sec:conclusion}

We have provided a new library which integrates NeRV-like methods to allow for the disentangling of their components, and the assembly of superior, time-optimal NeRVs.
We have distilled the principles necessary for the creation of such networks and proposed a highly effective novel configuration.
We proposed Hyper-network innovations for smaller hypo-networks with better quality, namely weight token masking and allocating additional base parameters for the hypo-network.
We extended XINC to analyze how these networks represent content and motion in videos.

\noindent\textbf{Acknowledgments.} This work was partially supported by NSF CAREER Award (\#2238769) and Dolby-UMD Joint Seed Grant. The authors would like to thank Fangjun Pu, Peng Yin, Birendra Kathariya, Tong Shao, and Guan-Ming Su for their feedback. The authors acknowledge UMD’s supercomputing resources made available for conducting this research. The U.S. Government is authorized to reproduce and distribute reprints for Governmental purposes notwithstanding any copyright annotation thereon. The views and conclusions contained herein are those of the authors and should not be interpreted as necessarily representing the official policies or endorsements, either expressed or implied, of NSF, Dolby, or the U.S. Government.

\clearpage
\setcounter{page}{1}
\maketitlesupplementary

\begin{figure*}[t]
\begin{center}
    \begin{minipage}{0.45\linewidth}
    \includegraphics[width=1.0\textwidth]{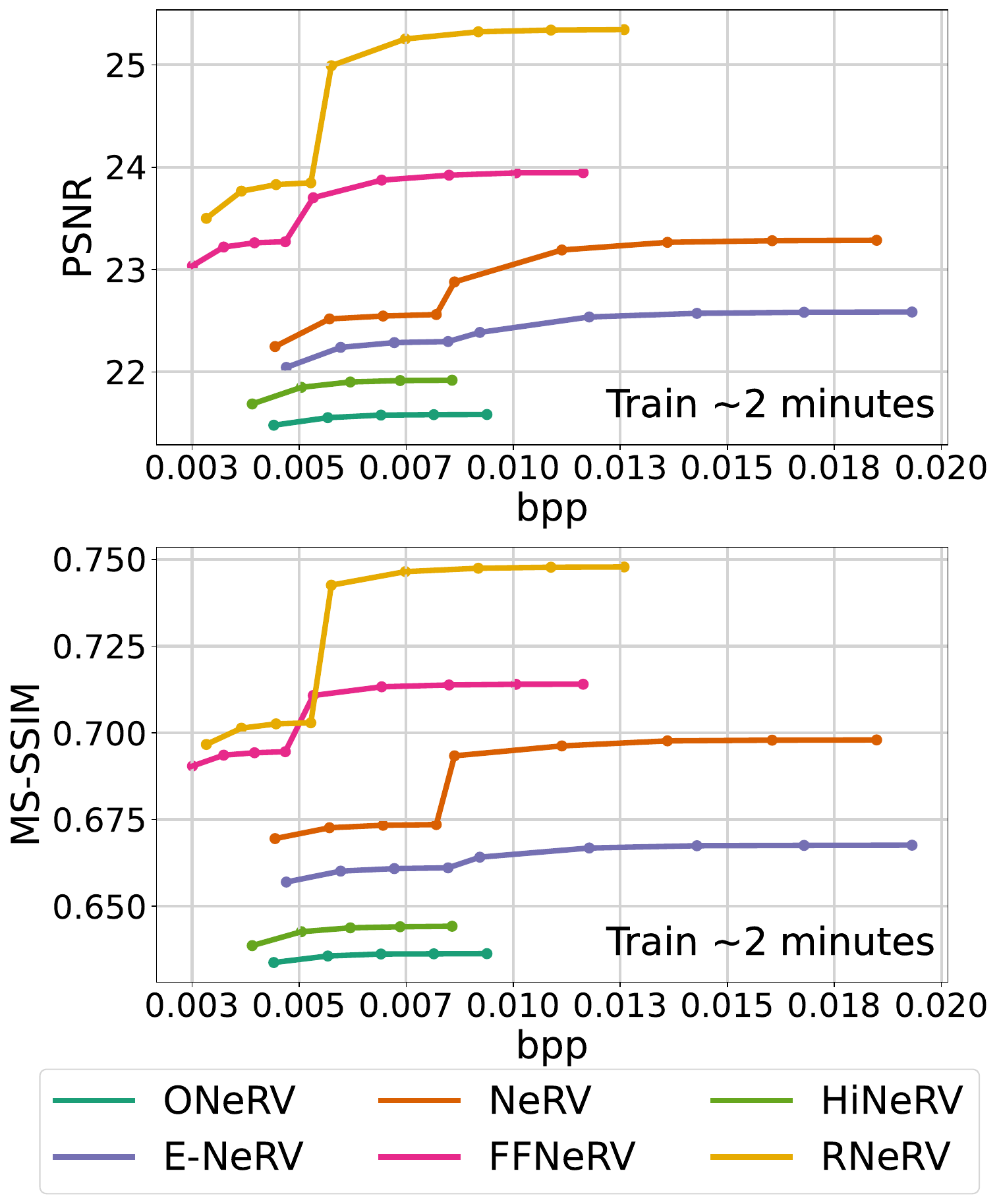}
        \caption{\textbf{Quality vs.\ size}, averaged across the 7 UVG videos we use in this paper, for \textbf{``short''} training time.}
        \label{fig:videos_average_short}
    \end{minipage}
    \hfill
    \begin{minipage}{0.45\linewidth}
    \includegraphics[width=1.0\textwidth]{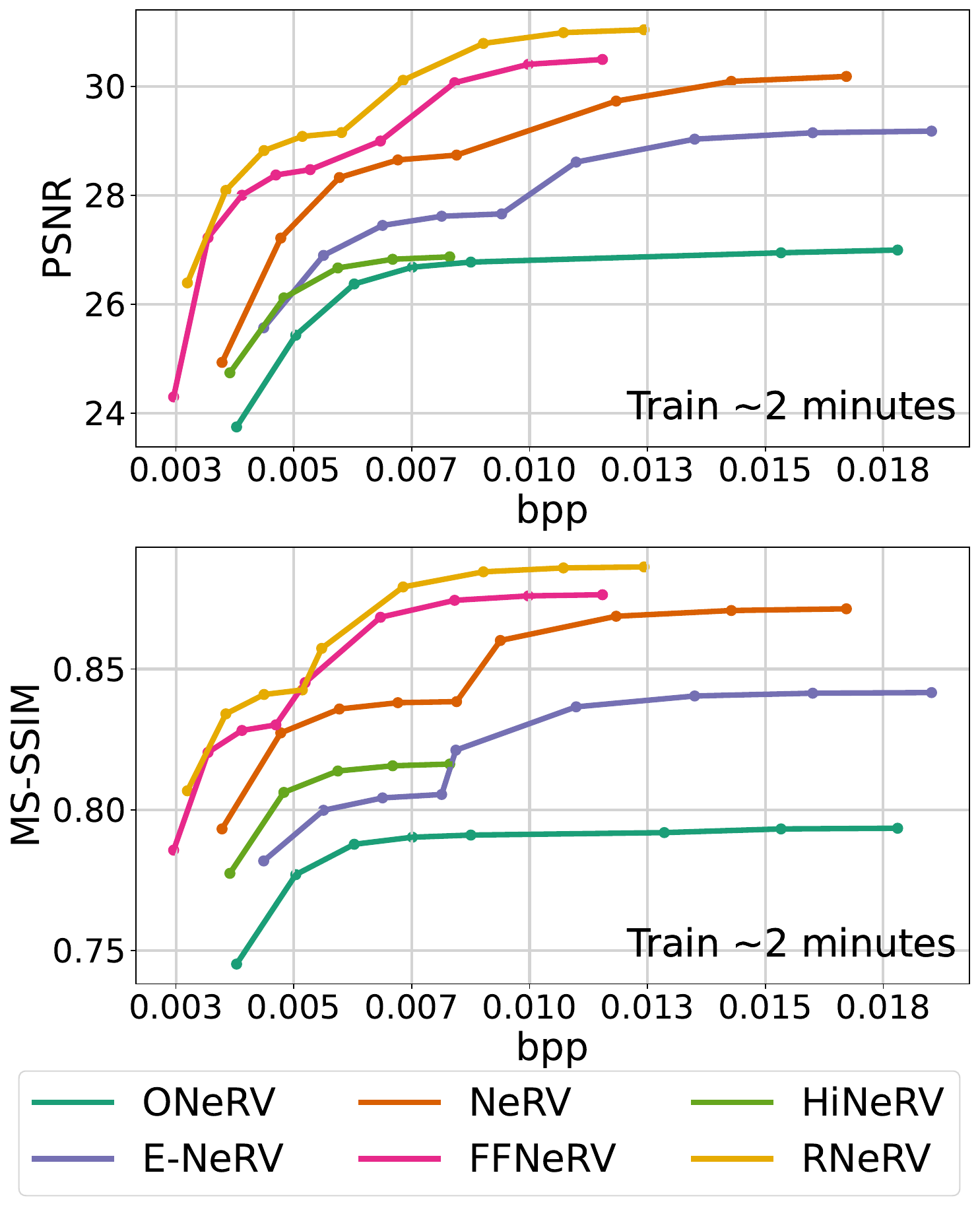}
        \caption{\textbf{Quality vs.\ size}, averaged across the 7 UVG videos we use in this paper, for \textbf{``medium''} training time.}
        \label{fig:videos_average_medium}
    \end{minipage}
\vspace{-1.0em}
\end{center}
\end{figure*}

\begin{figure*}[t]
\begin{center}
    \begin{minipage}{0.32\linewidth}
    \includegraphics[width=1.0\textwidth]{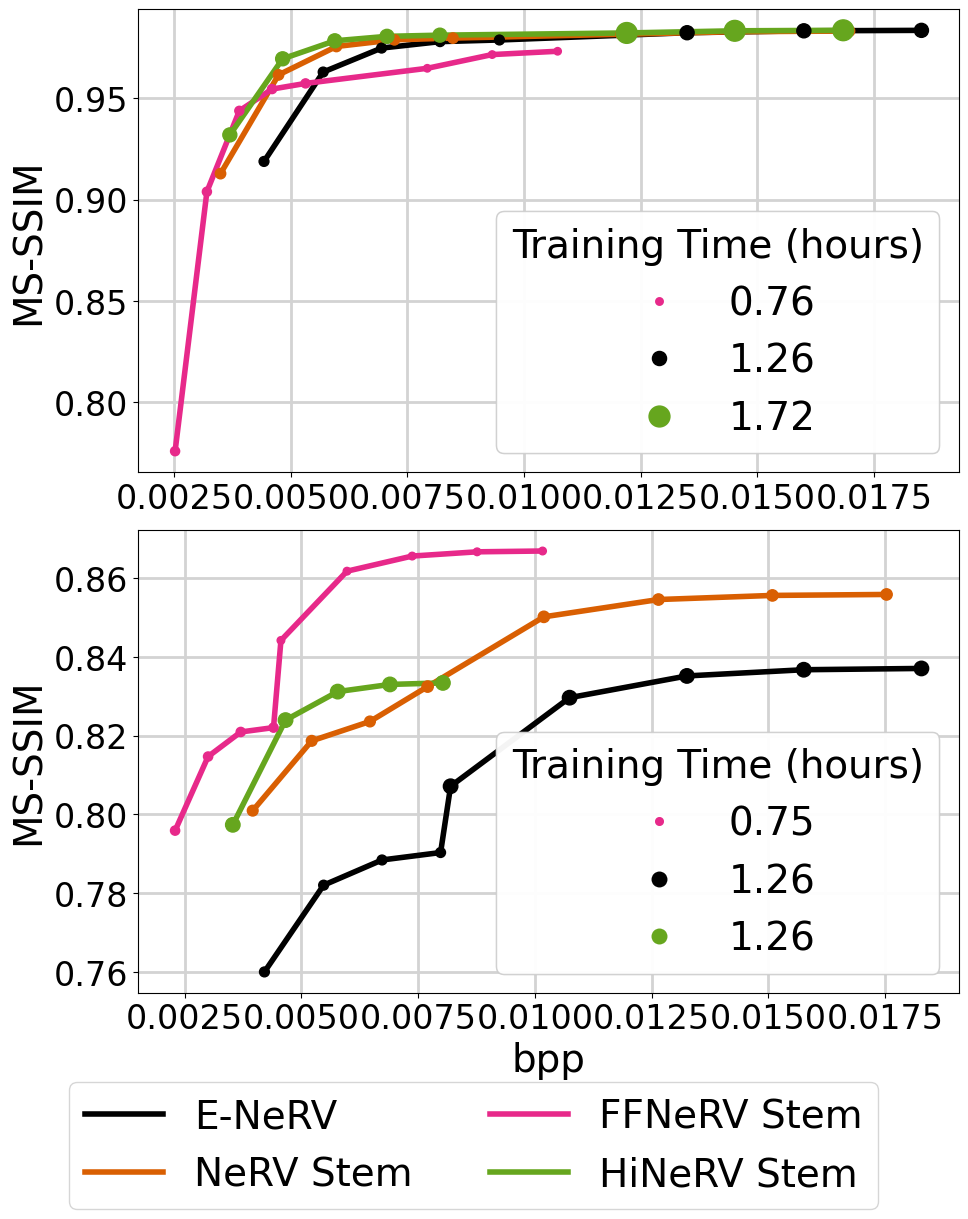}
        \caption{\textbf{Quality vs.\ size}, for HoneyBee (top) and Jockey (bottom) at 1080p for \textbf{E-NeRV} for various \textbf{position-stem} combinations.}
        \label{fig:enerv_stems_ssim_bpp}
    \end{minipage}
    \hfill
    \begin{minipage}{0.32\linewidth}
    \includegraphics[width=1.0\textwidth]{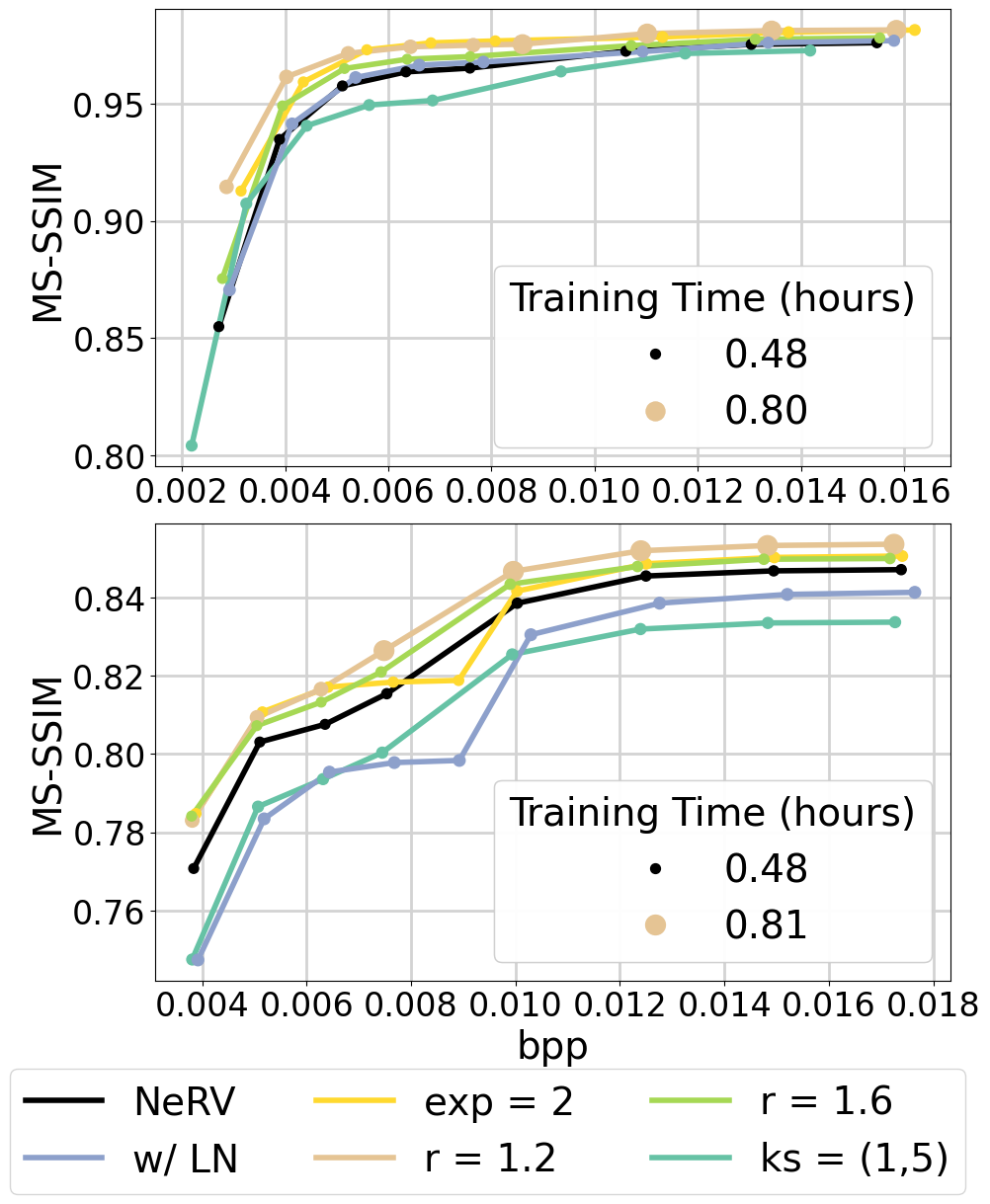}
        \caption{\textbf{Quality vs.\ size}, for HoneyBee (top) and Jockey (bottom) at 1080p for \textbf{NeRV} with different \textbf{parameter distributions}.}
        \label{fig:nerv_params_ssim_bpp}
    \end{minipage}
    \hfill
    \begin{minipage}{0.32\linewidth}
    \includegraphics[width=1.0\textwidth]{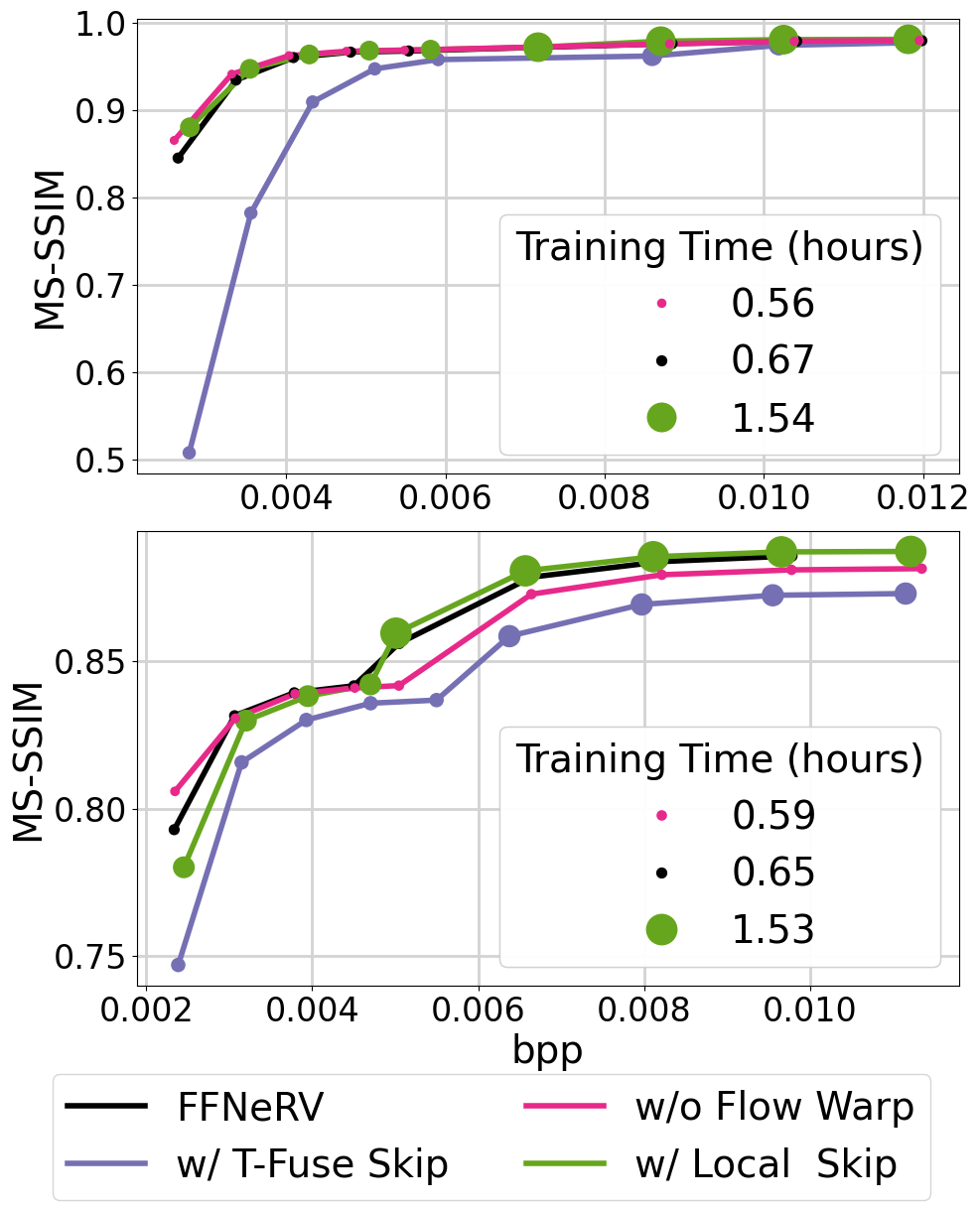}
        \caption{\textbf{Quality vs.\ size}, for HoneyBee (top) and Jockey (bottom) at 1080p for \textbf{FFNeRV} with different \textbf{skip} connections and without \textbf{flow} warping.}
        \label{fig:ffnerv_skips_flow_ssim_bpp}
    \end{minipage}
\vspace{-1.0em}
\end{center}
\end{figure*}

\begin{figure*}[t]
\begin{center}
    \begin{minipage}{0.32\linewidth}
    \includegraphics[width=1.0\textwidth]{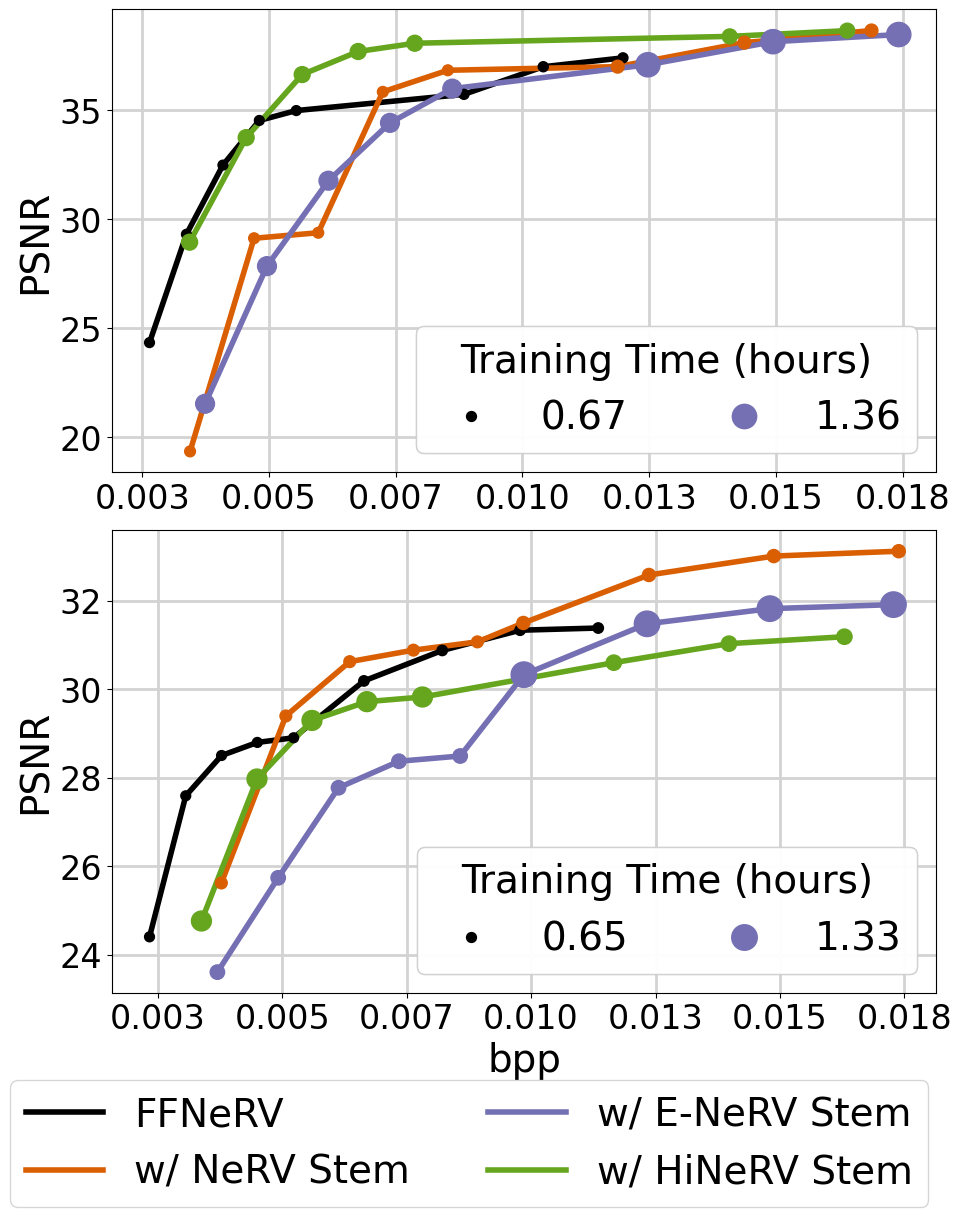}
        \caption{\textbf{Quality vs.\ size}, for HoneyBee (top) and Jockey (bottom) at 1080p for \textbf{FFNeRV} for various \textbf{position-stem} combinations.}
        \label{fig:ffnerv_stems_psnr_bpp}
    \end{minipage}
    \hfill
    \begin{minipage}{0.32\linewidth}
    \includegraphics[width=1.0\textwidth]{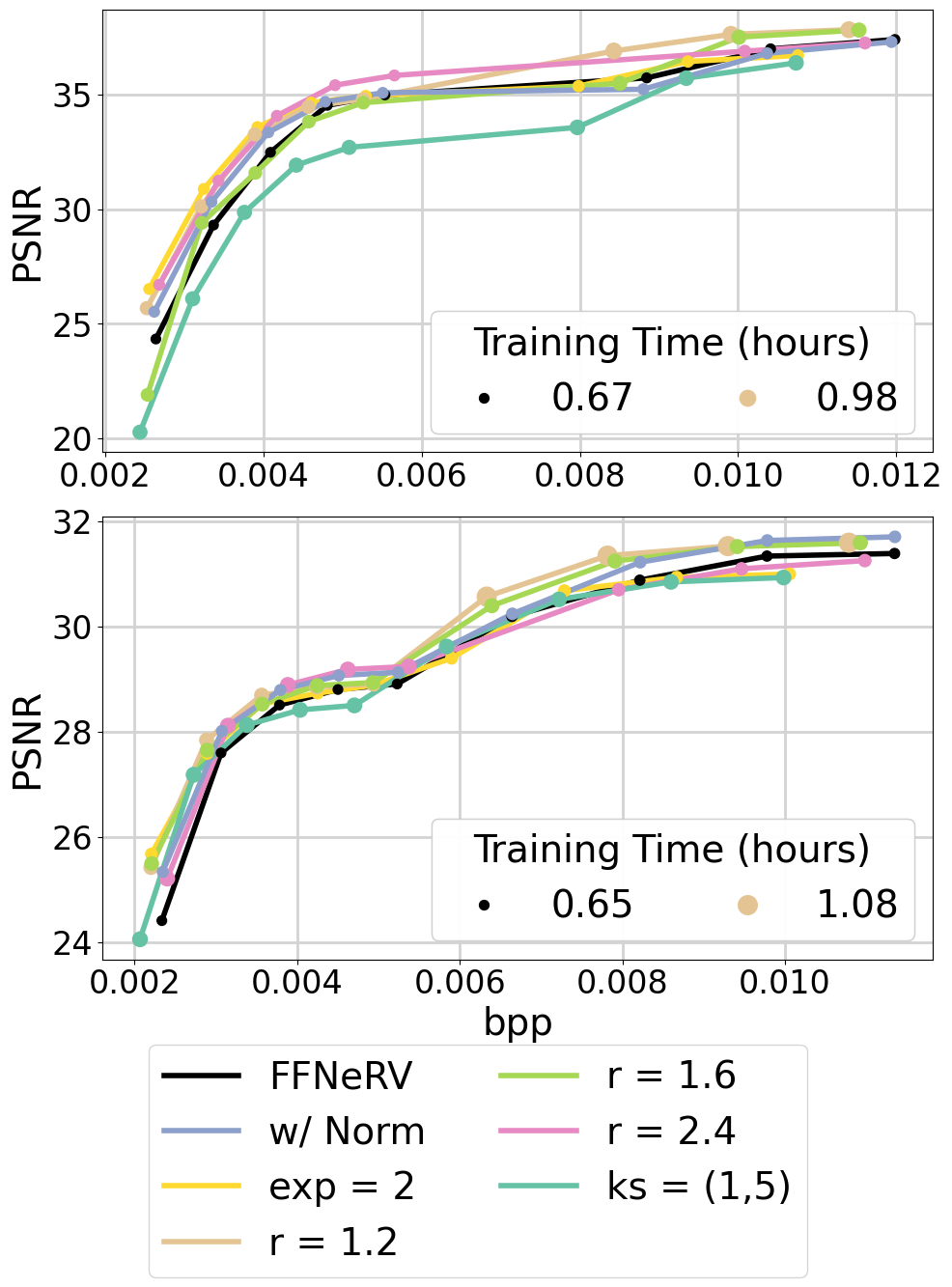}
        \caption{\textbf{Quality vs.\ size}, for HoneyBee (top) and Jockey (bottom) at 1080p for \textbf{FFNeRV} with different \textbf{parameter distributions}.}
        \label{fig:ffnerv_params_psnr_bpp}
    \end{minipage}
    \hfill
    \begin{minipage}{0.32\linewidth}
    \includegraphics[width=1.0\textwidth]{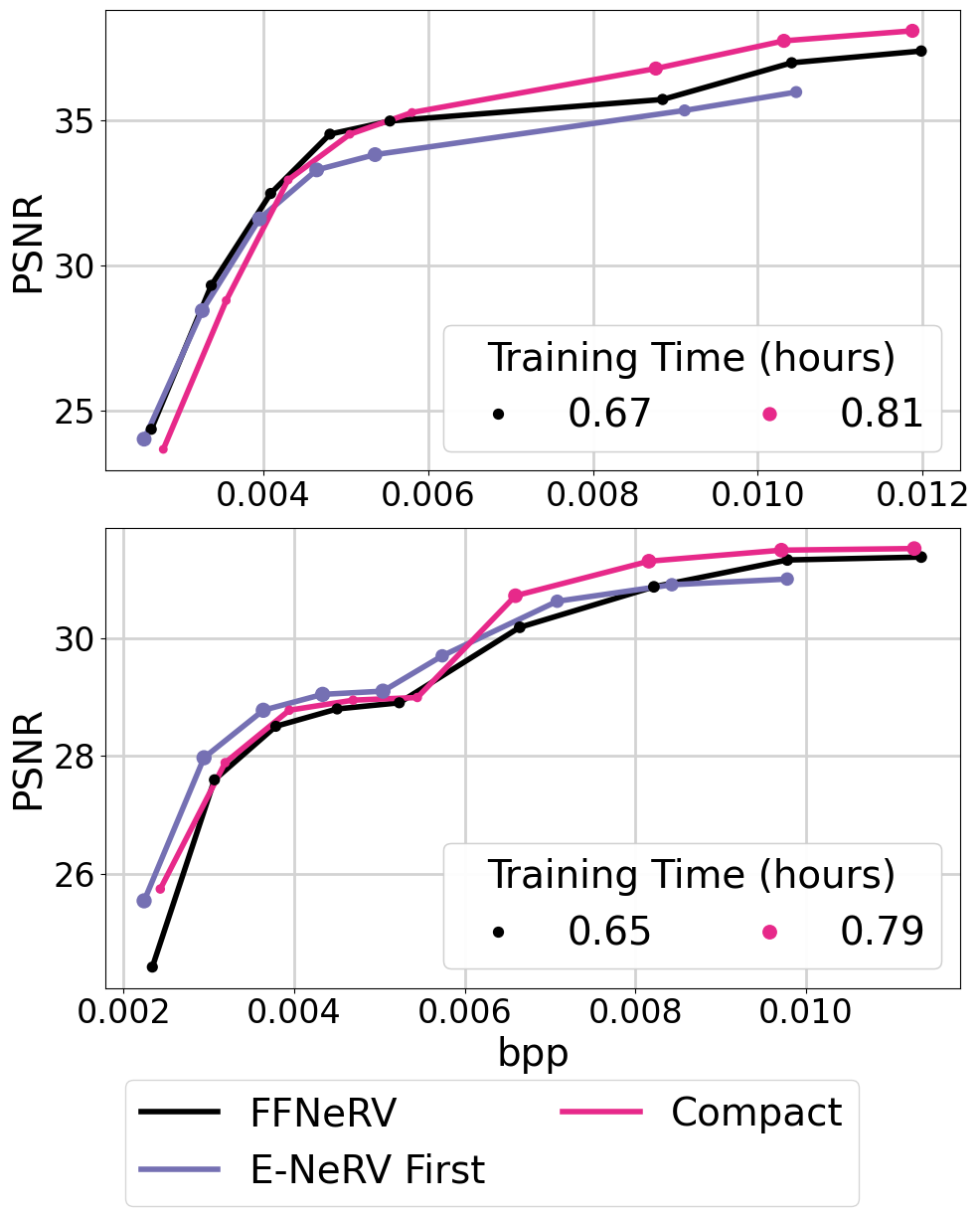}
        \caption{\textbf{Quality vs.\ size}, for HoneyBee (top) and Jockey (bottom) at 1080p for \textbf{FFNeRV} with different \textbf{block designs}.}
        \label{fig:ffnerv_blocks_psnr_bpp}
    \end{minipage}
\vspace{-1.0em}
\end{center}
\end{figure*}

\begin{figure*}[t]
\begin{center}
    \begin{minipage}{0.32\linewidth}
    \includegraphics[width=1.0\textwidth]{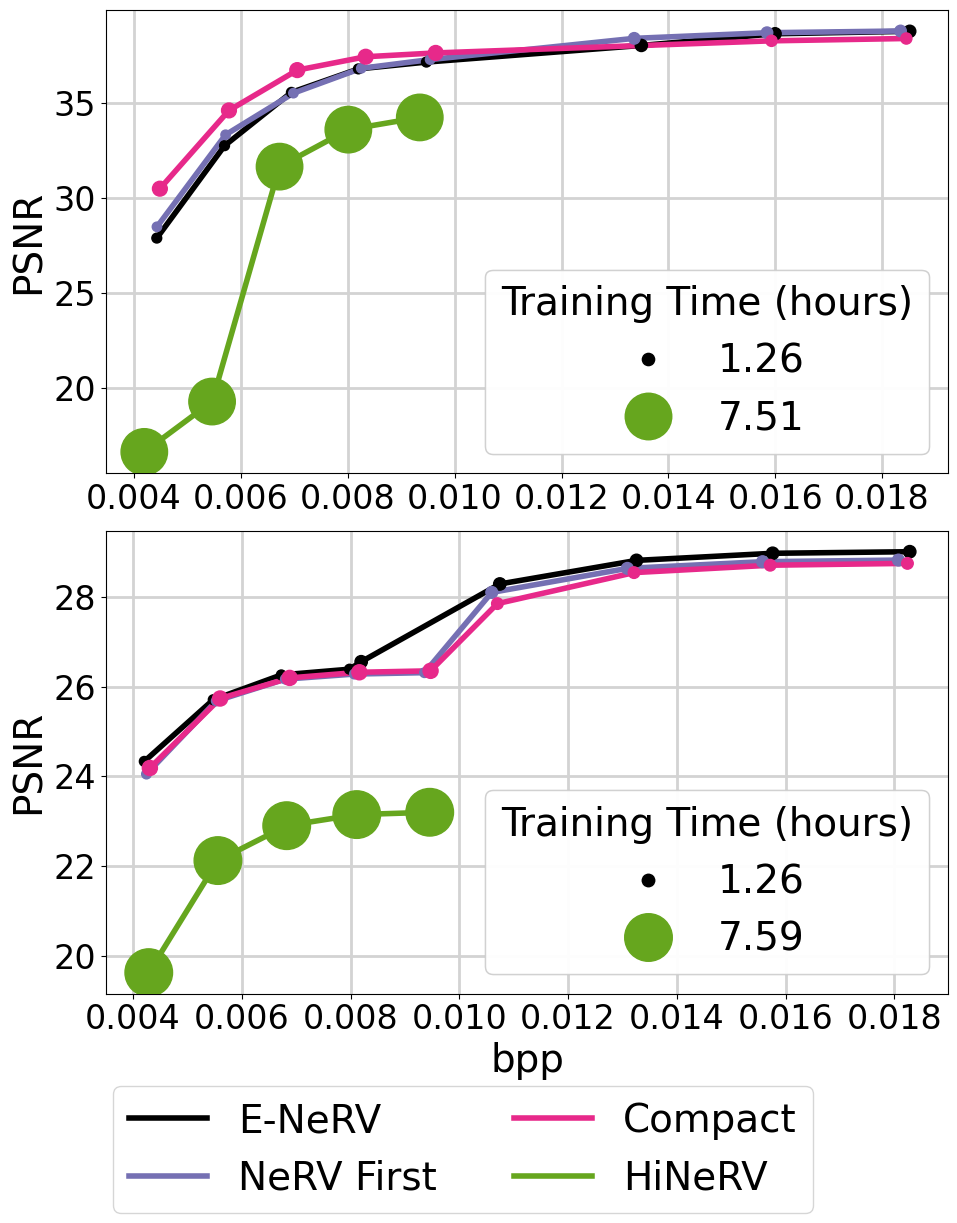}
        \caption{\textbf{Quality vs.\ size}, for HoneyBee (top) and Jockey (bottom) at 1080p for \textbf{E-NeRV} for various \textbf{block designs}.}
        \label{fig:enerv_blocks_psnr_bpp}
    \end{minipage}
    \hfill
    \begin{minipage}{0.32\linewidth}
    \includegraphics[width=1.0\textwidth]{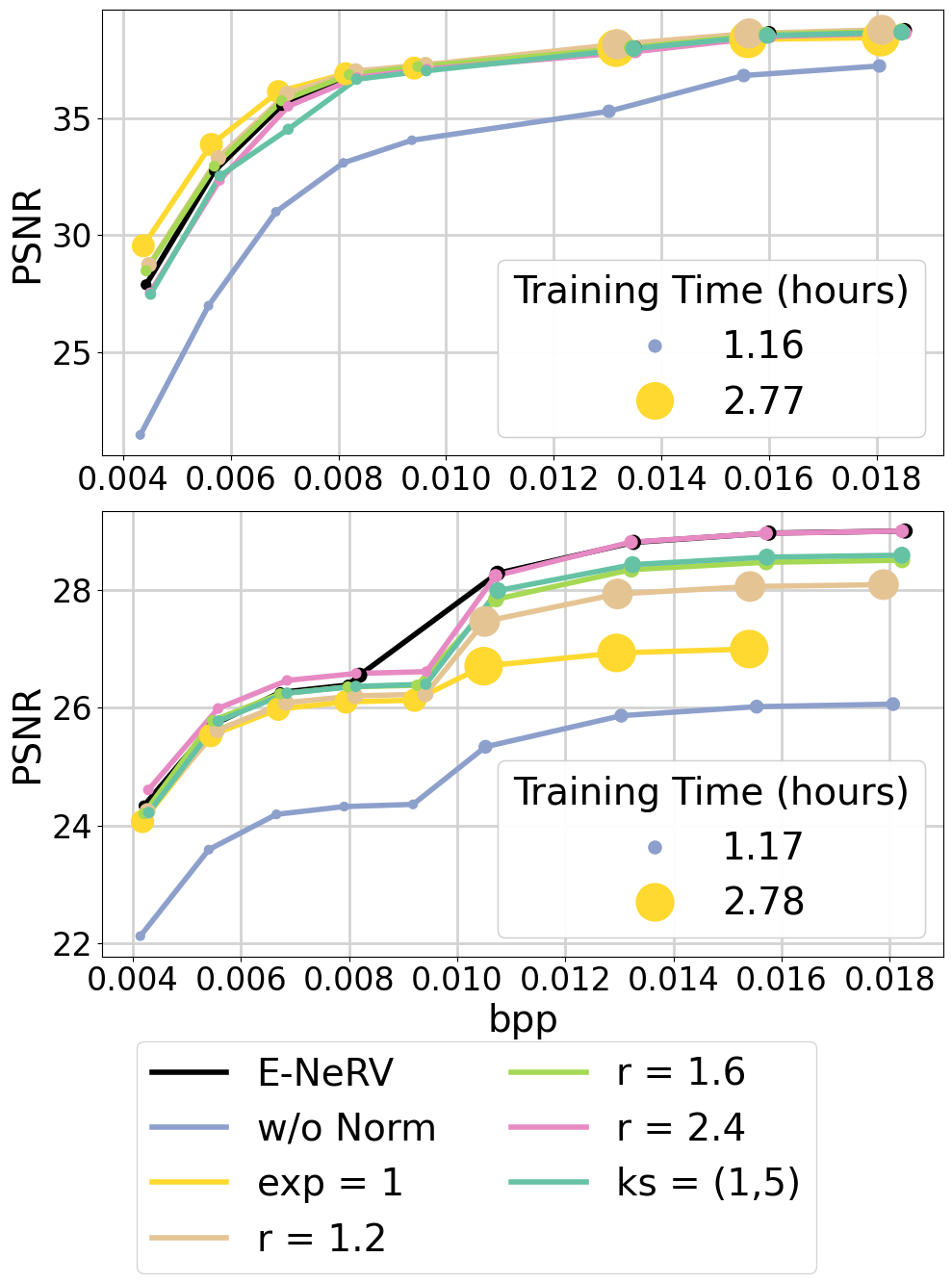}
        \caption{\textbf{Quality vs.\ size}, for HoneyBee (top) and Jockey (bottom) at 1080p for \textbf{E-NeRV} with different \textbf{parameter distributions}.}
        \label{fig:enerv_params_psnr_bpp}
    \end{minipage}
    \hfill
    \begin{minipage}{0.32\linewidth}
    \includegraphics[width=1.0\textwidth]{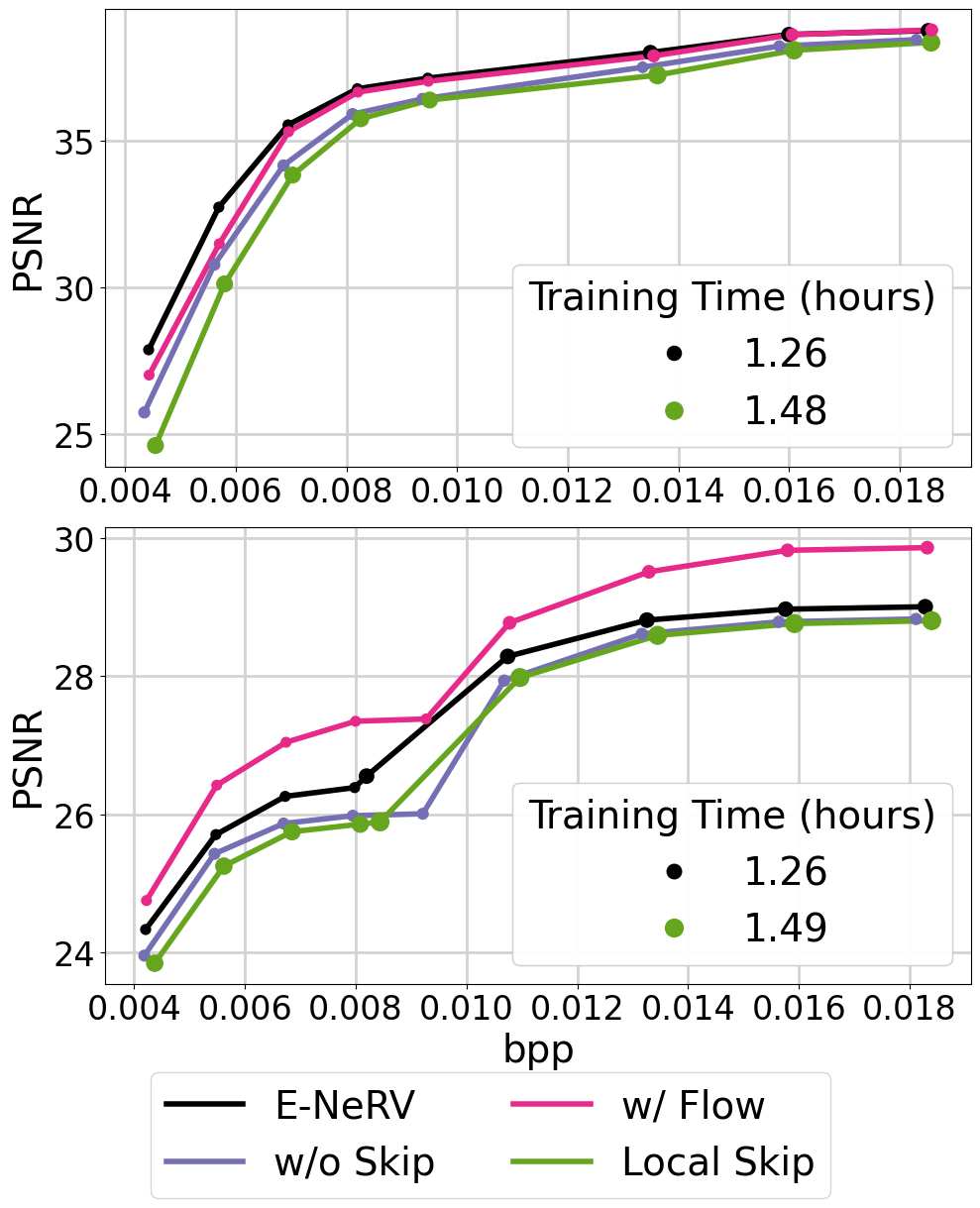}
        \caption{\textbf{Quality vs.\ size}, for HoneyBee (top) and Jockey (bottom) at 1080p for \textbf{E-NeRV} with different \textbf{skip} connections and without \textbf{flow} warping.}
        \label{fig:enerv_skips_psnr_bpp}
    \end{minipage}
\vspace{-1.0em}
\end{center}
\end{figure*}

\begin{figure*}[t]
\begin{center}
    \begin{minipage}{0.32\linewidth}
    \includegraphics[width=1.0\textwidth]{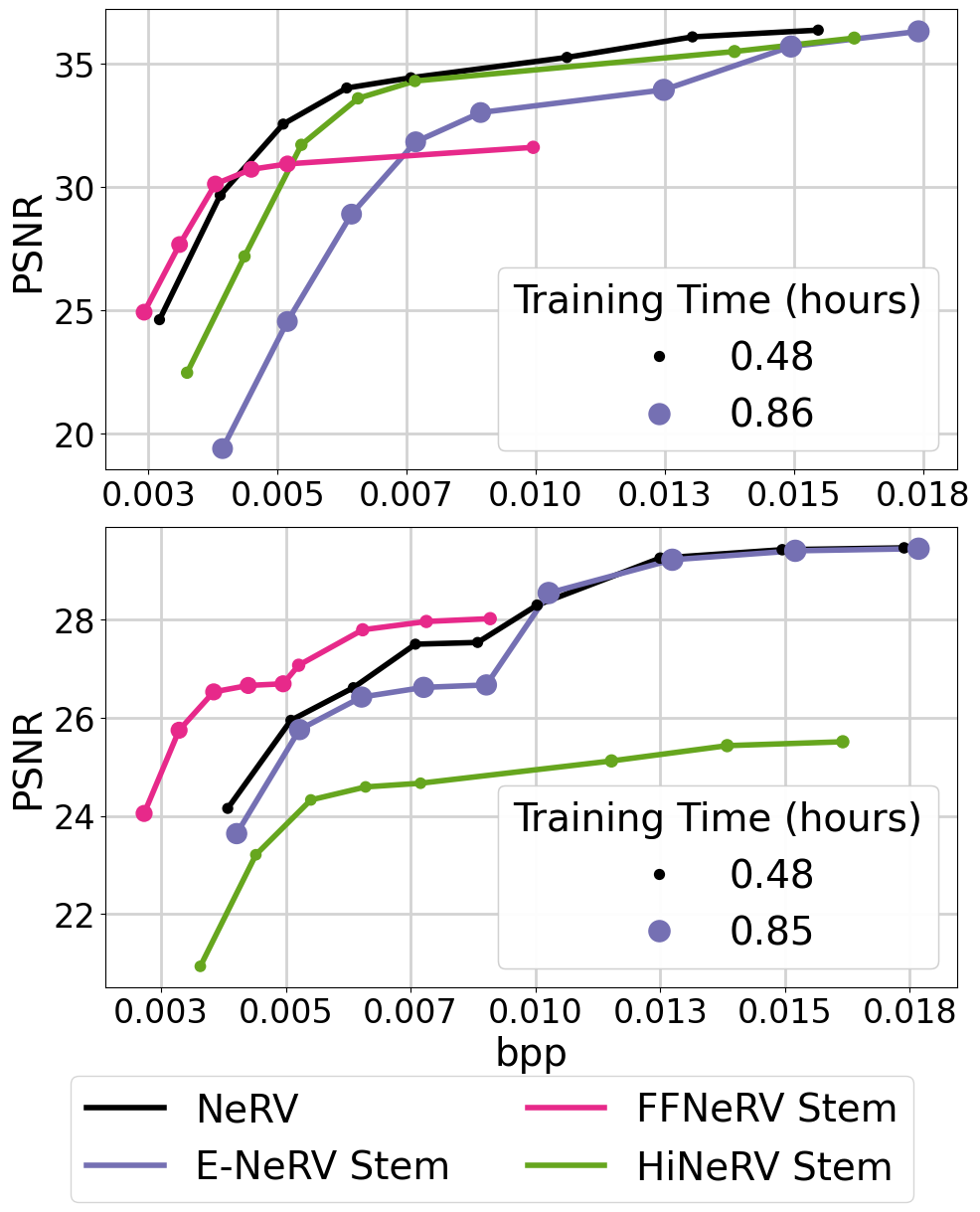}
        \caption{\textbf{Quality vs.\ size}, for HoneyBee (top) and Jockey (bottom) at 1080p for \textbf{NeRV} for various \textbf{position-stem} combinations.}
        \label{fig:nerv_stems_psnr_bpp}
    \end{minipage}
    \hfill
    \begin{minipage}{0.32\linewidth}
    \includegraphics[width=1.0\textwidth]{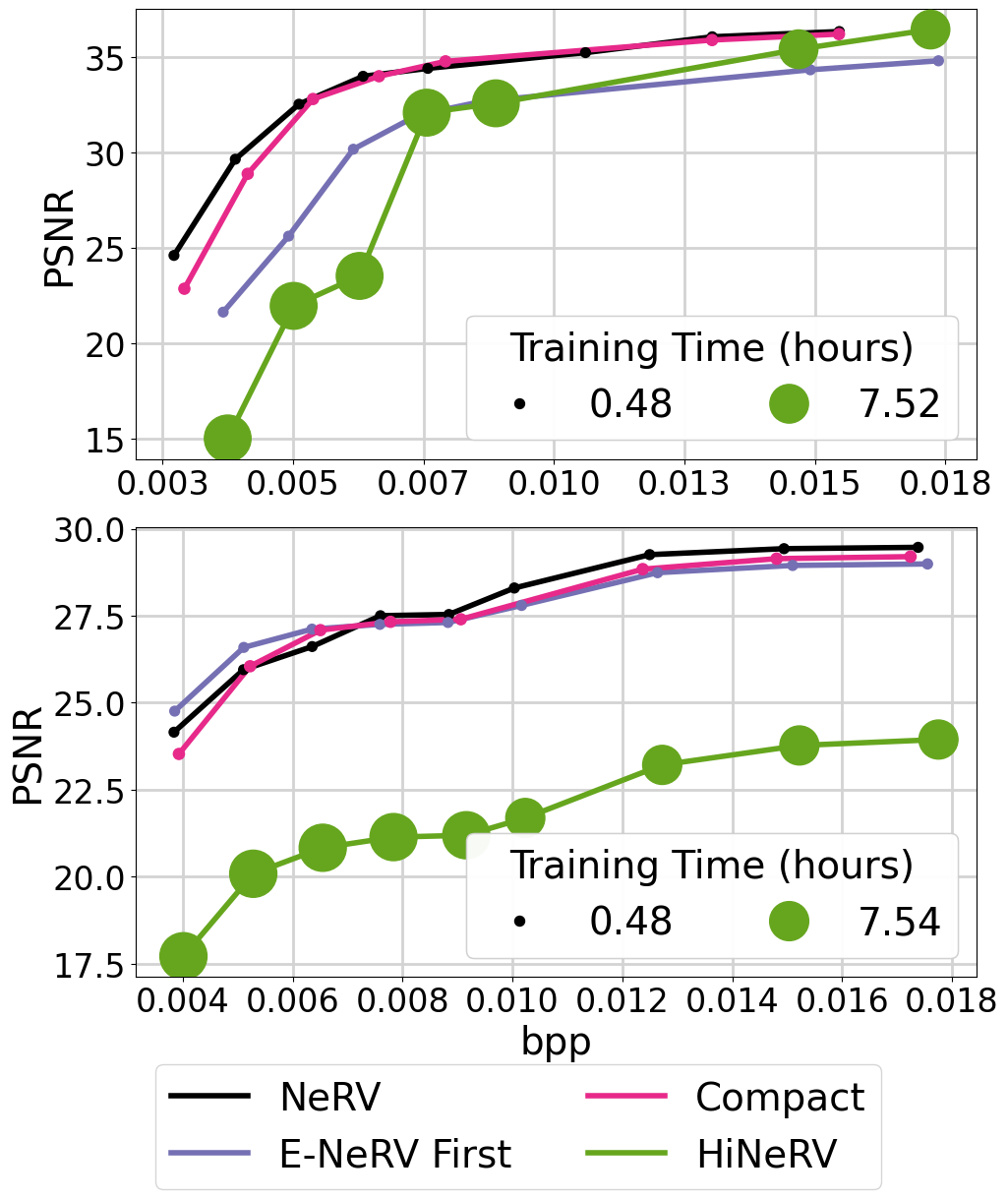}
        \caption{\textbf{Quality vs.\ size}, for HoneyBee (top) and Jockey (bottom) at 1080p for \textbf{NeRV} with different \textbf{block designs}.}
        \label{fig:nerv_blocks_psnr_bpp}
    \end{minipage}
    \hfill
    \begin{minipage}{0.32\linewidth}
    \includegraphics[width=1.0\textwidth]{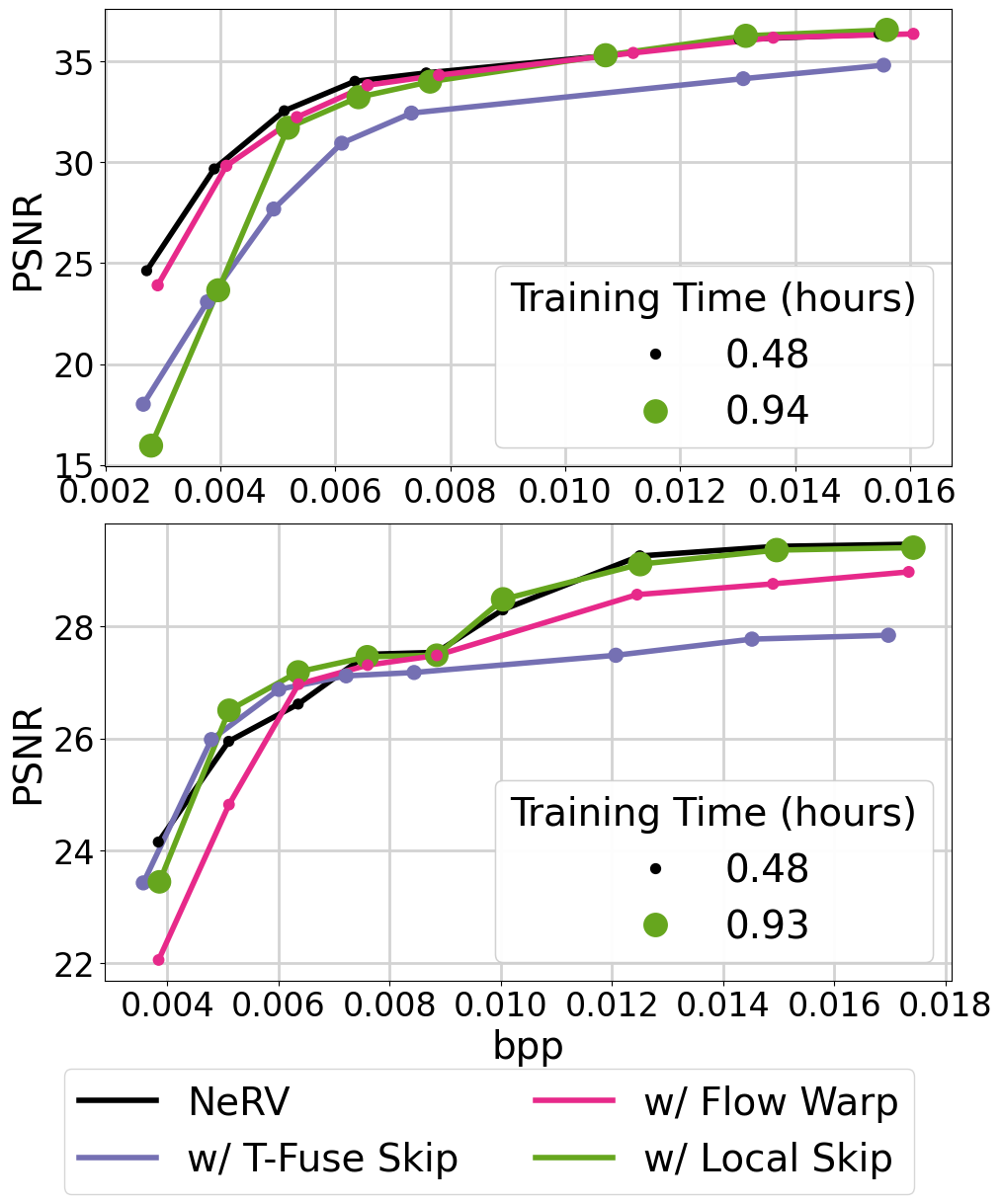}
        \caption{\textbf{Quality vs.\ size}, for HoneyBee (top) and Jockey (bottom) at 1080p for \textbf{NeRV} with different \textbf{skip} connections and without \textbf{flow} warping.}
        \label{fig:nerv_skips_psnr_bpp}
    \end{minipage}
\vspace{-1.0em}
\end{center}
\end{figure*}

\begin{figure*}[t]
\begin{center}
    \begin{minipage}{0.32\linewidth}
    \includegraphics[width=1.0\textwidth]{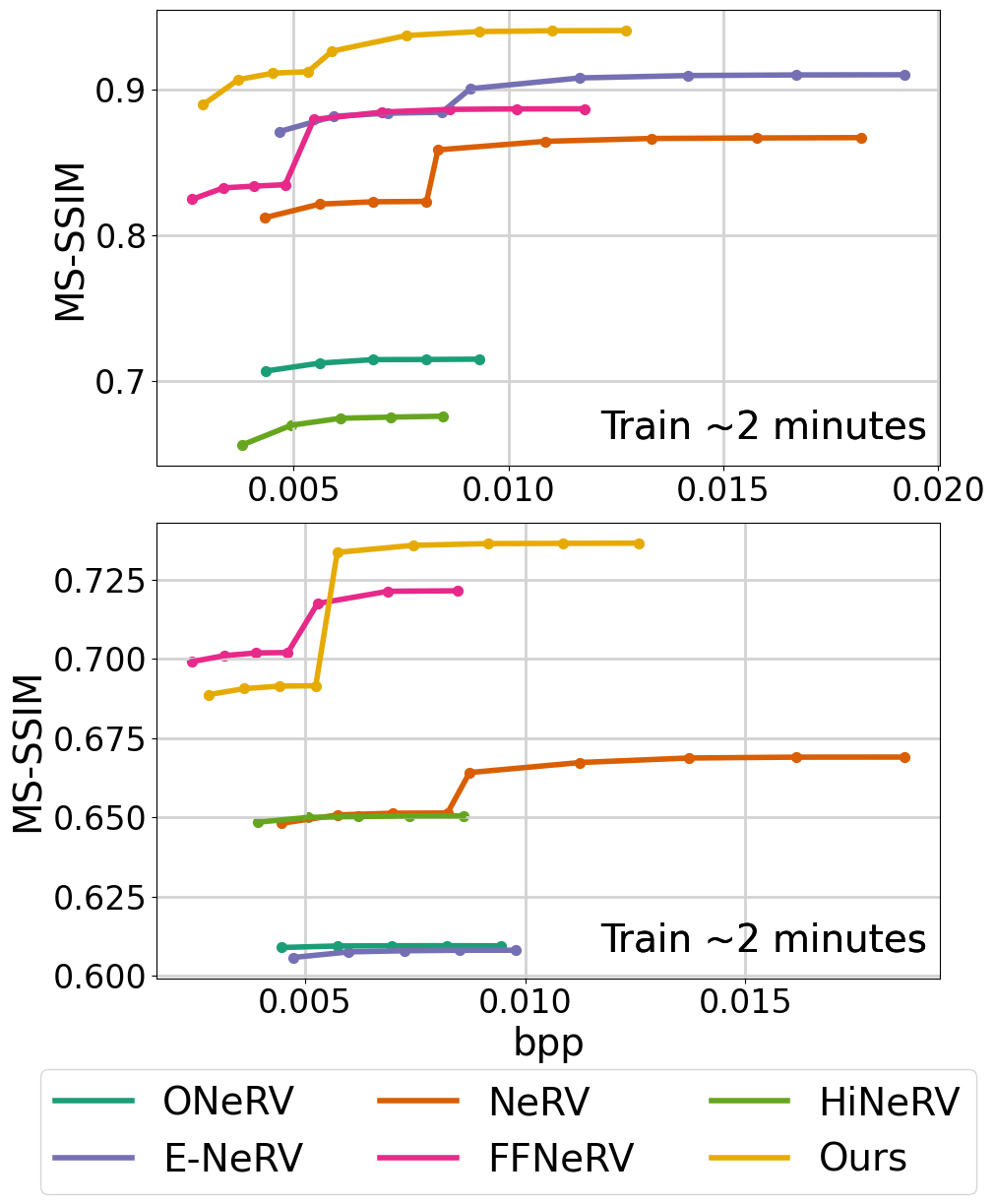}
        \caption{\textbf{Quality vs.\ size}, for HoneyBee (top) and Jockey (bottom) at 1080p with ``short'' training time.}
        \label{fig:short_time_ssim_bpp}
    \end{minipage}
    \hfill
    \begin{minipage}{0.32\linewidth}
    \includegraphics[width=1.0\textwidth]{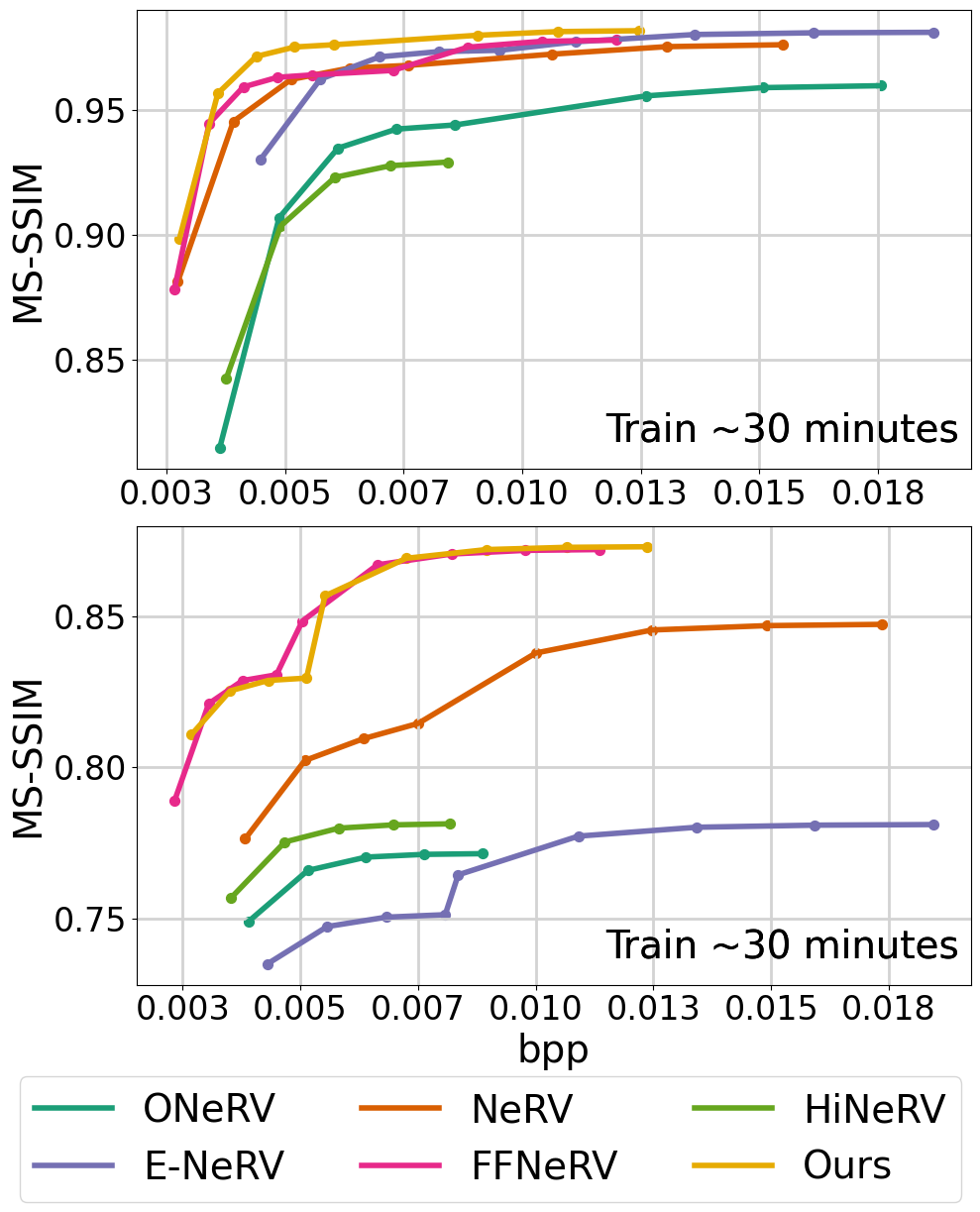}
        \caption{\textbf{Quality vs.\ size}, for HoneyBee (top) and Jockey (bottom) at 1080p with ``medium'' training time.}
        \label{fig:medium_time_ssim_bpp}
    \end{minipage}
    \hfill
    \begin{minipage}{0.32\linewidth}
    \includegraphics[width=1.0\textwidth]{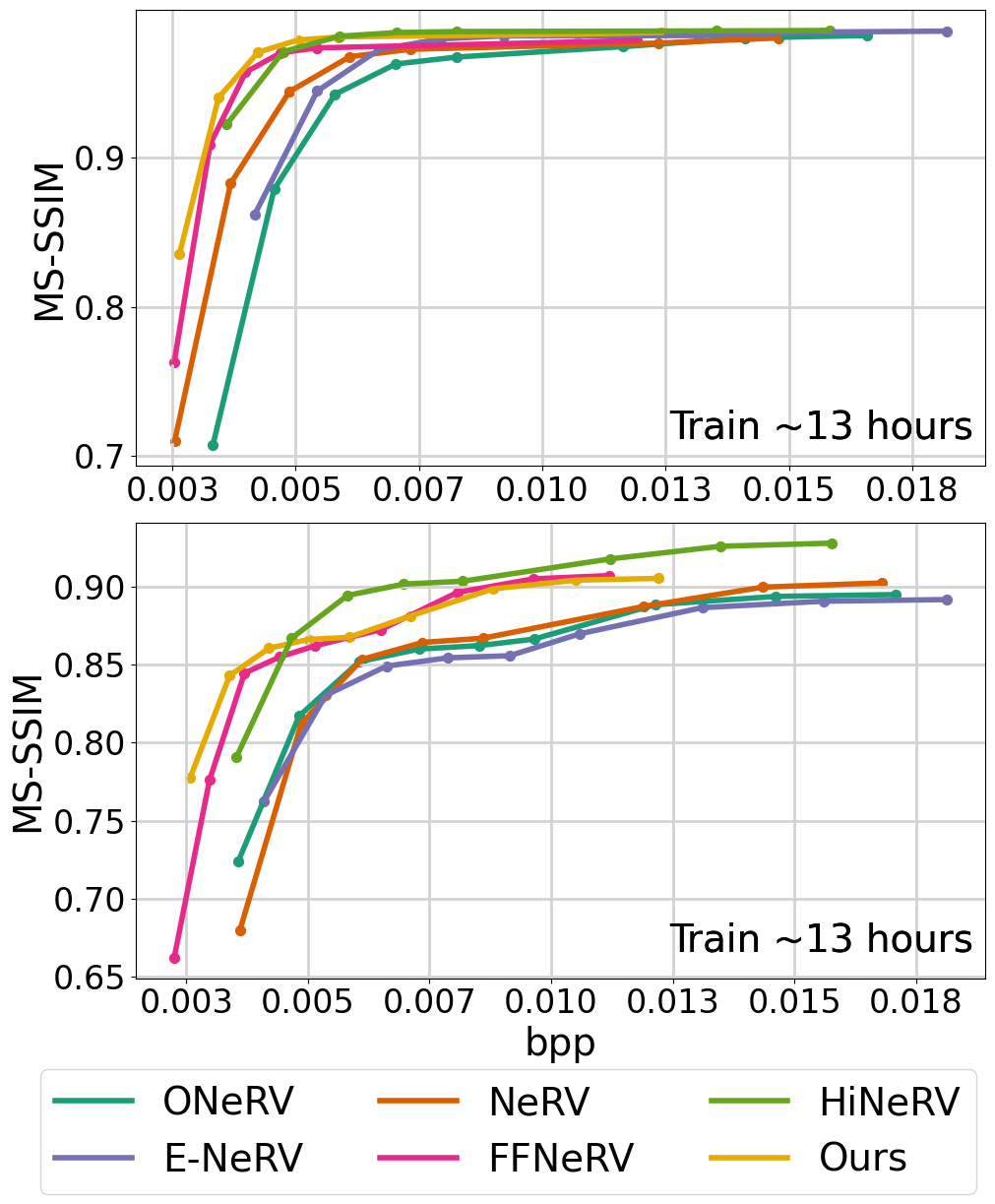}
        \caption{\textbf{Quality vs.\ size}, for HoneyBee (top) and Jockey (bottom) at 1080p with ``long'' training time.}
        \label{fig:long_time_ssim_bpp}
    \end{minipage}
    \vspace{-1.5em}
\end{center}
\end{figure*}

\begin{figure*}[t]
\begin{center}
    \begin{minipage}{0.45\linewidth}
    \includegraphics[width=1.0\textwidth]{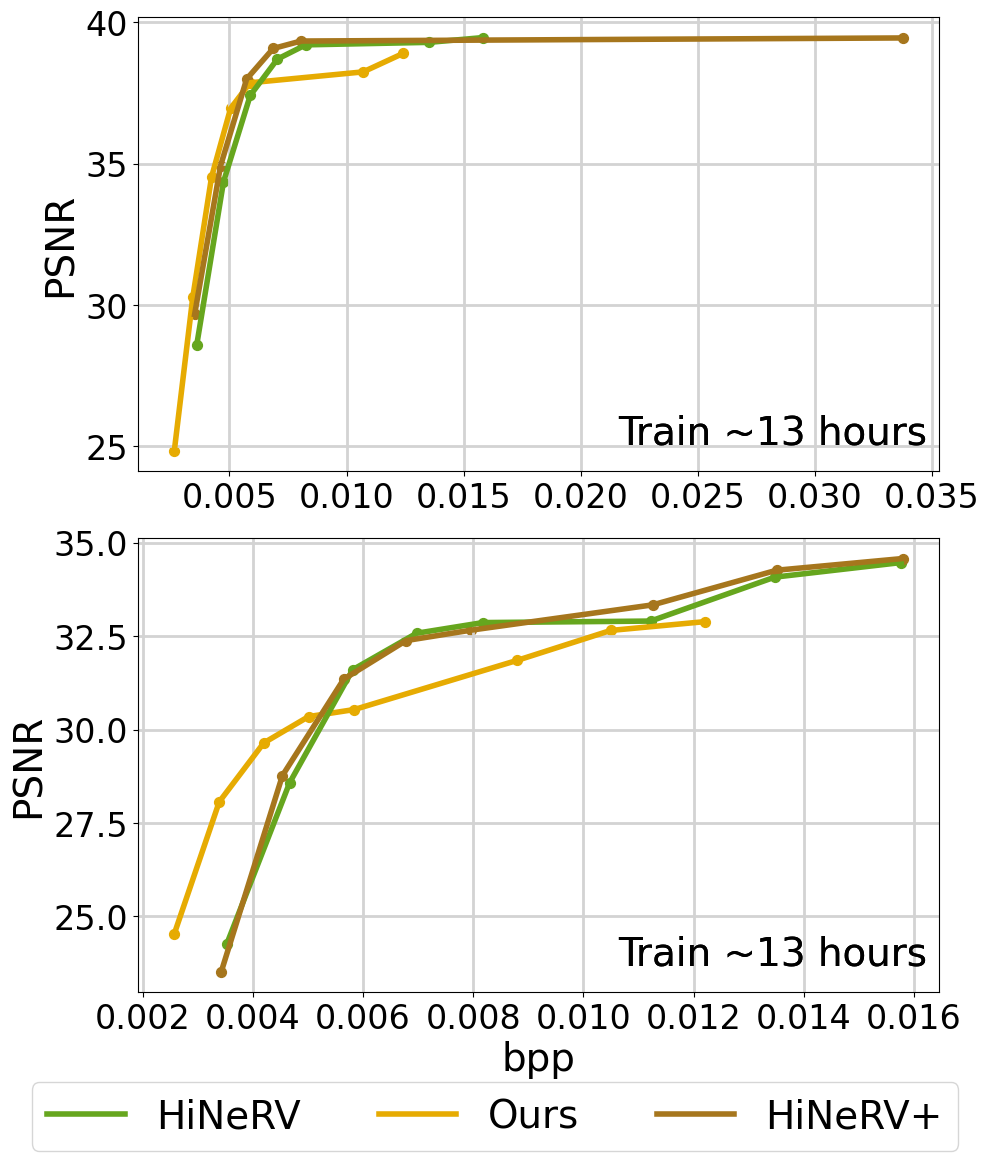}
        \caption{\textbf{Quality vs.\ size}, for HoneyBee (top) and Jockey (bottom) at 1080p.}
        \label{fig:videos_long_sup_psnr}
    \end{minipage}
    \hfill
    \begin{minipage}{0.45\linewidth}
    \includegraphics[width=1.0\textwidth]{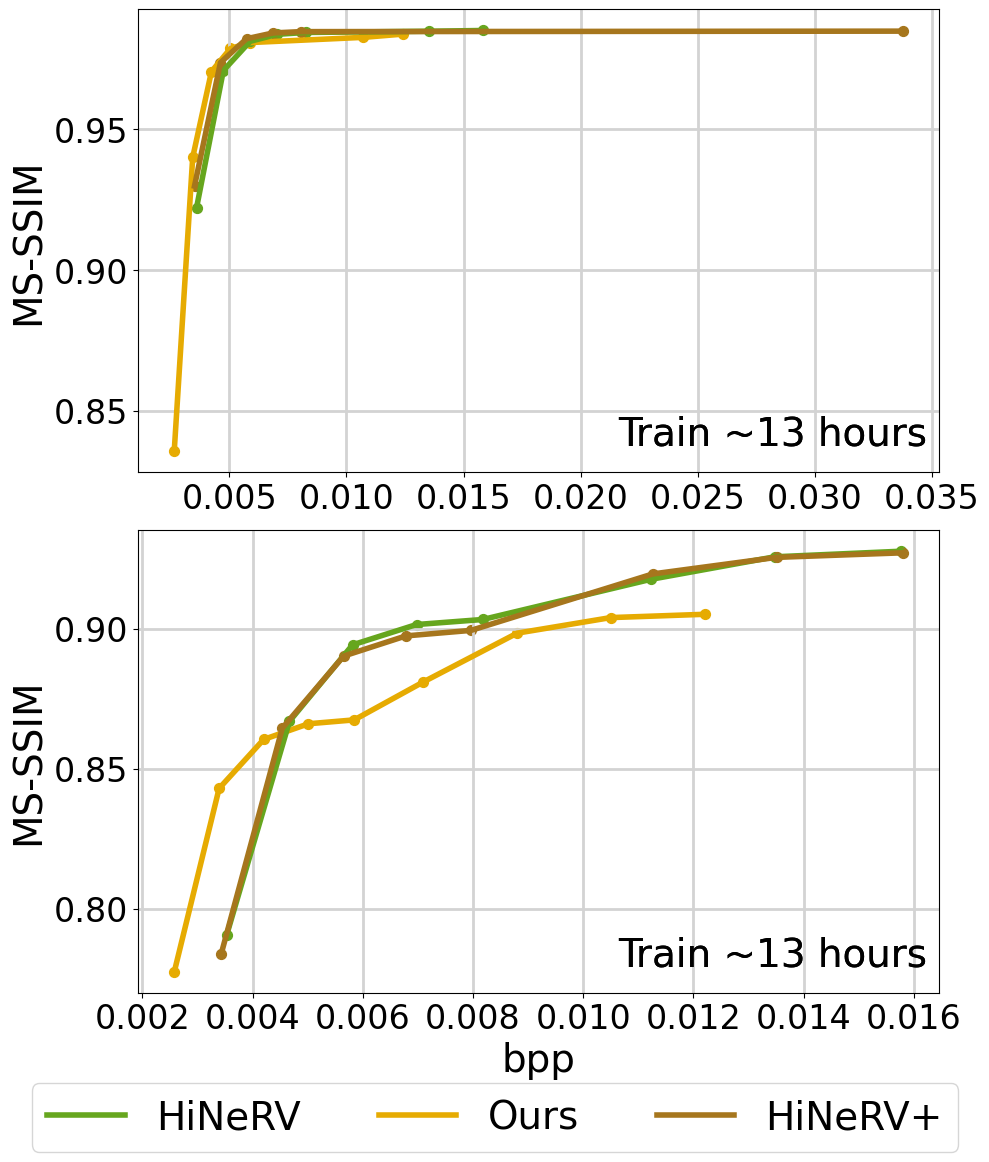}
        \caption{\textbf{Quality vs.\ size}, for HoneyBee (top) and Jockey (bottom) at 1080p.}
        \label{fig:videos_long_sup_ssim}
    \end{minipage}
\vspace{-1.0em}
\end{center}
\end{figure*}

\section{Additional Results}

We report PSNR and MS-SSIM vs. bpp for an average across the 7 videos we use in this paper, for ``short'' training time in Figure~\ref{fig:videos_average_short}, and for ``medium'' training time in Figure~\ref{fig:videos_average_medium}.
When we compute an average, we average in both size and quality simultaneously.
Specifically, for each model, for every video, we take the bpp and PSNR (or MS-SSIM) for the 3M parameter model, with 8 bit quantization plus arithmetic coding, and average them.
We do this for 7 bits, 6 bits, and so on.
We repeat the process for the 1.5M parameter models.

We report MS-SSIM vs. bpp corresponding to the figures in Section~\ref{subsec:investigating} in Figure~\ref{fig:enerv_stems_ssim_bpp}, Figure~\ref{fig:nerv_params_ssim_bpp}, and Figure~\ref{fig:ffnerv_skips_flow_ssim_bpp}.
We report more ablations for FFNeRV in terms of PSNR in Figure~\ref{fig:ffnerv_stems_psnr_bpp}, Figure~\ref{fig:ffnerv_params_psnr_bpp}, and Figure~\ref{fig:ffnerv_blocks_psnr_bpp}.
We report more ablations for E-NeRV in terms of PSNR in Figure~\ref{fig:enerv_blocks_psnr_bpp}, Figure~\ref{fig:enerv_params_psnr_bpp}, and Figure~\ref{fig:enerv_skips_psnr_bpp}.
We report more ablations for NeRV in terms of PSNR in Figure~\ref{fig:nerv_stems_psnr_bpp}, Figure~\ref{fig:nerv_blocks_psnr_bpp}, and Figure~\ref{fig:nerv_skips_psnr_bpp}.

We report MS-SSIM vs. bpp corresponding to the figures in Section~\ref{subsec:time} in Figure~\ref{fig:short_time_ssim_bpp}, Figure~\ref{fig:medium_time_ssim_bpp}, and Figure~\ref{fig:long_time_ssim_bpp}.
We also try a version of HiNeRV with $r = 1.6$ and compare it to HiNeRV and RNeRV in the ``long'' time setting, for PSNR in Figure~\ref{fig:videos_long_sup_psnr} and MS-SSIM in Figure~\ref{fig:videos_long_sup_ssim}.

We perform evaluations for the k400-trained HyperNeRV on the UVG videos, downsampled to 256 height and center-cropped.
We report the PSNR and MS-SSIM in Table~\ref{tab:hypernet_uvg_results}.
Keep in mind that these results are both higher bpp, lower resolution, and lower quality than anything we observe with the NeRV models themselves.
However, the trends are quite different -- HyperNeRV performs better on the videos with motion, whereas it struggles substantially on the HoneyBee video which is easy for most INR-based methods (perhaps it is out of the training distribution).

We also consider the performancne of the NeRV models with different hybrid encoders in Table~\ref{tab:hybrid_reconstruction_results}.
We reserve these for supplementary due to the results in the ``\# Encode'' column.
With their native settings, HNeRV adds some bpp without any quality gains.
To get some gains, the architecture would have to be reworked (possibly by shifting parameters, but this would add encoding time).
DiffNeRV adds a very large amount of extra information to store which renders any comparison with other methods unfair.
We leave the issue of reconciling the size discrepancies between these and the other methods for future work.

\section{Implementation Details}
\label{sec:implementation_details}

\subsection{NeRV Compression.}

We follow the quantization procedure as explained in HiNeRV~\cite{kwan2023hinerv}.
Any time we report bpp, we measure the actual space requirement after compressing with torchac~\cite{mentzer2019practical}.
We then measure the quality after loading the compressed model.
While our library supports quantization-aware training, we observe limited benefit and skip it.

To compute a PSNR/bpp or MS-SSIM/bpp curve, we train each method at a 1.5 million and 3 million parameter setting.
We then quantize each setting at 8, 7, 6, 5, and 4 bits, then compress it with torchac.
We then report the quality/size pairs in strictly ascending order; if some higher size achieves worse quality than a previous size, we do not report it.
This sometimes happens in the case where we transition from an 8-bit 1.5M parameter model, to a 4-bit 3M parameter model, for example.

There are rare cases where the quality/size curve performs well for the points corresponding to the 1.5M parameter model, and poorly for the 3M parameter model (such as with the FFNeRV stem in Figure~\ref{fig:nerv_stems_psnr_bpp}.
This is not necessarily a reflection of poor stem design, but rather a limitation of our study.
That is, every single component of a NeRV model affects the success of every other component.
It is likely that with that our specific configuration for that experiment is suboptimal; perhaps with a different $fc_\text{dim}$ or $r$ or $\texttt{exp}$ we would observe better performance.
In general we test multiple configurations to avoid such cases, but, as our study in part helps prove, it is quite difficult to be sure that one has found a true global optimum for all the different hyperparameters and components of these models.

\subsection{HyperNeRV.} 

Training details for the larger Hyper-Network Hypo-Network settings:
\begin{itemize}
    \item Video: The frames from the input video are organized into clips of 8 consecutive frames each. In contrast to \cite{chen2024fastencodingdecodingimplicit}, we train on all frames of the video (separated into clips). Each clip is treated as an independent input when predicting the weights of the hyponet during training and inference.
    \item Batch size: 32
    \item Tokenizer patch size: 64
    \item HypoNeRV position embedding dimension: 16
    \item HypoNeRV activation layer: GELU~\cite{hendrycks2023gaussianerrorlinearunits}
    \item HypoNeRV $fc_{\text{dim}}$: 20
    \item HypoNeRV kernel sizes (first layer to last): 1, 3, 3, 3
    \item HypoNeRV upscale factors (PixelShuffle strides): 4, 4, 4, 4
    \item HypoNeRV token numbers (first layer to last): 4, 80, 16, 0
    \item HypoNeRV token dimensions (first layer to last): 256, 240, 240, 0
    \item Transformer token dimension and feed-forward dimension for transformer encoder layers: 720 and 2880; 12 heads, 6 blocks, for HyperNeRV of size 47.9M.
    \item Optimizer: Adam
    \item Learning rate: 0.0001
\end{itemize}

\paragraph{HypoNeRV $fc_{\text{dim}} = 16$.} For HypoNeRV with $fc_{\text{dim}} = 16$, the following settings are altered from the base Hyper-Network training.
\begin{itemize}
    \item HypoNeRV $fc_{\text{dim}}$: 16
    \item HypoNeRV layers token numbers: 4, 64, 16, 0
    \item HypoNeRV layers token dimensions: 256, 288, 288, 0
\end{itemize}

\paragraph{Weight masking.} For weight masking, we use the following modified token settings while keeping the other settings same as the base training.

We have two configurations, a larger one, and a smaller one.

\paragraph{Larger weight masking model.}
\begin{itemize}
    \item HypoNeRV layers min token numbers: 1, 32, 4, 0
    \item HypoNeRV layers max token numbers: 2, 64, 8, 0
    \item HypoNeRV masking ratio: 0.5
    \item HypoNeRV token dimensions: 256, 144, 72, 0
\end{itemize}

\paragraph{Smaller weight masking model.}
\begin{itemize}
    \item HypoNeRV layers min token numbers: 0, 16, 2, 0
    \item HypoNeRV layers max token numbers: 0, 32, 4, 0
    \item HypoNeRV masking ratio: 0.5
    \item HypoNeRV token dimensions: 256, 144, 72, 0
\end{itemize}

\begin{table*}
\centering
\caption{HyperNeRV compression results on UVG at $256{\times}256$ resolution. PSNR/MS-SSIM.}
\label{tab:hypernet_uvg_results}
\setlength{\tabcolsep}{3pt}
\begin{tabular}{@{}l c ccccccc@{}}
\toprule
& & \multicolumn{7}{c}{Video} \\
\cmidrule(l){3-9}
Method & \#Params & Beauty & Bosphorus & HoneyBee & Jockey & ShakeNDry & YachtRide & ReadySetGo   \\
\midrule
HyperNeRV & 130k/24.1k & 30.22/0.8370 & 26.52/0.7212 & 21.33/0.5223 & 25.14/0.7376 & 25.66/0.5715 & 25.36/0.6728 & 21.14/0.5821 \\
\bottomrule
\end{tabular}
\end{table*}

\begin{table*}
\centering
\caption{\textbf{NeRV-methods with hybrid encoders}. Both videos are at 1080p, all methods are trained for 300 epochs. Grids and decoder parameters are included in the \# Params; we compute the size of hybrid encodings, \# Encode, (which must be stored in the bitstream) as $fc_h{\times}fc_w{\times}e_{\text{dim}}$. We also include results from the original methods for reference. We do not include these comparisons in the main paper since the relatively large size of the content and difference embeddings make fair comparison very challenging.}
\label{tab:hybrid_reconstruction_results}
\setlength{\tabcolsep}{3pt}
\begin{tabular}{@{}ll cc cc@{}}
\toprule
& & \multicolumn{2}{c}{Storage Size} & \multicolumn{2}{c}{Video} \\
\cmidrule(l){3-4}
\cmidrule(l){5-6}
Method & Encoding & \# Params & \# Encode & HoneyBee & Jockey \\
\midrule
\rowcolor{lightgray!30} &  & 1.5M & n/a & 37.27/0.9790 & 26.44/0.7907 \\
\rowcolor{lightgray!30}\multirow{-2}{*}{E-NeRV} & \multirow{-2}{*}{Fixed} & 3M & n/a & 38.81/0.9835 & 29.02/0.8372 \\
\multirow{2}{*}{E-NeRV} & \multirow{2}{*}{HNeRV} & 1.5M & 0.346M & 35.17/0.9699 & 23.73/0.7291 \\
& & 3M & 0.346M & 38.23/0.9822 & 27.21/0.8025 \\
\rowcolor{lightgray!30} & & 1.5M & 4.47M & 35.46/0.9713 & 29.88/0.8754 \\
\rowcolor{lightgray!30}\multirow{-2}{*}{E-NeRV} & \multirow{-2}{*}{DiffNeRV} & 3M & 4.47M & 38.64/0.9834 & 31.91/0.9004 \\
& & 1.5M & n/a & 35.15/0.9681 & 28.93/0.8422 \\
\multirow{-2}{*}{FFNeRV} & \multirow{-2}{*}{Grid} & 3M & n/a & 37.55/0.9797 & 31.41/0.8859 \\
\rowcolor{lightgray!30} & & 1.5M & 0.346M & 36.18/0.9749 & 27.94/0.8216 \\
\rowcolor{lightgray!30}\multirow{-2}{*}{FFNeRV} & \multirow{-2}{*}{HNeRV} & 3M & 0.346M & 37.68/0.9806 & 30.08/0.8574 \\
 & & 1.5M & n/a & \textbf{39.21}/\textbf{0.9844} & 32.25/0.8978 \\
\multirow{-2}{*}{HiNeRV} & \multirow{-2}{*}{Grid} & 3M & n/a & \textbf{39.39}/\textbf{0.9849} & 34.71/0.9300 \\
\rowcolor{lightgray!30} & & 1.5M & 11.1M & 38.49/0.9826 & \textbf{34.62}/\textbf{0.9384} \\
\rowcolor{lightgray!30}\multirow{-2}{*}{HiNeRV} & \multirow{-2}{*}{HNeRV} & 3M & 11.1M & 39.24/0.9846 & \textbf{35.64}/\textbf{0.9447} \\
& & 1.5M & 58.8M & 37.99/0.9812 & 33.12/0.9170 \\
\multirow{-2}{*}{HiNeRV} & \multirow{-2}{*}{DiffNeRV} & 3M & 58.8M & 38.95/0.9840 & 34.12/0.9268 \\
\rowcolor{lightgray!30} &  & 1.5M & 1.38M & 37.87/0.9805 & 29.90/0.8652 \\
\rowcolor{lightgray!30}\multirow{-2}{*}{HNeRV} & \multirow{-2}{*}{HNeRV} & 3M & 1.38M & 38.72/0.9832 & 31.34/0.8841 \\
& & 1.5M & 4.23M & 37.91/0.9793 & 31.01/0.8845 \\
\multirow{-2}{*}{DiffNeRV} & \multirow{-2}{*}{DiffNeRV} & 3M & 4.23M & 39.19/0.9829 & 32.41/0.9001 \\
\bottomrule
\end{tabular}
\end{table*}

\begin{figure}[ht]
    \centering
    \begin{minipage}{1.0\linewidth}
        \centering
        \includegraphics[width=0.7\textwidth]{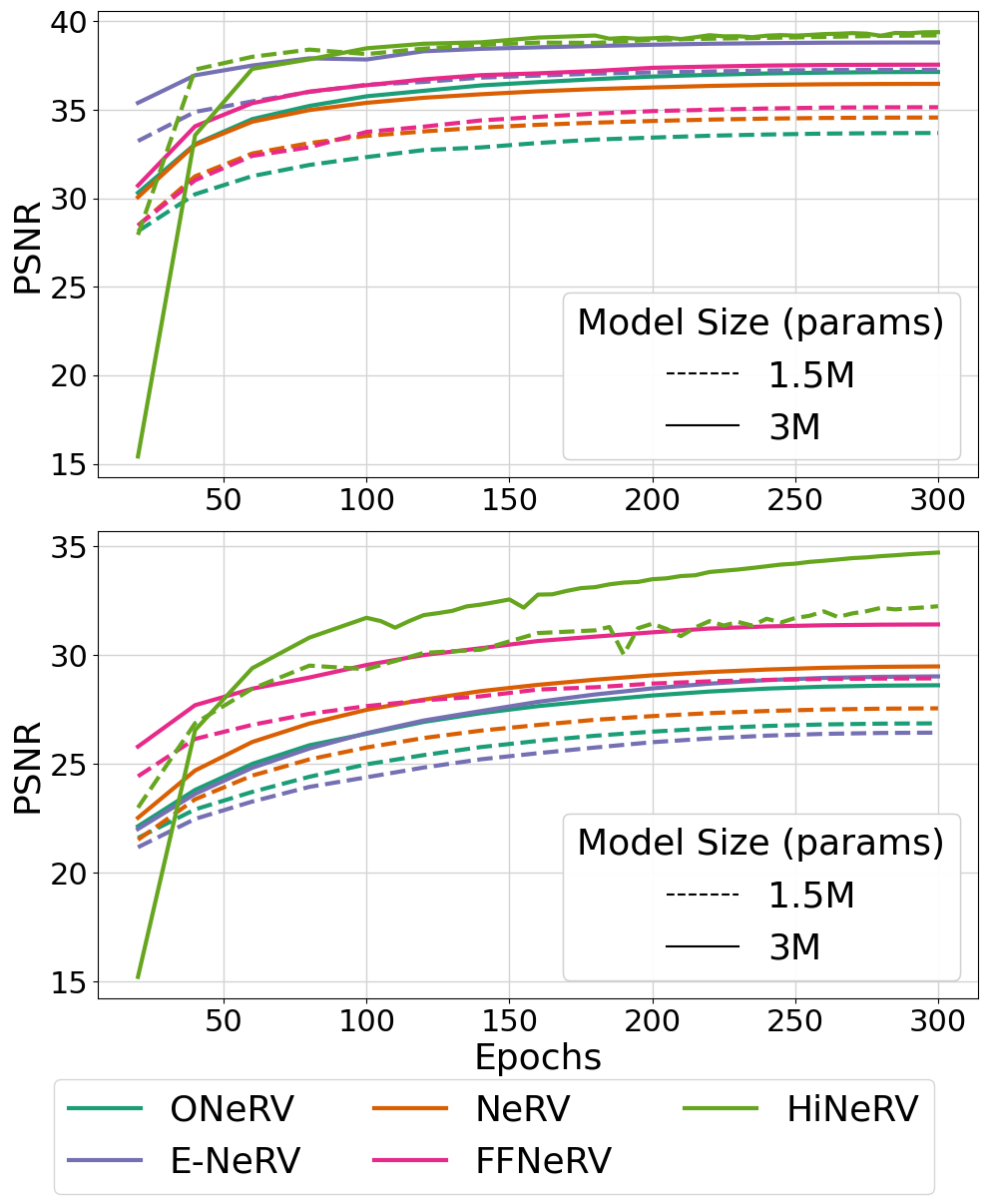}
        \caption{Reconstruction quality over training epochs, for UVG HoneyBee (top) and Jockey (bottom) at 1080p.}
        \label{fig:quality_epochs_main}
    \end{minipage}
\end{figure}

\begin{figure}[ht]
    \centering
    \begin{minipage}{1.0\linewidth}
        \centering
        \includegraphics[width=1.0\textwidth]{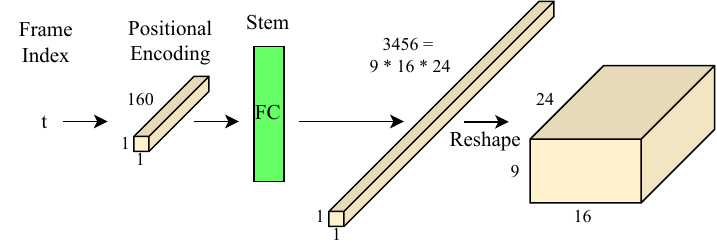}
        \caption{We show a detailed walkthrough of the NeRV stem, including sizes for a NeRV with 160-dimension positional encoding, $fc_\text{dim}=12$, $fc_w=16$, $fc_h=9$. See Figure~\ref{fig:nerv_block_tutorial} for a walkthrough of the NeRV block, whose input is this stem's output.}
        \label{fig:nerv_stem_tutorial}
    \end{minipage}
\end{figure}

\begin{figure*}[ht]
    \centering
    \begin{minipage}{1.0\linewidth}
        \centering
        \includegraphics[width=0.85\textwidth]{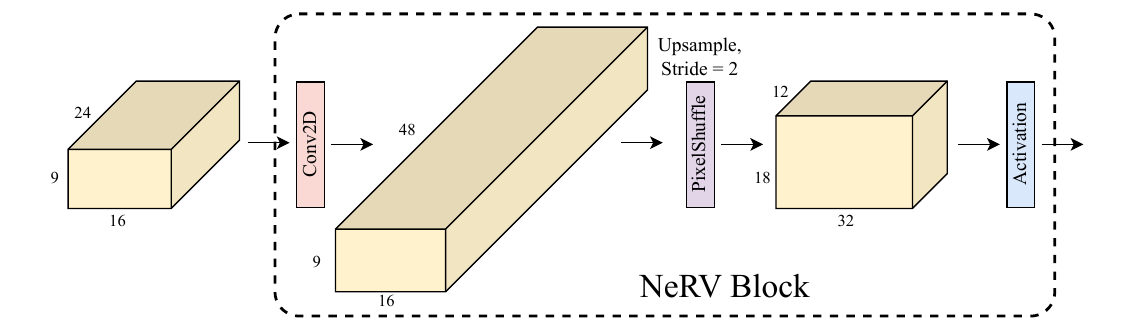}
        \caption{We show an example NeRV block with upsample stride $s=2$ and channel reduction $r=2$, for the features generated by the positional encoding and stem in Figure~\ref{fig:nerv_stem_tutorial}. See Appendix~\ref{sec:nerv_walkthrough} for more details.}
        \label{fig:nerv_block_tutorial}
    \end{minipage}
\end{figure*}

\section{NeRV Walkthrough}
\label{sec:nerv_walkthrough}

See a detailed diagram of the basic NeRV stem from HNeRV~\cite{chen2023hnerv} in Figure~\ref{fig:nerv_stem_tutorial}.
See a detailed diagram of the first NeRV block corresponding to that stem in Figure~\ref{fig:nerv_block_tutorial}.
From there, the features are upsampled by subsequent NeRV blocks, with upsampling dictated by the channel expansion (from convolution) and the PixelShuffle stride which converts channels to spatial dimensions.
Once at full frame/image resolution, the features are processed by a final convolutional layer to convert from feature dimension to color channel dimension (3).
The result is passed through a specialized activation, such as a sigmoid, tanh, or adding 0.5 to the output.
This final output is equivalent to the original image/frame.

\section{INR Hyper-Network Walkthrough}
\label{sec:hypernerv_walkthrough}

We provide a brief walkthrough of the hyper-network setup, although it is explained in prior work as well~\cite{chen2024fastencodingdecodingimplicit}.
To understand the INR hyper-network setup, it is easiest to start with the prediction of the hyper-network: the hypo-network.
For our purposes, this network is a simple sequence of 4 NeRV blocks, which take a time positional encoding as input, and upsample it (via convolution and PixelShuffle) until it is the size of the image the NeRV model is meant to represent.
This NeRV model has a total of 4 learnable convolutional layers.
So, our hyper-network must predict the parameters for 4 convolutional layers.

The model does so in two parts.
First, it initializes and learns a set of base, or ``shared'' parameters, which correspond to the hypo-network exactly.
That is, the base parameters contain all the weights and the biases corresponding to all 4 layers of the hypo-network.
If we want to have an 85.6k parameter hypo-network, we learn 85.6k base parameters.
Remember that the hyper-network learns at the dataset-level, not the video level.
In a realistic setting, this would be installed as an encoder, and the base parameters would be installed as part of the decoder.
They would not be transmitted over the wire in real-time, since they can be learned up front.
Or else, they would be transmitted, but only a single time.
So, we do not consider the base parameters when computing bpp or otherwise making compression storage considerations.

\begin{figure*}[ht]
    \centering
    \begin{minipage}{1.0\linewidth}
        \centering
        \includegraphics[width=0.85\textwidth]{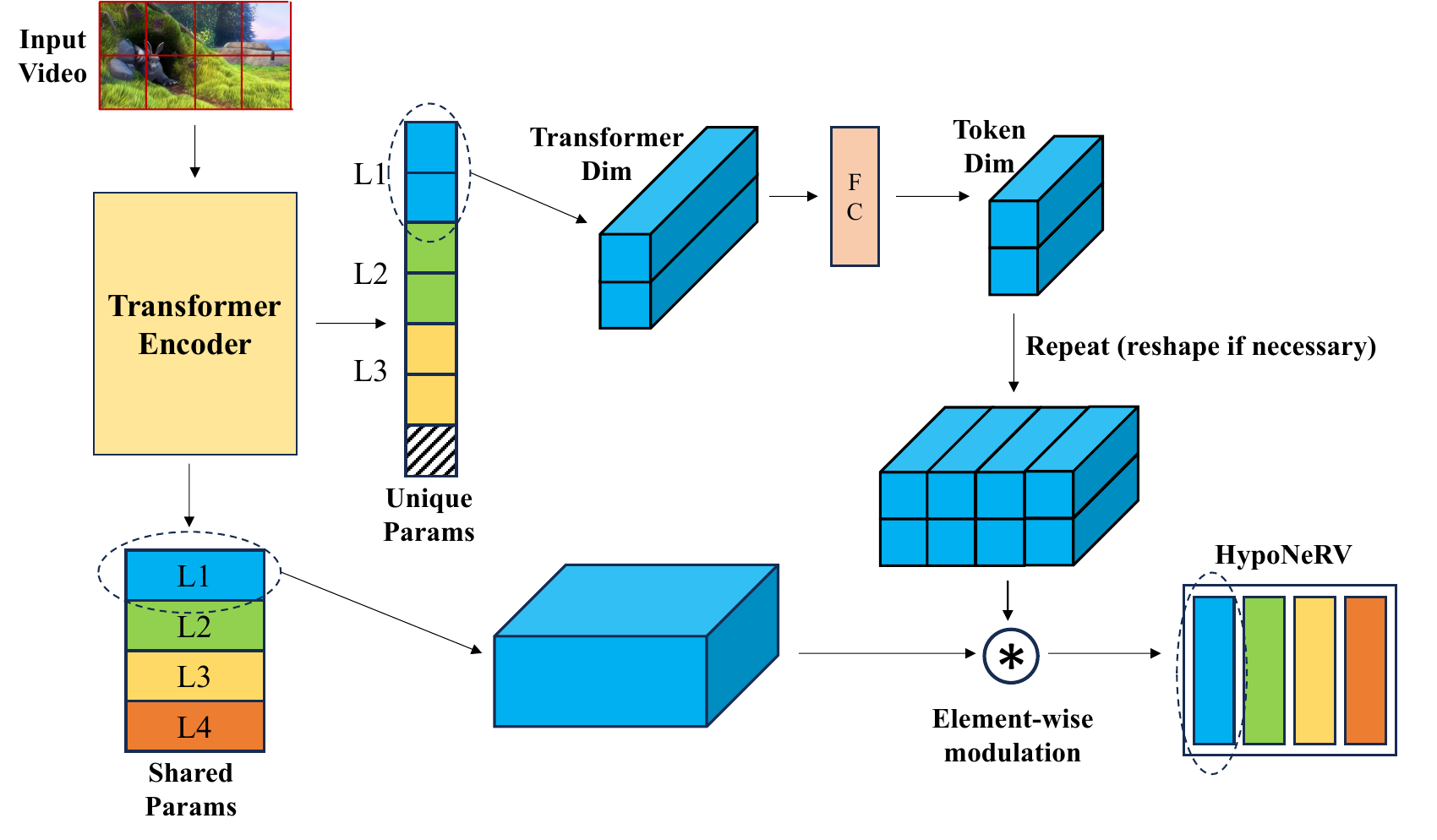}
        \caption{We show example token predictions, FC processing, repeat/reshape, and modulation with the ``shared'' parameters for a single layer of the hypo-network. The unique parameters (the predictions of the weight token FC layer, just before repeat/reshape), are what we store, quantize, encode, and measure for bpp.}
        \label{fig:hypernerv_tutorial}
    \end{minipage}
\end{figure*}

On their own, the base parameters cannot represent every single video.
Instead, they are modulated, by the transformer backbone of the hyper-network, for any given input clip.
To perform this modulation, we predict a set of ``unique parameters.''
We set designate a certain amount of learnable weight tokens, corresponding to the different layers of the hypo-network, and give these as input to the hyper-network, along with the video tokens.
We take the output weight tokens for each layer, and process them with fully-connected layers to convert from the transformer token dimension to a weight token dimension.
We then repeat and reshape the post-FC tokens, perform an elementwise multiplication with the shared parameters, and normalize.
The result is the hypo-network corresponding to the input clip.
We illustrate this process for a single layer in Figure~\ref{fig:hypernerv_tutorial}.

\section{XINC Analysis Walkthrough}
\subsection{Adapting XINC}
XINC, introduced in \cite{Padmanabhan_2024_CVPR}, is a framework designed to investigate how neurons in an image or video INR encode signals they are trained to represent. By dissecting the contributions of each neuron in every layer of the network to each output pixel, the framework aids in understanding the behavior of these INRs.
We adapt this framework to analyze the last (head) layer of the various NeRV variants we study in this work, as well as the HypoNeRV model. Specifically, we generate contribution maps for the kernels in the head layer of these models (Figures ~\ref{fig:xinc_main}, \ref{fig:xinc_hyponerv_grid} and \ref{fig:xinc_suppl_all_nervs}) and leverage these maps for motion analysis (Figures \ref{fig:xinc_motion} and \ref{fig:xinc_motion_all_nervs}). 

For the ONeRV, NeRV, HNeRV, E-NeRV and FFNeRV networks, integrating XINC is relatively straightforward since their head layers consist of a single learnable convolutional layer, akin to the NeRV model studied in \cite{Padmanabhan_2024_CVPR}, followed by an optional activation function. Given an input $v_{in}$ of shape $h \times w \times ch_{in}$ and $ch_{out}$ output channels, each of the $ch_{in} * ch_{out}$ convolutional kernels is treated as a neuron with a distinct contribution map. We compute these maps by independently convolving each kernel over $v_{in}$ and storing the outputs to produce a set of $h \times w \times (ch_{in} * ch_{out})$ maps. These maps are subsequently passed through an optional activation layer to obtain final kernel contributions.

For HiNeRV, we begin by following the approach outlined above. However, since HiNeRV is a patch-based method, we stitch together contributions to pixels in different patches of a frame to construct the full-frame kernel contributions. 

Adapting XINC for the HypoNeRV requires additional considerations due to the presence of a PixelShuffle operation in the head layer, which redistributes channel information into the spatial domain. To correctly interpret kernel contributions at the output resolution, we account for this rearrangement by following the procedure outlined in \cite{Padmanabhan_2024_CVPR}. We omit applying proximity correction since there are no downstream convolutional layers that follow the head layer, as was originally applied in XINC.

\subsection{XINC on HyperNeRV}

\begin{figure}[t]
  \centering
   \includegraphics[width=0.95\linewidth]{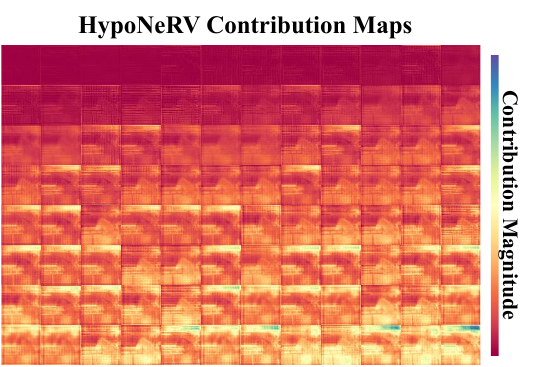}

    \caption{\textbf{XINC contribution maps} on the last (head) layer of HypoNeRV, for Jockey at 256×256 resolution. We sort kernels by total contribution magnitude and select a subset of uniformly sampled kernels in the head layer.
    For the sake of interpretability, we remove locations at which kernel contributions are rendered zero due to the strided PixelShuffle following the convolutional layer.
    HypoNeRV kernels exhibit the full spectrum of contribution magnitudes for various parts of the scene, ranging from low (dark red) to high (blue/purple).
    }
   \label{fig:xinc_hyponerv_grid}
\end{figure}

\begin{figure}[t]
  \centering
   \includegraphics[width=0.95\linewidth]{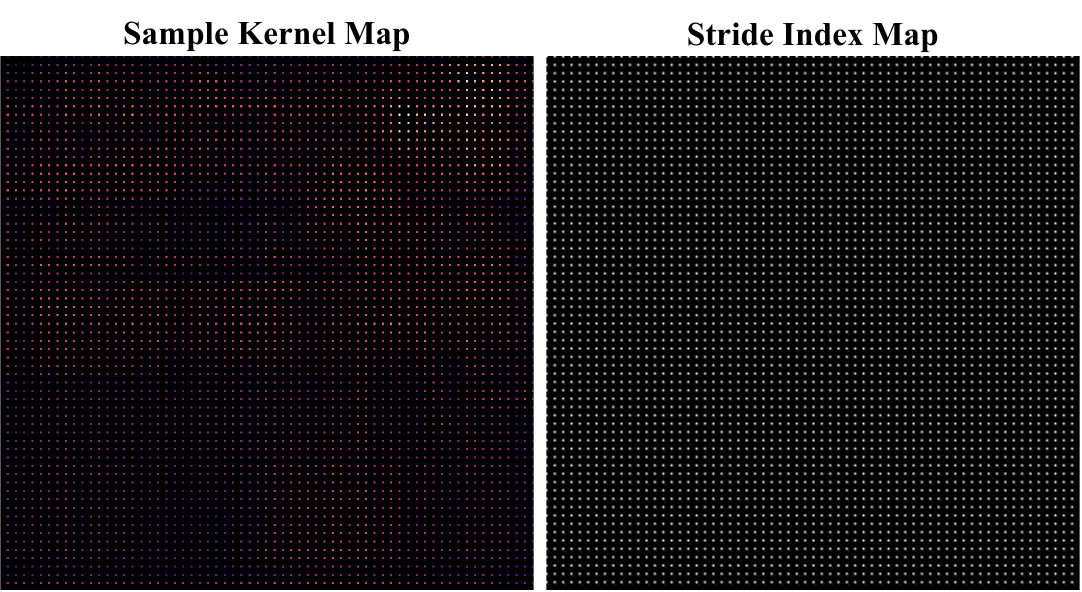}

    \caption{\textbf{Visualization of PixelShuffle}'s effect on kernel contribution patterns in the last (head) layer of HypoNeRV, for Jockey at 256x256. Left: Contribution map for a single kernel, showing the sparse activation pattern created by PixelShuffle's channel-to-space rearrangement. Black regions indicate locations where the kernel's contributions are zeroed out. Right: Reference map showing the potential contribution locations (white pixels) available to a kernel within a PixelShuffle group.
    Best viewed at full scale, otherwise the reader's PDF viewer may perform interpolation and render strange artifacts and make it appear as if the values on the right hand plot are not uniform. Similar distortions can occur for the plot on the left.}
   \label{fig:xinc_strided_index_map}
\end{figure}

\begin{figure}[t]
  \centering
   \includegraphics[width=0.95\linewidth]{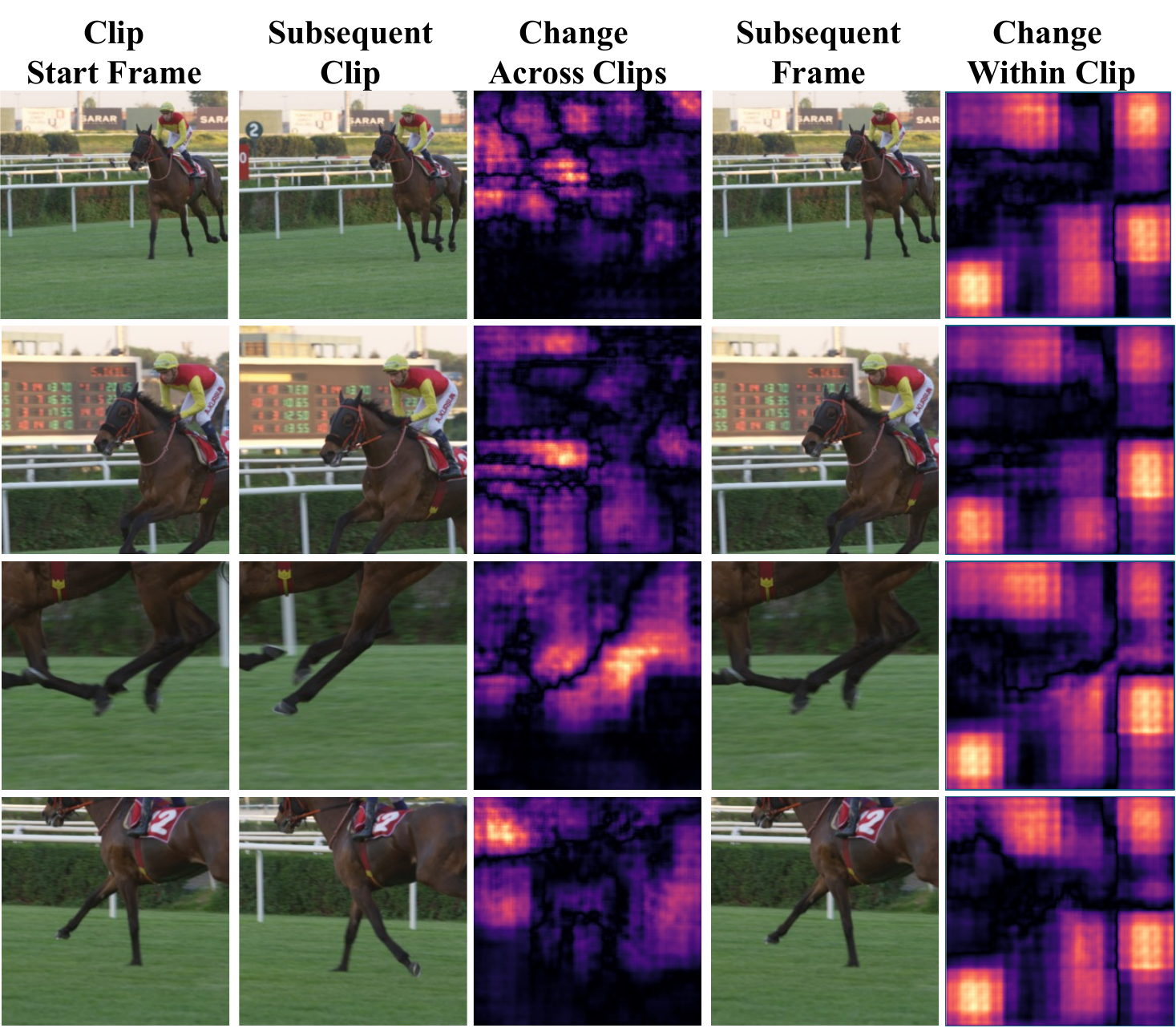}
    \caption{XINC motion analysis for fluctuation in total contribution of HypoNeRV neurons corresponding to two types of transitions: between first frames of consecutive clips (middle column) and between consecutive frames within the same clip (rightmost column). For transitions between clips, neurons rearrange their contributions in broader regions around areas of motion while ignoring static elements such as grass. For smaller motions between consecutive frames within a clip, neurons unexpectedly rearrange their contributions in static regions, showing little temporal faithfulness in their inner representation.}
  \label{fig:xinc_hyponet_motion}
\end{figure}

Figure~\ref{fig:xinc_hyponerv_grid} presents the XINC contribution maps for the last layer of HypoNeRV, which consists of a convolutional layer followed by PixelShuffle. Since the head layer contains over 750 neurons, we sort them by total contribution magnitude and display a subset of uniformly sampled representative kernels. Unlike the NeRV variants shown in Figure~\ref{fig:xinc_main}, HypoNeRV's head layer features a higher number of kernels, necessitated by the channel-to-space rearrangement imposed by PixelShuffle to output the 3 channel frame. The selected contributions span a diverse range, from high (blue/purple) to low (dark red), highlighting the variation in neuron responses across different parts of the scene.
Figure~\ref{fig:xinc_strided_index_map} illustrates how PixelShuffle modifies the contribution patterns of individual kernels in the head layer. The left panel shows a single kernel’s contribution map, where the stride-based shuffling operation enforces a sparse activation pattern with specific zeroed-out locations. The right panel provides a reference map indicating valid contribution sites within a PixelShuffle group, revealing the structured nature of these transformations. Each kernel in a group of $stride^2$ kernels operated on by PixelShuffle can contribute to exactly one position within the $stride \times stride$ local window. Due to the interpolation artifacts that may arise, this visualization is best examined at full resolution.

Figure~\ref{fig:xinc_motion_all_nervs} analyzes how HypoNeRV neurons adjust their contributions over time by comparing two types of transitions: (1) between the first frames of consecutive clips and (2) between consecutive frames within the same clip. In clip-to-clip transitions, contributions shift primarily in dynamic regions, reflecting scene changes while largely disregarding static backgrounds like grass. However, for frame-to-frame transitions within a clip, neurons unexpectedly redistribute their contributions even in stationary regions, suggesting an internal representation that does not strictly adhere to temporal consistency. This behavior indicates that while the model effectively captures broader scene dynamics, its learned representations may not maintain stable spatial allocations for finer-grained motion and preserve local temporal coherence.

\subsection{Supplementary Results on NeRV variants}

\begin{figure*}[t]
  \centering
   \includegraphics[width=.9\linewidth]{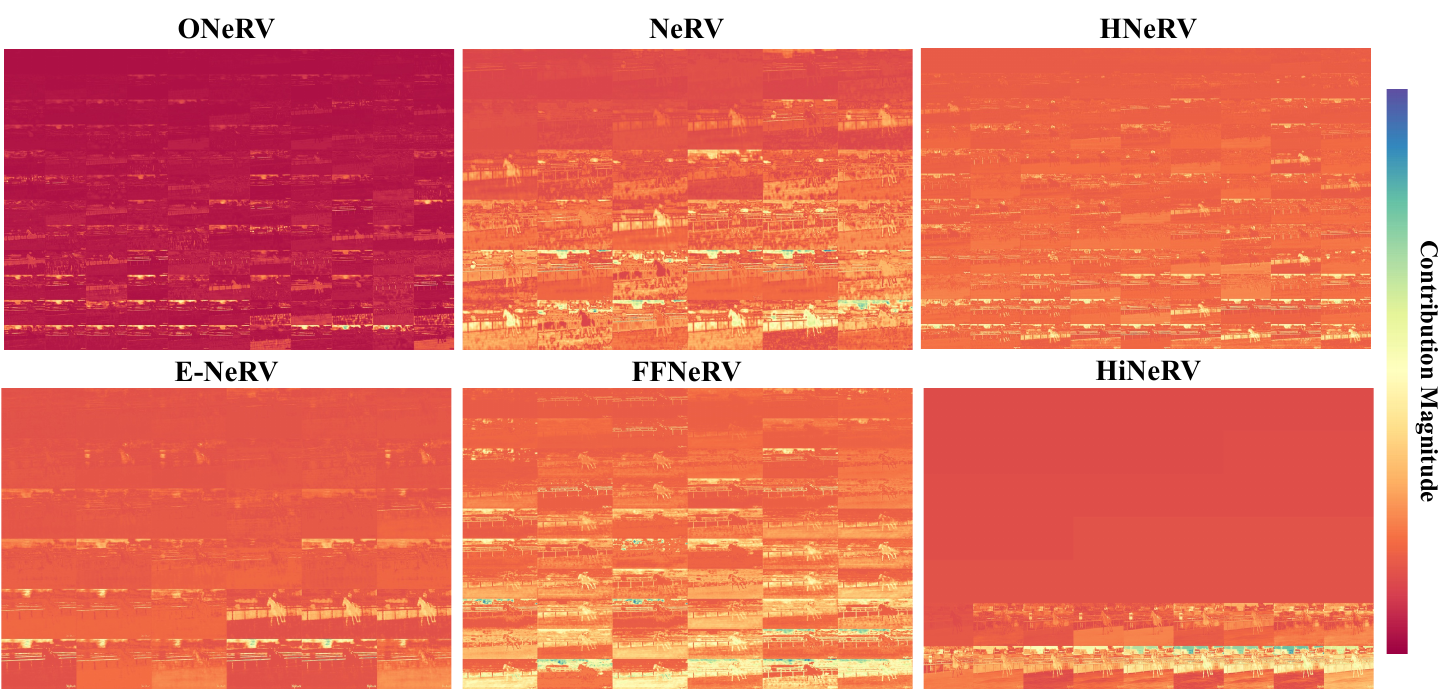}
   \caption{\textbf{XINC contribution maps} from the last (head) layer on Jockey at 1080p. We supplement Figure ~\ref{fig:xinc_main} by showing kernel contribution maps sorted by magnitude for additional NeRV variants - NeRV, HNeRV and E-NeRV.}
   \label{fig:xinc_suppl_all_nervs}
\end{figure*}

\begin{figure}[t]
  \centering
   \includegraphics[width=0.95\linewidth]{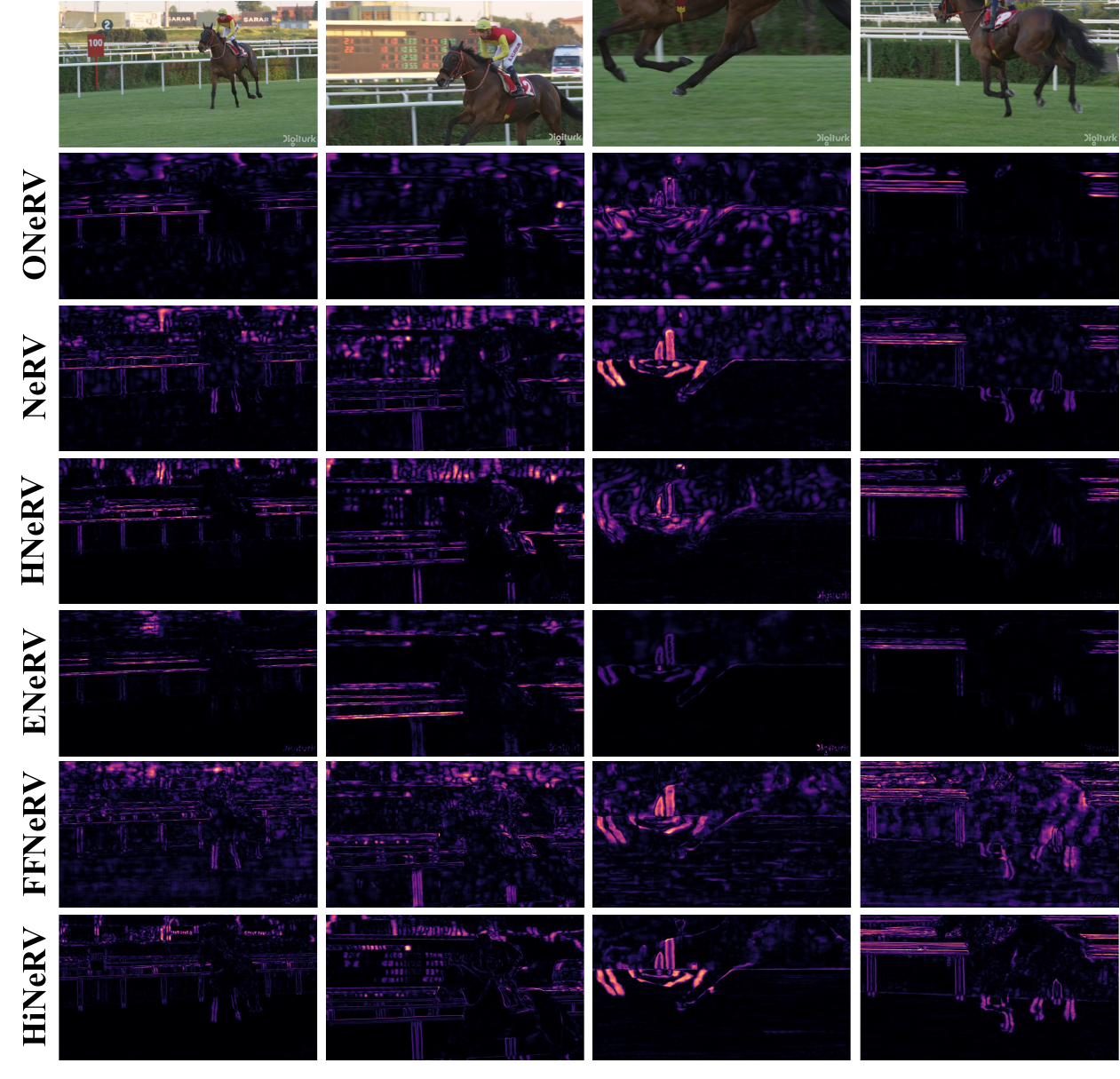}

    \caption{XINC motion analysis for the last (head) layer of all different NeRV variants, for Jockey at 1080p. We supplement Figure ~\ref{fig:xinc_motion} by visualizing the fluctuation in total contribution for kernels of additional NeRV variants - NeRV, HNeRV and E-NeRV.}
   \label{fig:xinc_motion_all_nervs}
\end{figure}

Figure~\ref{fig:xinc_suppl_all_nervs} and Figure~\ref{fig:xinc_motion_all_nervs} supplement Figure~\ref{fig:xinc_main} and Figure~\ref{fig:xinc_motion} respectively to extend the analysis to additional contribution maps and motion analysis results for NeRV, HNeRV and E-NeRV. NeRV has more structured changes driven by motion, whereas the weaker E-NeRV model demonstrates less pronounced adaptation. These findings further highlight the differences in representational capacity among the NeRV variants.


\begin{thebibliography}{60}
\providecommand{\natexlab}[1]{#1}
\providecommand{\url}[1]{\texttt{#1}}
\expandafter\ifx\csname urlstyle\endcsname\relax
  \providecommand{\doi}[1]{doi: #1}\else
  \providecommand{\doi}{doi: \begingroup \urlstyle{rm}\Url}\fi

\bibitem[Agustsson et~al.(2020)Agustsson, Minnen, Johnston, Balle, Hwang, and Toderici]{Agustsson_2020_CVPR}
Eirikur Agustsson, David Minnen, Nick Johnston, Johannes Balle, Sung~Jin Hwang, and George Toderici.
\newblock Scale-space flow for end-to-end optimized video compression.
\newblock In \emph{CVPR}, 2020.

\bibitem[Ba et~al.(2016)Ba, Kiros, and Hinton]{ba2016layernormalization}
Jimmy~Lei Ba, Jamie~Ryan Kiros, and Geoffrey~E. Hinton.
\newblock Layer normalization, 2016.

\bibitem[Chen et~al.(2021)Chen, He, Wang, Ren, Lim, and Shrivastava]{chen2021nerv}
Hao Chen, Bo He, Hanyu Wang, Yixuan Ren, Ser~Nam Lim, and Abhinav Shrivastava.
\newblock Nerv: Neural representations for videos.
\newblock \emph{Advances in Neural Information Processing Systems}, 34:\penalty0 21557--21568, 2021.

\bibitem[Chen et~al.(2022)Chen, Gwilliam, He, Lim, and Shrivastava]{chen2022cnerv}
Hao Chen, Matt Gwilliam, Bo He, Ser-Nam Lim, and Abhinav Shrivastava.
\newblock Cnerv: Content-adaptive neural representation for visual data, 2022.

\bibitem[Chen et~al.(2023)Chen, Gwilliam, Lim, and Shrivastava]{chen2023hnerv}
Hao Chen, Matthew Gwilliam, Ser-Nam Lim, and Abhinav Shrivastava.
\newblock Hnerv: A hybrid neural representation for videos.
\newblock In \emph{Proceedings of the IEEE/CVF Conference on Computer Vision and Pattern Recognition}, pages 10270--10279, 2023.

\bibitem[Chen et~al.(2024)Chen, Xie, Lim, and Shrivastava]{chen2024fastencodingdecodingimplicit}
Hao Chen, Saining Xie, Ser-Nam Lim, and Abhinav Shrivastava.
\newblock Fast encoding and decoding for implicit video representation, 2024.

\bibitem[Chen and Wang(2022)]{chen2022transformers}
Yinbo Chen and Xiaolong Wang.
\newblock Transformers as meta-learners for implicit neural representations, 2022.

\bibitem[Dupont et~al.(2021)Dupont, Goli{\'n}ski, Alizadeh, Teh, and Doucet]{dupont2021coin}
Emilien Dupont, Adam Goli{\'n}ski, Milad Alizadeh, Yee~Whye Teh, and Arnaud Doucet.
\newblock Coin: Compression with implicit neural representations.
\newblock \emph{arXiv preprint arXiv:2103.03123}, 2021.

\bibitem[Dupont et~al.(2022)Dupont, Loya, Alizadeh, Golinski, Teh, and Doucet]{dupont2022coin++}
Emilien Dupont, Hrushikesh Loya, Milad Alizadeh, Adam Golinski, Yee~Whye Teh, and Arnaud Doucet.
\newblock Coin++: Data agnostic neural compression.
\newblock \emph{arXiv preprint arXiv:2201.12904}, 1\penalty0 (2):\penalty0 4, 2022.

\bibitem[Girish et~al.(2023)Girish, Shrivastava, and Gupta]{girish2023shacirascalablehashgridcompression}
Sharath Girish, Abhinav Shrivastava, and Kamal Gupta.
\newblock Shacira: Scalable hash-grid compression for implicit neural representations, 2023.

\bibitem[Haydarov et~al.(2024)Haydarov, Muhamed, Shen, Lazarevic, Skorokhodov, Galappaththige, and Elhoseiny]{Haydarov_2024_CVPR}
Kilichbek Haydarov, Aashiq Muhamed, Xiaoqian Shen, Jovana Lazarevic, Ivan Skorokhodov, Chamuditha~Jayanga Galappaththige, and Mohamed Elhoseiny.
\newblock Adversarial text to continuous image generation.
\newblock In \emph{Proceedings of the IEEE/CVF Conference on Computer Vision and Pattern Recognition (CVPR)}, pages 6316--6326, 2024.

\bibitem[He et~al.(2023)He, Yang, Wang, Wu, Chen, Huang, Ren, Lim, and Shrivastava]{He_2023_CVPR}
Bo He, Xitong Yang, Hanyu Wang, Zuxuan Wu, Hao Chen, Shuaiyi Huang, Yixuan Ren, Ser-Nam Lim, and Abhinav Shrivastava.
\newblock Towards scalable neural representation for diverse videos.
\newblock In \emph{Proceedings of the IEEE/CVF Conference on Computer Vision and Pattern Recognition (CVPR)}, pages 6132--6142, 2023.

\bibitem[Hendrycks and Gimpel(2023)]{hendrycks2023gaussianerrorlinearunits}
Dan Hendrycks and Kevin Gimpel.
\newblock Gaussian error linear units (gelus), 2023.

\bibitem[Kay et~al.(2017)Kay, Carreira, Simonyan, Zhang, Hillier, Vijayanarasimhan, Viola, Green, Back, Natsev, Suleyman, and Zisserman]{kay2017kineticshumanactionvideo}
Will Kay, Joao Carreira, Karen Simonyan, Brian Zhang, Chloe Hillier, Sudheendra Vijayanarasimhan, Fabio Viola, Tim Green, Trevor Back, Paul Natsev, Mustafa Suleyman, and Andrew Zisserman.
\newblock The kinetics human action video dataset, 2017.

\bibitem[Khani et~al.(2021)Khani, Sivaraman, and Alizadeh]{khani2021efficient}
Mehrdad Khani, Vibhaalakshmi Sivaraman, and Mohammad Alizadeh.
\newblock Efficient video compression via content-adaptive super-resolution.
\newblock \emph{ICCV}, 2021.

\bibitem[Kim et~al.(2022{\natexlab{a}})Kim, Lee, Kim, Cho, and Han]{kim2022generalizable}
Chiheon Kim, Doyup Lee, Saehoon Kim, Minsu Cho, and Wook-Shin Han.
\newblock Generalizable implicit neural representations via instance pattern composers.
\newblock \emph{arXiv preprint arXiv:2211.13223}, 2022{\natexlab{a}}.

\bibitem[Kim et~al.(2023)Kim, Bauer, Theis, Schwarz, and Dupont]{kim2023c3highperformancelowcomplexityneural}
Hyunjik Kim, Matthias Bauer, Lucas Theis, Jonathan~Richard Schwarz, and Emilien Dupont.
\newblock C3: High-performance and low-complexity neural compression from a single image or video, 2023.

\bibitem[Kim et~al.(2024)Kim, Lee, and Kang]{Kim_2024}
Jina Kim, Jihoo Lee, and Je-Won Kang.
\newblock \emph{SNeRV: Spectra-Preserving Neural Representation for Video}, page 332–348.
\newblock Springer Nature Switzerland, 2024.

\bibitem[Kim et~al.(2022{\natexlab{b}})Kim, Yu, Lee, and Shin]{kim2022scalable}
Subin Kim, Sihyun Yu, Jaeho Lee, and Jinwoo Shin.
\newblock Scalable neural video representations with learnable positional features.
\newblock In \emph{Advances in Neural Information Processing Systems}, 2022{\natexlab{b}}.

\bibitem[Kingma and Ba(2017)]{kingma2017adammethodstochasticoptimization}
Diederik~P. Kingma and Jimmy Ba.
\newblock Adam: A method for stochastic optimization, 2017.

\bibitem[Kwan et~al.(2023)Kwan, Gao, Zhang, Gower, and Bull]{kwan2023hinerv}
Ho~Man Kwan, Ge Gao, Fan Zhang, Andrew Gower, and David Bull.
\newblock Hinerv: Video compression with hierarchical encoding-based neural representation.
\newblock In \emph{Advances in Neural Information Processing Systems}, pages 72692--72704. Curran Associates, Inc., 2023.

\bibitem[Kwan et~al.(2024)Kwan, Gao, Zhang, Gower, and Bull]{kwan2024nvrcneuralvideorepresentation}
Ho~Man Kwan, Ge Gao, Fan Zhang, Andrew Gower, and David Bull.
\newblock Nvrc: Neural video representation compression, 2024.

\bibitem[Ladune et~al.(2023)Ladune, Philippe, Henry, Clare, and Leguay]{ladune2023coolchiccoordinatebasedlowcomplexity}
Théo Ladune, Pierrick Philippe, Félix Henry, Gordon Clare, and Thomas Leguay.
\newblock Cool-chic: Coordinate-based low complexity hierarchical image codec, 2023.

\bibitem[Le~Gall(1991)]{mpeg}
Didier Le~Gall.
\newblock Mpeg: A video compression standard for multimedia applications.
\newblock \emph{Commun. ACM}, 1991.

\bibitem[Lee et~al.(2023)Lee, Rho, Ko, and Park]{Lee_2023}
Joo~Chan Lee, Daniel Rho, Jong~Hwan Ko, and Eunbyung Park.
\newblock Ffnerv: Flow-guided frame-wise neural representations for videos.
\newblock In \emph{Proceedings of the 31st ACM International Conference on Multimedia}, page 7859–7870. ACM, 2023.

\bibitem[Li et~al.(2021)Li, Li, and Lu]{li2021deepcontextualvideocompression}
Jiahao Li, Bin Li, and Yan Lu.
\newblock Deep contextual video compression, 2021.

\bibitem[Li et~al.(2022{\natexlab{a}})Li, Li, and Lu]{Li_2022}
Jiahao Li, Bin Li, and Yan Lu.
\newblock Hybrid spatial-temporal entropy modelling for neural video compression.
\newblock In \emph{Proceedings of the 30th ACM International Conference on Multimedia}. ACM, 2022{\natexlab{a}}.

\bibitem[Li et~al.(2023)Li, Li, and Lu]{Li_2023_CVPR}
Jiahao Li, Bin Li, and Yan Lu.
\newblock Neural video compression with diverse contexts.
\newblock In \emph{Proceedings of the IEEE/CVF Conference on Computer Vision and Pattern Recognition (CVPR)}, pages 22616--22626, 2023.

\bibitem[Li et~al.(2024)Li, Li, and Lu]{Li_2024_CVPR}
Jiahao Li, Bin Li, and Yan Lu.
\newblock Neural video compression with feature modulation.
\newblock In \emph{Proceedings of the IEEE/CVF Conference on Computer Vision and Pattern Recognition (CVPR)}, pages 26099--26108, 2024.

\bibitem[Li et~al.(2022{\natexlab{b}})Li, Wang, Pi, Xu, Mei, and Liu]{li2022enervexpediteneuralvideo}
Zizhang Li, Mengmeng Wang, Huaijin Pi, Kechun Xu, Jianbiao Mei, and Yong Liu.
\newblock E-nerv: Expedite neural video representation with disentangled spatial-temporal context, 2022{\natexlab{b}}.

\bibitem[Liu et~al.(2019)Liu, Chen, Lu, Shen, and Ma]{liu2019neural}
Haojie Liu, Tong Chen, Ming Lu, Qiu Shen, and Zhan Ma.
\newblock Neural video compression using spatio-temporal priors.
\newblock \emph{arXiv preprint arXiv:1902.07383}, 2019.

\bibitem[Maiya et~al.(2023)Maiya, Girish, Ehrlich, Wang, Lee, Poirson, Wu, Wang, and Shrivastava]{maiya2023nirvana}
Shishira~R Maiya, Sharath Girish, Max Ehrlich, Hanyu Wang, Kwot~Sin Lee, Patrick Poirson, Pengxiang Wu, Chen Wang, and Abhinav Shrivastava.
\newblock Nirvana: Neural implicit representations of videos with adaptive networks and autoregressive patch-wise modeling.
\newblock In \emph{Proceedings of the IEEE/CVF Conference on Computer Vision and Pattern Recognition}, pages 14378--14387, 2023.

\bibitem[Maiya et~al.(2024)Maiya, Gupta, Gwilliam, Ehrlich, and Shrivastava]{maiya2024latent}
Shishira~R Maiya, Anubhav Gupta, Matthew Gwilliam, Max Ehrlich, and Abhinav Shrivastava.
\newblock Latent-inr: A flexible framework for implicit representations of videos with discriminative semantics.
\newblock In \emph{European Conference on Computer Vision}, pages 285--302. Springer, 2024.

\bibitem[Mentzer et~al.(2019)Mentzer, Agustsson, Tschannen, Timofte, and Van~Gool]{mentzer2019practical}
Fabian Mentzer, Eirikur Agustsson, Michael Tschannen, Radu Timofte, and Luc Van~Gool.
\newblock Practical full resolution learned lossless image compression.
\newblock In \emph{Proceedings of the IEEE Conference on Computer Vision and Pattern Recognition (CVPR)}, 2019.

\bibitem[Mercat et~al.(2020)Mercat, Viitanen, and Vanne]{mercat2020uvg}
Alexandre Mercat, Marko Viitanen, and Jarno Vanne.
\newblock Uvg dataset: 50/120fps 4k sequences for video codec analysis and development.
\newblock In \emph{Proceedings of the 11th ACM multimedia systems conference}, pages 297--302, 2020.

\bibitem[Mildenhall et~al.(2020)Mildenhall, Srinivasan, Tancik, Barron, Ramamoorthi, and Ng]{mildenhall2020nerf}
Ben Mildenhall, Pratul~P. Srinivasan, Matthew Tancik, Jonathan~T. Barron, Ravi Ramamoorthi, and Ren Ng.
\newblock Nerf: Representing scenes as neural radiance fields for view synthesis, 2020.

\bibitem[Müller et~al.(2022)Müller, Evans, Schied, and Keller]{M_ller_2022}
Thomas Müller, Alex Evans, Christoph Schied, and Alexander Keller.
\newblock Instant neural graphics primitives with a multiresolution hash encoding.
\newblock \emph{ACM Transactions on Graphics}, 41\penalty0 (4):\penalty0 1–15, 2022.

\bibitem[Padmanabhan et~al.(2024)Padmanabhan, Gwilliam, Kumar, Maiya, Ehrlich, and Shrivastava]{Padmanabhan_2024_CVPR}
Namitha Padmanabhan, Matthew Gwilliam, Pulkit Kumar, Shishira~R Maiya, Max Ehrlich, and Abhinav Shrivastava.
\newblock Explaining the implicit neural canvas: Connecting pixels to neurons by tracing their contributions.
\newblock In \emph{Proceedings of the IEEE/CVF Conference on Computer Vision and Pattern Recognition (CVPR)}, pages 10957--10967, 2024.

\bibitem[Rippel et~al.(2019)Rippel, Nair, Lew, Branson, Anderson, and Bourdev]{Rippel_2019_ICCV}
Oren Rippel, Sanjay Nair, Carissa Lew, Steve Branson, Alexander~G. Anderson, and Lubomir Bourdev.
\newblock Learned video compression.
\newblock In \emph{ICCV}, 2019.

\bibitem[Rippel et~al.(2021)Rippel, Anderson, Tatwawadi, Nair, Lytle, and Bourdev]{rippel2021elfvc}
Oren Rippel, Alexander~G. Anderson, Kedar Tatwawadi, Sanjay Nair, Craig Lytle, and Lubomir Bourdev.
\newblock Elf-vc: Efficient learned flexible-rate video coding.
\newblock In \emph{ICCV}, 2021.

\bibitem[Saethre et~al.(2024)Saethre, Azevedo, and Schroers]{Saethre_2024_CVPR}
Jens~Eirik Saethre, Roberto Azevedo, and Christopher Schroers.
\newblock Combining frame and gop embeddings for neural video representation.
\newblock In \emph{Proceedings of the IEEE/CVF Conference on Computer Vision and Pattern Recognition (CVPR)}, pages 9253--9263, 2024.

\bibitem[Saragadam et~al.(2023)Saragadam, LeJeune, Tan, Balakrishnan, Veeraraghavan, and Baraniuk]{saragadam2023wire}
Vishwanath Saragadam, Daniel LeJeune, Jasper Tan, Guha Balakrishnan, Ashok Veeraraghavan, and Richard~G. Baraniuk.
\newblock Wire: Wavelet implicit neural representations, 2023.

\bibitem[Shi et~al.(2016)Shi, Caballero, Huszár, Totz, Aitken, Bishop, Rueckert, and Wang]{shi2016realtimesingleimagevideo}
Wenzhe Shi, Jose Caballero, Ferenc Huszár, Johannes Totz, Andrew~P. Aitken, Rob Bishop, Daniel Rueckert, and Zehan Wang.
\newblock Real-time single image and video super-resolution using an efficient sub-pixel convolutional neural network, 2016.

\bibitem[Sitzmann et~al.(2020)Sitzmann, Martel, Bergman, Lindell, and Wetzstein]{sitzmann2020implicit}
Vincent Sitzmann, Julien Martel, Alexander Bergman, David Lindell, and Gordon Wetzstein.
\newblock Implicit neural representations with periodic activation functions.
\newblock \emph{Advances in neural information processing systems}, 33:\penalty0 7462--7473, 2020.

\bibitem[Skorokhodov et~al.(2021)Skorokhodov, Ignatyev, and Elhoseiny]{skorokhodov2021adversarialgenerationcontinuousimages}
Ivan Skorokhodov, Savva Ignatyev, and Mohamed Elhoseiny.
\newblock Adversarial generation of continuous images, 2021.

\bibitem[Soomro et~al.(2012)Soomro, Zamir, and Shah]{soomro2012ucf101dataset101human}
Khurram Soomro, Amir~Roshan Zamir, and Mubarak Shah.
\newblock Ucf101: A dataset of 101 human actions classes from videos in the wild, 2012.

\bibitem[Strümpler et~al.(2022)Strümpler, Postels, Yang, van Gool, and Tombari]{strümpler2022implicitneuralrepresentationsimage}
Yannick Strümpler, Janis Postels, Ren Yang, Luc van Gool, and Federico Tombari.
\newblock Implicit neural representations for image compression, 2022.

\bibitem[Sullivan et~al.(2012)Sullivan, Ohm, Han, and Wiegand]{hevc}
Gary~J. Sullivan, Jens-Rainer Ohm, Woo-Jin Han, and Thomas Wiegand.
\newblock Overview of the high efficiency video coding (hevc) standard.
\newblock \emph{IEEE Transactions on Circuits and Systems for Video Technology}, 2012.

\bibitem[Tancik et~al.(2020)Tancik, Srinivasan, Mildenhall, Fridovich-Keil, Raghavan, Singhal, Ramamoorthi, Barron, and Ng]{tancik2020fourfeat}
Matthew Tancik, Pratul~P. Srinivasan, Ben Mildenhall, Sara Fridovich-Keil, Nithin Raghavan, Utkarsh Singhal, Ravi Ramamoorthi, Jonathan~T. Barron, and Ren Ng.
\newblock Fourier features let networks learn high frequency functions in low dimensional domains.
\newblock \emph{NeurIPS}, 2020.

\bibitem[Wang et~al.(2003)Wang, Simoncelli, and Bovik]{wang2003multiscale}
Zhou Wang, Eero~P Simoncelli, and Alan~C Bovik.
\newblock Multiscale structural similarity for image quality assessment.
\newblock In \emph{The Thrity-Seventh Asilomar Conference on Signals, Systems \& Computers, 2003}, pages 1398--1402. Ieee, 2003.

\bibitem[Wiegand et~al.(2003)Wiegand, Sullivan, Bjontegaard, and Luthra]{H264}
T. Wiegand, G.J. Sullivan, G. Bjontegaard, and A. Luthra.
\newblock Overview of the h.264/avc video coding standard.
\newblock \emph{IEEE Transactions on Circuits and Systems for Video Technology}, 2003.

\bibitem[Wu et~al.(2024)Wu, Quan, He, Lai, Li, Yu, Lin, and Yang]{wu2024qs}
Chang Wu, Guancheng Quan, Gang He, Xin-Quan Lai, Yunsong Li, Wenxin Yu, Xianmeng Lin, and Cheng Yang.
\newblock Qs-nerv: Real-time quality-scalable decoding with neural representation for videos.
\newblock In \emph{Proceedings of the 32nd ACM International Conference on Multimedia}, pages 2584--2592, 2024.

\bibitem[Xu et~al.(2022)Xu, Wang, Jiang, Fan, and Wang]{xu2022signal}
Dejia Xu, Peihao Wang, Yifan Jiang, Zhiwen Fan, and Zhangyang Wang.
\newblock Signal processing for implicit neural representations, 2022.

\bibitem[Xu et~al.(2024)Xu, Feng, Qin, Ge, Peng, and Wang]{xu2024vqnervvectorquantizedneural}
Yunjie Xu, Xiang Feng, Feiwei Qin, Ruiquan Ge, Yong Peng, and Changmiao Wang.
\newblock Vq-nerv: A vector quantized neural representation for videos, 2024.

\bibitem[Yan et~al.(2024)Yan, Ke, Zhou, Qiu, Shi, and Jiang]{Yan_2024_CVPR}
Hao Yan, Zhihui Ke, Xiaobo Zhou, Tie Qiu, Xidong Shi, and Dadong Jiang.
\newblock Ds-nerv: Implicit neural video representation with decomposed static and dynamic codes.
\newblock In \emph{Proceedings of the IEEE/CVF Conference on Computer Vision and Pattern Recognition (CVPR)}, pages 23019--23029, 2024.

\bibitem[Yu et~al.(2022)Yu, Tack, Mo, Kim, Kim, Ha, and Shin]{yu2022generatingvideosdynamicsawareimplicit}
Sihyun Yu, Jihoon Tack, Sangwoo Mo, Hyunsu Kim, Junho Kim, Jung-Woo Ha, and Jinwoo Shin.
\newblock Generating videos with dynamics-aware implicit generative adversarial networks, 2022.

\bibitem[Zhang et~al.(2024)Zhang, Yang, He, Ge, Xu, Wang, Qin, and Zhang]{Zhang_2024_CVPR}
Xinjie Zhang, Ren Yang, Dailan He, Xingtong Ge, Tongda Xu, Yan Wang, Hongwei Qin, and Jun Zhang.
\newblock Boosting neural representations for videos with a conditional decoder.
\newblock In \emph{Proceedings of the IEEE/CVF Conference on Computer Vision and Pattern Recognition (CVPR)}, pages 2556--2566, 2024.

\bibitem[Zhang et~al.(2021)Zhang, van Rozendaal, Brehmer, Nagel, and Cohen]{zhang2021implicitneuralvideocompression}
Yunfan Zhang, Ties van Rozendaal, Johann Brehmer, Markus Nagel, and Taco Cohen.
\newblock Implicit neural video compression, 2021.

\bibitem[Zhao et~al.(2023)Zhao, Asif, and Ma]{Zhao_2023_CVPR}
Qi Zhao, M.~Salman Asif, and Zhan Ma.
\newblock Dnerv: Modeling inherent dynamics via difference neural representation for videos.
\newblock In \emph{Proceedings of the IEEE/CVF Conference on Computer Vision and Pattern Recognition (CVPR)}, pages 2031--2040, 2023.

\bibitem[Zhao et~al.(2024)Zhao, Asif, and Ma]{Zhao_2024_CVPR}
Qi Zhao, M.~Salman Asif, and Zhan Ma.
\newblock Pnerv: Enhancing spatial consistency via pyramidal neural representation for videos.
\newblock In \emph{Proceedings of the IEEE/CVF Conference on Computer Vision and Pattern Recognition (CVPR)}, pages 19103--19112, 2024.

\end{thebibliography}
\end{document}